\documentclass[biber]{nowfnt} 

\usepackage[USenglish]{babel}
\usepackage{csquotes}

\usepackage{amsmath}
\usepackage{algpseudocode}
\usepackage{caption}
\usepackage{subcaption}
\usepackage{xcolor}
\usepackage{todonotes}
\usepackage{algorithm}
\usepackage{algpseudocode}

\title{Hyperparameter 
Optimization \\in Machine Learning}

\maintitleauthorlist{
Luca Franceschi \and
Michele Donini \and
Valerio Perrone \and
Aaron Klein \and
C\'{e}dric Archambeau \and
Matthias Seeger \and
Massimiliano Pontil \and
Paolo Frasconi
}

\issuesetup
{%
 issue         = Preprint,
 pubyear       = 2024,
 }

\usepackage{mwe}

\author[1,6]{Franceschi,Luca}
\author[2]{Donini,Michele}
\author[1]{Perrone,Valerio}
\author[3]{Klein,Aaron}
\author[2]{Archambeau,C\'{e}dric}
\author[1]{Seeger,Matthias}
\author[4]{Pontil,Massimiliano}
\author[5]{Frasconi,Paolo}

\affil[1]{Amazon Web Services; work not related to position at Amazon}
\affil[2]{Helsing}
\affil[3]{ScaDS.AI, Leipzig University}
\affil[4]{University College London \& Istituto Italiano di Tecnologia}
\affil[5]{Università di Firenze}
\affil[6]{Work partially done while LF was at UCL and IIT}

\articledatabox{\nowfntstandardcitation}

\usepackage{bm}

\usepackage{tikz}
\usepackage{txfonts}
\usepackage{comment}
\usetikzlibrary{shapes,arrows}

\DeclareMathAlphabet\mathbfcal{OMS}{cmsy}{b}{n}

\DeclareMathOperator*{\argmax}{arg\,max}
\DeclareMathOperator*{\argmin}{arg\,min}

\makeatletter
\newcommand{\bigcomp}{%
  \DOTSB
  \mathop{\vphantom{\sum}\mathpalette\bigcomp@\relax}%
  \slimits@
}
\newcommand{\bigcomp@}[2]{%
  \begingroup\m@th
  \sbox\z@{$#1\sum$}%
  \setlength{\unitlength}{0.9\dimexpr\ht\z@+\dp\z@}%
  \vcenter{\hbox{%
    \begin{picture}(1,1)
    \bigcomp@linethickness{#1}
    \put(0.5,0.5){\circle{1}}
    \end{picture}%
  }}%
  \endgroup
}
\newcommand{\bigcomp@linethickness}[1]{%
  \linethickness{%
      \ifx#1\displaystyle \
       2\fontdimen8\textfont\else
      \ifx#1\textstyle 1.65\fontdimen8\textfont\else
      \ifx#1\scriptstyle 1.65\fontdimen8\scriptfont\else
      1.65\fontdimen8\scriptscriptfont\fi\fi\fi 3
  }%
}

\def\transpose#1{#1^\intercal}

\newcommand{\weightspace}{\mathcal{W}}

\usepackage{dsfont}
\def\RSet{\mathds{R}}

\def\NSet{\mathds{N}}

\DeclareMathOperator{\expectation}{\mathds{E}}

\usepackage{commath} 

\def\ONE#1{\mathds{1}\!\left\{#1\right\}}
\def\transpose#1{#1^\mathsf{T}}

\usepackage{xcolor}

\newcommand{\inputspace}{\mathcal{X}}

\newcommand{\outputspace}{\mathcal{Y}}

\newcommand{\valmetric}{\mathscr{M}}
\newcommand{\hypothesis}{h}
\newcommand{\hypothesisspace}{\mathcal{H}}

\newcommand{\dotprod}[2]{\langle #1 , #2 \rangle}

\newcommand{\learningalgo}{A}

\usepackage[scr=rsfso]{mathalfa}
\newcommand{\setofdata}{\mathscr{D}}

\newcommand{\hpoap}{\upsilon}

\newcommand{\Dtrain}{\mathcal{D}}
\newcommand{\Dval}{\mathcal{V}}

\newcommand{\objtrain}{\mathcal{J}}

\newcommand{\surrogm}{\hat{f}}

\def\lambdacomp#1{\lambda^{(#1)}}
\def\Lambdacomp#1{\Lambda^{(#1)}}

\def\barLambdacomp#1{\bar{\Lambda}^{(#1)}}
\def\hpoapcomp#1{\hpoap^{(#1)}}

\usepackage{tabularx}
\usepackage{booktabs}

\makeatletter
\newcommand{\insertheaderrule}{\rlap{\rule[-.3\normalbaselineskip]{\textwidth}{.4pt}}}
\let\old@evenhead\@evenhead \let\old@oddhead\@oddhead
\def\@evenhead{\insertheaderrule\old@evenhead}
\def\@oddhead{\insertheaderrule\old@oddhead}
\makeatother

\usepackage{fixme}
\fxsetup{
    status=draft,
    author=,
    layout=inline,
    theme=color
}
\definecolor{fxnote}{HTML}{268bd2}
\definecolor{fxtarget}{RGB}{220,50,47}

\newcommand{\taskloss}{\mathcal{L}}

\newcommand{\ffid}[2]{f_{#2}(#1)}
\newcommand{\bmin}{b_{min}}
\newcommand{\bmax}{b_{max}}

\begin{document}

\makeabstracttitle

\begin{abstract}
Hyperparameters are configuration variables controlling the 
behavior of machine learning algorithms. 
They are ubiquitous in machine learning and artificial intelligence and the choice of their values determines the effectiveness of systems based on these technologies.
Manual hyperparameter search is often time-consuming and
becomes infeasible when the number of hyperparameters is large. 
Automating the search is an important step  towards advancing, streamlining, and systematizing machine learning, freeing researchers and practitioners alike from the
burden of finding a good set of hyperparameters 
by trial and error.
In this survey, we present a unified treatment of hyperparameter optimization, providing the reader with examples, insights into the state-of-the-art,
and numerous links to further reading. 
We cover the main families of techniques to automate hyperparameter search, often referred to as hyperparameter optimization or tuning, including random and quasi-random search, bandit-, model-, population-, and gradient-based approaches. 
We further discuss extensions, including online, constrained, and multi-objective formulations, touch upon connections with other fields, such as meta-learning and neural architecture search, and conclude with open questions and future research directions.
\end{abstract}

\chapter{Introduction}
\label{ch1}

In this chapter, we informally introduce the main subject of this monograph:
algorithms for optimizing hyperparameters. Hyperparameters are configuration variables that are
ubiquitous in machine learning methods and can strongly influence
generalization and overall performance. 
The wide array of existing algorithms for
optimizing hyperparameters is the product of several years of research.
Their adoption is already widespread in academia and in the industry.
Our goal is to provide a comprehensive introduction to the fundamentals of
these methodologies, with the aim of encouraging their broader adoption -- making 
them standard practice —- and 
providing a solid foundation for further research in the field.
Such efforts help strengthen and
advance the current state-of-the-art, foster transparency and improve
the reproducibility of scientific and engineering results. This is a timely
necessity as the application of AI has rapidly expanded during the last
decade and has solidified its position as a primary driver of innovation in
many areas of science and industry.

The digitalization of human activities and interactions has led to the
generation of data at an unprecedented scale.
On the one hand, data can now be stored more affordably, thanks to
advancements in microelectronics and the emergence of cloud computing. On
the other hand, progress in hardware and low-power chip design have led to
an exponential increase in computing capabilities for both cloud
infrastructure, notably graphical processing units (GPUs), and on edge devices, like mobile phones.
The convergence of these two key trends has been instrumental in the
pervasive adoption of machine learning. Recent
breakthroughs in algorithms and architectures for deep learning have further
reinforced this foundation. Learning algorithms can now sift through vast
amounts of data to discover and extract patterns that can then be used to
guide decisions with little or even no human intervention. But this is
largely due to a dramatic increase in model complexity, both in terms of
size and structure. Modern architectures often comprise several intertwined
modules, each introducing its own set of hyperparameters.
Without systematic tools, configuring these hyperparameters jointly 
can be complex and overwhelming.

One notable and often cited illustration of the recent advances made in
machine learning is AlphaGo~\citep{silver:2017:mastering_go}, a computer 
program developed at the London-based
company \emph{DeepMind}. AlphaGo won against one of the world champions 
of the game of Go, Lee Sedol, in 2016. This event generated considerable 
press coverage (and was even made into a movie) as it was believed at the time 
that no computer program could win a game of Go against a human before 
several decades to come. What is less known is that the success of AlphaGo 
critically depended on how a set of hyperparameters were adjusted
automatically by another computer
program~\citep{chen:2018:bayesian_optimization_alphago}. This computer 
program relied on Bayesian optimization (see Section~\ref{sec:bk:hpo:bayesopt}), a general black-box technique
that in this case sequentially predicts
and then evaluates the performance of
learning algorithms, such as those behind
AlphaGo, when their hyperparameters are set to specific values.

\section{A Simple Illustrative Example}

\begin{figure}
  \centering
  \begin{subfigure}[b]{0.45\textwidth}
    \centering
    \includegraphics[width=\textwidth]{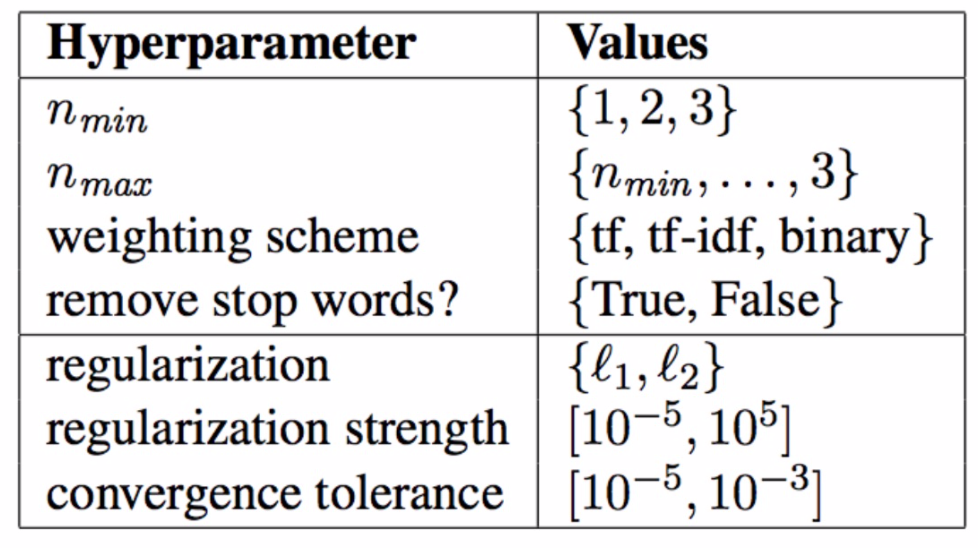}
    \caption{Hyperparameters.}
    \label{tab: sentiment analysis hyperparams}
  \end{subfigure}
  \hfill
  \begin{subfigure}[b]{0.45\textwidth}
    \centering
    \includegraphics[width=0.75\textwidth]{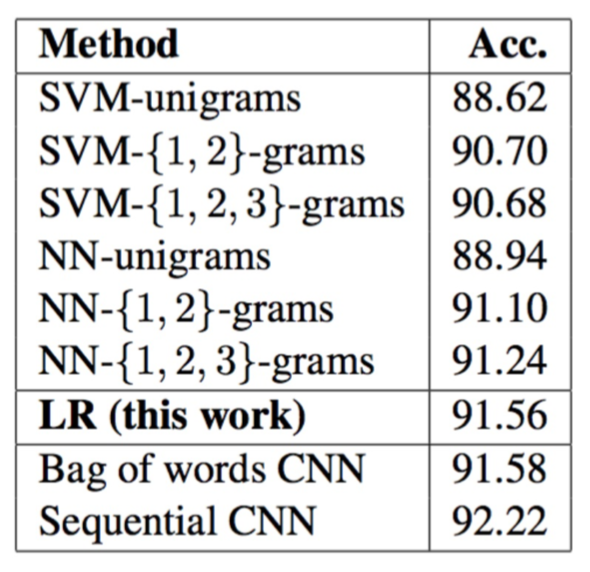}
    \caption{Accuracy.}
    \label{tab: sentiment analysis results}
  \end{subfigure}
  \caption{Results on sentiment analysis reported in
    \citep{yogotama2015-emnlp15}. Hyperparameters and their domains (both
    continuous and discrete) are listed on the left. Here
    $[n_{min},n_{max}]$ denotes the $n$-gram range. SVM: support vector
    machine; LR: logistic regression; NN: multi-layer perceptron; CNN:
    convolutional neural network.}
\end{figure}

To illustrate the practical importance of carefully selecting the hyperparameters, 
we consider the problem of \emph{sentiment analysis}. \citet{yogotama2015-emnlp15} 
studied the effect of hyperparameters in this context. 
More specifically, the authors framed the problem as a standard binary classification problem as was
commonly done in the literature, where the task of the classifier is to predict whether the text expresses a negative or a positive sentiment. 
They compared simple logistic regression, trained
by stochastic gradient descent, to convolutional neural networks, which achieved
state-of-the-art results at the time of the publication. Figure~\ref{tab: sentiment analysis 
hyperparams} shows the hyperparameters that were searched over. These included 
the choice of text features used (e.g., removal of stop words or not), the type of 
regularization (e.g., $L_1$ or $L_2$) and optimization algorithm parameters
(e.g., convergence tolerance). The experimental results they reported on the Amazon 
electronics dataset are reproduced in Figure~\ref{tab: sentiment analysis results}. 
Interestingly, they showed that an optimized logistic regression
with a bag of words representation of the text
performed similarly to a
convolutional neural network and
only slightly worse than a sequential convolutional neural network.
Hence, by carefully optimizing the hyperparameters, the authors could achieve a result close to the state-of-the-art, yet using a much simpler model.
In Section~\ref{sec:lrhpo}, we will come back to this example, explaining in detail the role of individual hyperparameters.

As this simple example reveals, claims about state-of-the-art performance
improvements need to be interpreted carefully: empirical results can vary
greatly depending on the chosen hyperparameters, with a possibly significant
impact on the conclusions drawn.
Unfortunately, it is not uncommon for published results to omit values for
specific hyperparameters used in the experiments, or to omit details
explaining how these values were chosen. In these cases, reproducibility may
be hindered~\citep{haibe-kains:2020:transparency}.

\section{What Is Hyperparameter Optimization?}
\label{sec:1:whatishpo}
A typical supervised learning algorithm receives training data and
outputs a predictor (e.g., a mapping from textual reviews
to binary sentiments, as in our previous example). 
The quality of the predictor can then be
measured on new data (the validation set) using an evaluation
metric, such as the error rate.
Since the predictor depends on the chosen
hyperparameters, the validation error also depends on those hyperparameters.
The mapping from hyperparameter values to the performance measured on
the validation set is called the \textit{response function}. 

In a nutshell, hyperparameter
optimization (HPO) consists in finding the hyperparameters that optimize
the response function. What distinguishes the HPO problem from conventional
optimization problems, such as those arising in convex or combinatorial
optimization, is its \textit{nested nature}: evaluating the response
function at a specific point requires running the learning algorithm,
which in turn often involves solving another optimization problem to fit a model to
the available (training) data.

\section{Why Is Hyperparameter Optimization Difficult?}

First, the nested nature of hyperparameter optimization problems means that,
in most cases, the response function is not available in closed 
form, but is defined only implicitly.
The function may be stochastic (due to the inherent randomness of
the training algorithm), have multiple local optima, exhibit wide
flat regions, or be non-smooth, non-continuous, or non-differentiable.
These characteristics rule out the application of standard gradient-based optimization algorithms.
Moreover, evaluating the response function at a given hyperparameter assignment
can be very expensive:
for instance, training a modern deep architecture may require significant compute,
ranging from several hours to several days on a GPU cluster.
As a result, much of the field has focused on developing algorithms that make minimal assumptions, such as only requiring \textit{observability} for instance.
This perspective becomes clearer when we return to the example of AlphaGo: it does not matter whether the mathematical
expression linking the hyperparameters of AlphaGo (i.e., how the algorithm
is configured) to the probability of winning a game is explicitly known; what matters
is that, for a given set of hyperparameters, we
can observe whether AlphaGo won or lost the game.

Second, the hyperparameter search space is often complex and heterogeneous.
Some hyperparameters could be continuous (e.g., the learning rate), but others could
be integer-valued (e.g., the number of layers in a deep neural network) or categorical
to encode choices.
Additional structure can also arise in the search space. 
Notably, \textit{conditional} hyperparameters are variables that are only
relevant depending on the value taken by others. For example, 
the number of units in the $j$-th layer of a neural network is only
relevant if the network has at least $j$ layers.
If the domain to be searched is a small discrete set, an exhaustive search
can be performed by evaluating all possible hyperparameter configurations
within this set. 
Unfortunately, this is rarely the case. 
In most scenarios, the
number of configurations can be infinite or even uncountable.

These challenges mean that, in practice, the focus of HPO is often
to find ``reasonably good'' configurations as quickly as possible ---
which could still mean days of computation --- rather than to find the global
optimum of the response function. 
Hence, HPO consists in determining a small set of promising
hyperparameters on which to evaluate the response function.

\section{Historical Remarks}
Although HPO has significantly developed in recent years, especially after
the rise of deep learning, its foundations are much older. Here we briefly
summarize some important landmarks. Model selection in statistics attracted
significant attention in the 1970s and it is perhaps the earliest precursor
to HPO in machine learning. Indeed, HPO itself is sometimes referred to as
``model selection'' -- although the expression originally only denoted the
simpler problem of finding the appropriate \textit{dimension} of a model.
Unlike modern approaches to HPO, which are based on a validation set, early
model selection methods, such as Akaike's information
criterion~\citep{akaike:1974:new_look_statistical} and the Bayesian
information criterion~\citep{schwarz1978estimating}, were based on
the idea of penalizing model complexity by adding a term to the likelihood
function. 
In the context of small, analytically treatable model, 
penalties make sense as they avoid 
always selecting the largest (best fitting)
model~\citep{schwarz1978estimating}. A closely related idea, rooted in
information theory, is the minimum description length
principle~\citep{rissanen1978modeling}, where model complexity is measured
by the number of bits required to encode it.

Interestingly, some algorithmic ideas behind modern HPO methods are even
older and can be traced back to well before the 1970s. The HPO problem in
machine learning is related to optimal experimental
design, an area that has a long tradition in statistics since the
introduction of the \emph{response surface methodology}
\citep{box:1951:experimental_attainment_optimum};
see~\citep{myers16:_respon} for a recent account. One of the core ideas in
that context is 
the approximation, by means of a surrogate model, of an otherwise
unknown true response function that needs to be optimized. Model-based
algorithms for HPO share the very same idea and the HPO objective function
draws its name from these methodologies.

Other ideas behind modern HPO originate from a community at the
intersection between AI and operations research, interested in a different
class of problems: the configuration of heuristics and meta-heuristics in
\emph{problem solving}. The goal in these cases is not generalization as in
machine learning, but finding solutions quickly, for example to certain instances of
(weighted) Boolean satisfiability. The LEX
system~\citep{mitchell:1983:learning_experimentation_acquiringa} already
embodied some ideas behind model-based approaches to HPO, which we cover in
Chapter~\ref{ch:mbm}. LEX ran the to-be-configured algorithm to produce
examples from which a \emph{generalizer} is trained to produce better
heuristics. Later, sequential model-based configuration (SMAC) was proposed
to solve global optimization and propositional satisfiability
problems~\citep{hutter:2009:experimental_investigation_modelbased,%
  hutter_sequential_2011}. It was eventually applied to supervised
learning~\citep{snoek_practical_2012,%
  thornton2013auto}. SMAC is related to Bayesian
optimization~\citep{mockus:1975:bayesian_methods_seeking,%
  sacks:1989:design_analysis_computer,%
  jones1998efficient}
as it is one of the
techniques at the core of current model-based HPO approaches.

Yet another class of early approaches is based on the idea of racing, where
candidate configurations are run in parallel and, as soon as sufficient
statistical evidence is collected against a candidate, the run is stopped
and the candidate replaced by a new one. This idea was initially formulated
for model selection in machine
learning~\citep{maron:1994:hoeffding_races_accelerating}, and then applied
to configuration of
meta-heuristics~\citep{birattari:2002:racing_algorithm_configuring}. Modern
multi-fidelity strategies for HPO also
discard unpromising candidates but in a more general setting (see Chapter~\ref{ch:multi_fidelity}).

Finally, population-based algorithms also originated in the area of problem
solving with seminal works of \citet{rechenberg65} and
\citet{fogel:1965:intelligent_decisionmaking_simulation}, boosted by the
introduction of genetic inspired operators by \citet{holland75}. While their
application to optimization in machine learning has a long tradition, the
use of population-based algorithms in HPO is more recent and preceded by the
idea of gradient-based approaches~\citep{bengio_gradient-based_2000}.

\section{Outline of the Monograph}

In this chapter, we introduced the HPO problem, motivating its importance and its place in machine learning.
In the next chapter, we will formalize many concept introduced here and further present several examples of hyperparameters and HPO problems.
In Chapter~\ref{ch:ea}, we will introduce elementary
algorithms that search over a predefined or random 
grid of hyperparameters. 
While conceptually simple, random search strategies 
based on uniform sampling are attractive as they are embarrassingly 
parallel and exhibit an anytime behavior, meaning that stopping the search at any time yields a valid experiment. 
However, they require the evaluation of 
a relatively large number hyperparameters to find a good solution. In Chapter~\ref{ch:mbm}, we will study
model-based strategies, such as Bayesian optimization. These approaches learn a surrogate 
model of the
response function 
based on the evaluations observed so far. 
Model-based approaches are sample-efficient, which means that they typically require evaluating
a much smaller number of hyperparameters (compared to grid or random search) to find a satisfactory solution. 
However, they are inherently sequential, which can negatively impact the run time of the 
optimization algorithm. In Chapter~\ref{ch:multi_fidelity}, we will discuss how randomized 
approaches can be made resource-efficient by stopping unpromising evaluations early. We will also 
show that they can be married to model-based approaches in order to sample the hyperparameter 
search space in a non-uniform fashion. Chapter~\ref{ch:popM} covers population-based methods, stochastic algorithms 
for global optimization that draw 
inspirations from natural phenomena.
When applied to 
HPO, some techniques in this family, such as population-based training, maintain a set of predictors whose
hyperparameters are updated dynamically, unlike the aforementioned approaches 
where hyperparameters are fixed for the entire training duration. In Chapter~\ref{ch:gradient-based}, we will
discuss how gradient-based approaches can be applied to the optimization of continuous 
hyperparameters through hypergradients, which is more time-efficient than search-based algorithms. Finally, 
we close the monograph by discussing advanced use cases. 
In particular, we study how  to adapt HPO to constrained and multi-objective settings.
Indeed, HPO 
algorithms provide a powerful set of tools to meet competing requirements in applications (e.g., maximizing the 
predictive performance while mitigating the prediction bias) and satisfy practical constraints 
(e.g., maximal prediction latency).
We also discuss connections with with neural architecture search and meta-learning, and touch upon transfer learning of hyperparameters for foundation models.
The two closing chapters are dedicated to an overview of HPO software frameworks (Chapter \ref{ch:hposys}) and emerging topics and open questions in the field (Chapter \ref{ch:open_questions}).

\section{Related Surveys}

The topic of this monograph has been surveyed in some other papers,
briefly by \citet{hutter_beyond_2015} and more extensively in the works of
\citet{bischl:2023:hyperparameter_optimization_foundations} and
\citet{yang:2020:hyperparameter_optimization_machine}. \citet{arlot:2010:survey_crossvalidation_procedures} surveyed the
model selection problem from a complementary statistical perspective.
Specific aspects have been reviewed in other papers, in particular
Bayesian optimization by~\citet{Frazier:08,%
  shahriari:2016:taking_human_out,%
  wang:2023:recent_advances_bayesian}, neural architecture
search by~\citet{ren:2021:comprehensive_survey_neurala,%
  heuillet:2024:efficient_automation_neural}, evolutionary
algorithms by~\citet{li:2023:survey_evolutionary_deep}, and
meta-learning by~\citet{hospedales2020meta}.
HPO is a core topic in automated machine learning (AutoML). The subject has been
covered extensively in the book by \citet{hutter2019automated} and more recently in the survey by \citet{he:2021:automl}.

This monograph aims to provide a broad methodological introduction to the problem settings and principal algorithmic approaches in HPO. 
It is intended to support both practitioners aiming to understand the principles behind the methods they use or consider adopting, and researchers entering the 
field or seeking a reference on specific algorithm classes or aspects of hyperparameter optimization.

\section{Notation}

We denote by $\RSet$ and $\NSet$ the set of real and natural numbers, respectively.
For $K$ a positive integer, we indicate by $[K]=\{1, \dots, K\}$ the subset of the natural numbers from $1$ to $K$, included.
We use calligraphic letters, like $\mathcal{D}$, to indicate sets, and script-style calligraphic, like $\mathscr{D}$ to indicate collections.
We reserve the Greek letter $\lambda$ to denote hyperparameters, $\Lambda$ to indicate hyperparameter spaces, and $f$ to denote the response function.
We use $\mathbb{P}$ to indicate a probability measure.
Whenever confusion may arise, we further add a subscript like $\mathbb{P}_{x\sim\nu}$ to specify that the random variable $x$ evaluated by $\mathbb{P}$ follow the distribution $\nu$.
We use $\mathbb{E}$ with the same convention to indicate the expected value.
We mostly reserve the letter $p$ to indicate density or probability mass functions, and denote by $\mathcal{P}(\mathcal{X})$ the space of densities over the set $\mathcal{X}$.
We denote by $\mathcal{N}(\mu, \Sigma)$ the multivariate Gaussian distribution with mean $\mu$ and covariance matrix $\Sigma$.


\chapter{Hyperparameter Optimization}
\label{ch:hpo}

Here, we formalize 
hyperparameter optimization (HPO) along the concepts introduced in the previous chapter.
We discuss the nested structure that characterizes an HPO problem 
and the archetypal settings encountered in supervised machine learning.
To fix ideas, we revisit in greater detail the example of the introduction and
showcase different application scenarios from the literature. 
We conclude the chapter with an overview of 
practical desiderata for HPO algorithms
that will guide our discussions in the remainder of the monograph.

\section{Learning Algorithms with Hyperparameters}

Hyperparameters are variables that configure the behavior of a learning algorithm, shaping and modifying its inductive bias, 
i.e., the set of assumptions the algorithm implicitly relies on in order to generalize.
We view a learning algorithm $A$ as a higher-order, possibly
stochastic, function that maps \textit{data}
and \textit{hyperparameters} to a model $h_\lambda$, capable of accomplishing a
given task:
\begin{equation}
    \label{eq:algodef}
    A: \setofdata \times \Lambda \to \hypothesisspace; \quad h_{\lambda} = A(\Dtrain, \lambda)
\end{equation}
where $\setofdata$ denotes the collection of all finite datasets that the algorithm can process,  $\Lambda$ the \textit{hyperparameter space} (the Cartesian
product of the domains of individual hyperparameters), $\hypothesisspace$ the \textit{model space}
that represents the class of models that the algorithm may output,
and $\lambda\in\Lambda$ a hyperparameter vector (or tuple).
By writing $h_{\lambda}=A(\Dtrain, \lambda)$ in  Eq.~\eqref{eq:algodef} 
we emphasize the dependence of the model returned 
by the algorithm on $\lambda$,
but omit the dependence on the training data $\Dtrain$.
This choice is consistent with the fact that 
one often considers the dataset
``fixed and given'' when setting up an HPO problem.

\subsection{Logistic Regression for Sentiment Analysis: An HPO Perspective}
\label{sec:lrhpo}
As a concrete example, let us revisit in more detail the problem of sentiment analysis introduced in Chapter~\ref{ch1} and, in particular, the logistic regression
model considered by \citet{yogotama2015-emnlp15}.
Data in this case is a set of pairs $\{(x,y)\mid x\in\inputspace, y\in\outputspace\}$
where $\inputspace$ is the space of documents and $\outputspace=\{0, 1\}$  is the space of labels, expressing a negative or a positive sentiment.
The hyperparameter space of the learning algorithm is the
Cartesian product $\Lambda=\Lambdacomp{1}\times\dots\times\Lambdacomp{7}$ of the seven domains reported in the column ``Values'' of
Table~\ref{tab: sentiment analysis hyperparams}.
The first four rows are related to a function $\Phi_{\lambda}:\inputspace\to\RSet^{n_\lambda}$, the feature extractor, that maps documents to $n_{\lambda}$-dimensional vector representation.
The last three rows are related to  regularization
and optimization.
The model space is defined as follows:
\begin{equation}
    \hypothesisspace = \left\lbrace 
       h_{\lambda}:\inputspace\to \outputspace \mid  h_{\lambda}(x) = \ONE{w^\intercal \Phi_{\lambda}(x)>0}, \; w \in \RSet^{n_{\lambda}}, \; \lambda\in\Lambda
    \right\rbrace,
    \label{eq:logreg_model_sapace}
\end{equation}
where $\ONE{s}$ is the indicator function: $\ONE{s}=1$ if $s$ is true and $0$ otherwise.
In other words, $\hypothesisspace$ is the union of the spaces of all the $n_{\lambda}$-dimensional linear classifiers composed with feature extractor $\Phi_{\lambda}$, as $\lambdacomp{1}$, $\lambdacomp{2}$, $\lambdacomp{3}$, and  $\lambdacomp{4}$ vary in their respective domains $\Lambdacomp{1}$, $\Lambdacomp{2}$, $\Lambdacomp{3}$ and $\Lambdacomp{4}$.

In the context of a logistic regression model, learning consists in finding the parameter vector $w$ that minimizes the negative log-likelihood of the data under this model.
The first
four hyperparameters determine in which space this minimization 
is carried out.
For instance, the larger the range of the $n$-grams considered, the more input features and, consequently, parameters the model will have. 
This range is controlled by the hyperparameters $\lambdacomp{1}$ and $\lambdacomp{2}$.
Unigrams alone ($\lambdacomp{1}=\lambdacomp{2}=1$) are very likely not sufficient for sentiment analysis as the sentences
$x_1=$ \texttt{the restaurant is not busy but the food is tasty} and
$x_2=$ \texttt{the restaurant is busy but the food is not tasty} are indistinguishable since $\phi_{\lambda}(x_1) = \phi_{\lambda}(x_2)$.
However, using longer $n$-grams typically requires increasingly larger corpora to avoid overfitting, since the number of possible features (sequences of words of length $n$) grows exponentially with $n$.

\begin{figure}
    \centering
    \includegraphics[width=0.9\textwidth]{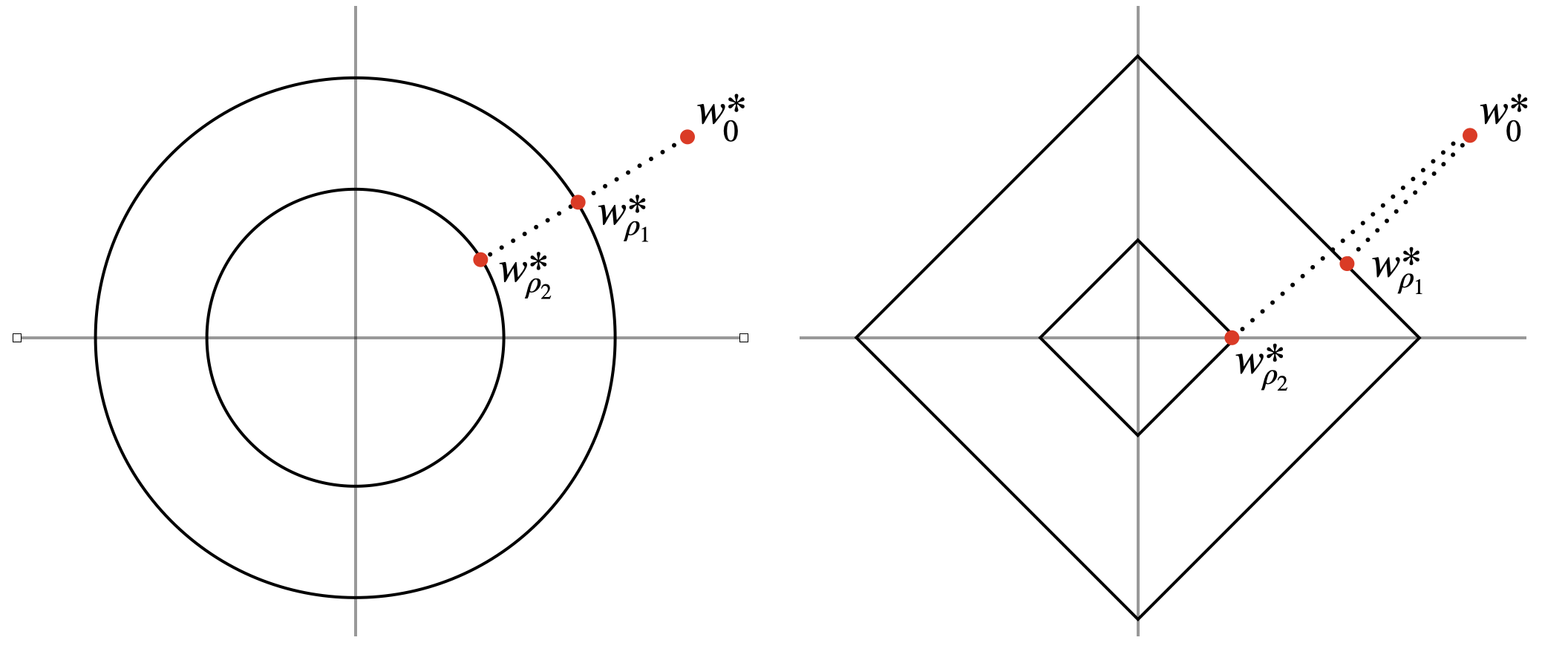}
    \caption{Geometrical representations of the $L^2$ (left) and $L^1$ (right) regularizers. 
    $w^*_0$ represents the minimizer of the training loss $\objtrain_{\lambda}$ when the regularization coefficient $\lambdacomp{6}$ is $0$ (i.e. no regularization). $w^*_{\rho_1}$ and $w^*_{\rho_2}$ are, instead, the minimizers of the regularized losses for two different values of $\lambdacomp{6}=\rho_i$ for $i\in \{1, 2\}$, with $0 < \rho_1<\rho_2$.
    These correspond to the projections of $w^*_0$ onto $L^2$ and $L^1$ balls with varying radii that are inversely proportional to $\lambdacomp{6}$.
  }
    \label{fig:reg}
\end{figure}

The remaining hyperparameters, $\lambdacomp{5}, \lambdacomp{6}$, and $\lambdacomp{7}$, do not directly affect the model space, but rather determine the way in which the optimization is carried out.
Specifically,
the algorithm $A$
outputs the model whose parameter vector $w$ minimizes the training objective $\objtrain_{\lambda}:\RSet^{n_{\lambda}}\times\setofdata\to \RSet$ defined as follows:
\begin{equation}
\label{eq:logreg}
    \objtrain_{\lambda}(w, \Dtrain) = 
    \frac{1}{|\Dtrain|}\sum_{(x,y)\in \Dtrain} -\log \mathbb{P}\left(y|\Phi_{\lambda}(x); w\right) + 
    \lambdacomp{6} \Omega_{\lambdacomp{5}}(w),
\end{equation}
where $\Omega_{\lambdacomp{5}}(w)$ is either the $L^1$ or the $L^2$ norm depending on the value of $\lambdacomp{5}$, and the probability that the model assigns a document to express a positive sentiment, encoded by $y=1$, is given by
\begin{equation*}
    \mathbb{P}\left(1|\Phi_{\lambda}(x); w\right) = \frac{1}{1 + e^{-w^\intercal \Phi_{\lambda}(x)}}.
\end{equation*}
Hyperparameter $\lambdacomp{5}$ exposes a binary choice between two different forms of regularization: the first favors sparse solutions, the second penalizes components of $w$ with comparatively large values.
Figure \ref{fig:reg} offers a geometric interpretation of these regularization schemes. 
The hyperparameter $\lambdacomp{6}$ controls the strength of such regularization; if, for instance, $\lambdacomp{5}=L^1$, one can expect the number of $0$'s in the solution vector to be proportional to the value of $\lambdacomp{6}$.
Finally, $\lambdacomp{7}$ controls the convergence tolerance of the
numerical solver used to
minimize the objective in Eq.~\eqref{eq:logreg}.

\paragraph{}
Although logistic regression is a relatively simple algorithm, its
practical application already exposes several hyperparameters.  Other
algorithms may require many more hyperparameters. This happens for
example in deep learning, where feature extraction is internalized in
the underlying models, and the optimization problems become more
challenging.

The actual hyperparameter space depends on the application and how the problem is framed.  For instance, the range of the regularization coefficient $\lambdacomp{6}$
in our example could in principle take any
non-negative real value: it does not need to be limited to the
interval $[10^{-5}, 10^5]$. Furthermore, one may want to use both types
of regularization simultaneously in Eq.~\eqref{eq:logreg}, instead of picking only one of the two, or one may consider the
usage of also tetra-grams when extracting features.  What to include
in $\Lambda$ is a design choice that can have important implications in practice.
One of the reasons
\citet{yogotama2015-emnlp15} succeeded to
achieve strong predictive performance with a
simple linear classifier is that they included in $\Lambda$ hyperparameters related to the feature extractor $\Phi_{\lambda}$, whereas other authors opted for an \textit{a priori fixed} map $\Phi$ instead of optimizing over it.

Hence, by exposing a (relatively) larger number of hyperparameters of the learning algorithm, such as parameters associated with the feature extractors or parameters associated with the optimizer or the regularizer, one can substantially expand the algorithm's model space $\hypothesisspace$.
These additional degrees of freedom may in turn raise the odds of finding a better model for the task at hand, at the price of having to search over a possibly much larger hyperparameter space.

\section{The Hyperparameter Optimization Problem}
\label{sec:the_HPO_problem}

Hyperparameter configurations need to be evaluated according to a given performance metric, that we denote by $\valmetric$.
This metric takes as input a trained model and a \textit{validation set},
and returns a scalar value, such as the validation error\footnote{In Section \ref{sec:multi-objective-hpo} we discuss the case where $\valmetric$ returns multiple scalars.}.
The objective of an HPO problem, the \textit{response function}, is then constructed from the performance metric. 
Specifically, given a learning algorithm $\learningalgo:\setofdata\times \Lambda\to
  \hypothesisspace$, a metric $\valmetric:\hypothesisspace\times\setofdata\to\RSet$, a training set $\Dtrain$ and a validation set $\Dval$,
  the hyperparameter
  optimization problem is defined as follows:
\begin{eqnarray}
\label{eq:hpo0}
    \min_{\lambda\in\Lambda} f(\lambda)
    &=& \valmetric\left(h_{\lambda}, \Dval \right)\\
    \label{eq:hpo1}
    \mbox{such that} \quad \hypothesis_{\lambda} &=& A(\Dtrain,\lambda).
\end{eqnarray}

The \textit{nested structure} of this problem involves minimizing a function whose dependence from its decision variables (the hyperparameters) is implicit through the execution of the learning algorithm.
Thus, in principle, to evaluate $f$ at a given point $\lambda$, we must run 
$A(\Dtrain, \lambda)$
until termination (e.g. convergence).
 Hence, evaluating the response function can be expensive, as many learning algorithms may take hours, or even days to terminate. We will discuss how we can relax this requirement using low-fidelity approximations in Chapter \ref{ch:multi_fidelity}.
The second major difficulty inherent to most HPO problems is the irregularity of the search space $\Lambda$.
For instance, in our sentiment analysis example, the range of
$n$-grams considered by the feature extractor
$\Phi_{\lambda}$ is a discrete variable, ruling out the application of out-of-the-box continuous
algorithms for optimizing $f$.

When $A$ internally minimizes a training objective, such as the
function $\objtrain_{\lambda}$ in the case of logistic regression,
Problem \eqref{eq:hpo0}--\eqref{eq:hpo1} becomes an instance of
\textit{bilevel programming}~\citep{colson2007overview,dempe2002foundations,dempe2020bilevel} where the algorithm plays
the role of the inner problem \citep{franceschi2018bilevel}.
We will return on this view in Chapter \ref{ch:gradient-based} where we discuss gradient-based HPO techniques that explicitly leverage this structure.
A key observation is that we cannot collapse the optimization problem into one level, where we jointly minimize both model parameters and hyperparameters.
For instance, if we were to optimize the objective $\objtrain_{\lambda}$ from Eq.~\eqref{eq:logreg} jointly with respect to both $w$ and $\lambda$, we would obtain as solutions the smallest values in $\Lambda$, namely $\lambdacomp{6}=10^{-5}$ and $\lambdacomp{7}=10^{-5}$, irrespective of the training set considered. This would defeat the purpose of these two hyperparameters which are limiting the model
complexity.

\subsection{Estimating Generalization}
\label{sec:generalization}

Most of the material and techniques for HPO we
covered in this
monograph may be used regardless of the specific nature and meaning of $\valmetric$ and, hence, of $f$.
However, since many hyperparameters control the inductive bias 
of their respective learning algorithms, 
the archetypal setting
consists in
a response function that estimates 
the generalization of the learning algorithm's output models.
\begin{figure}
    \centering
    \includegraphics[width=0.33\textwidth]{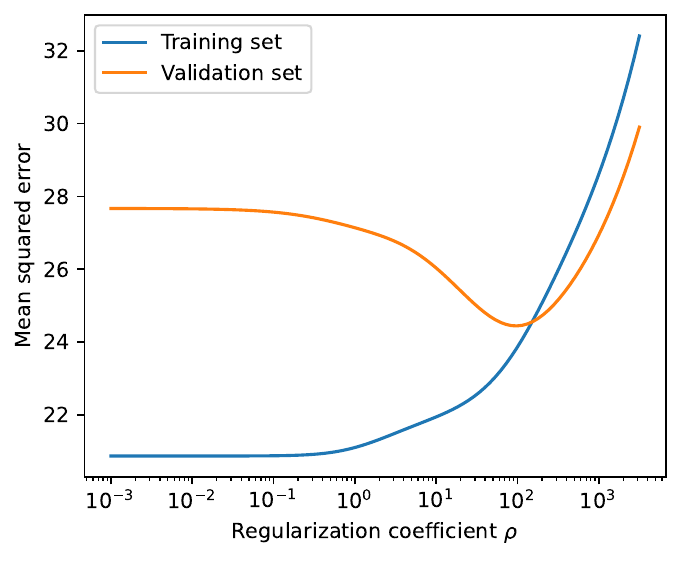}
    \includegraphics[width=0.33\textwidth]{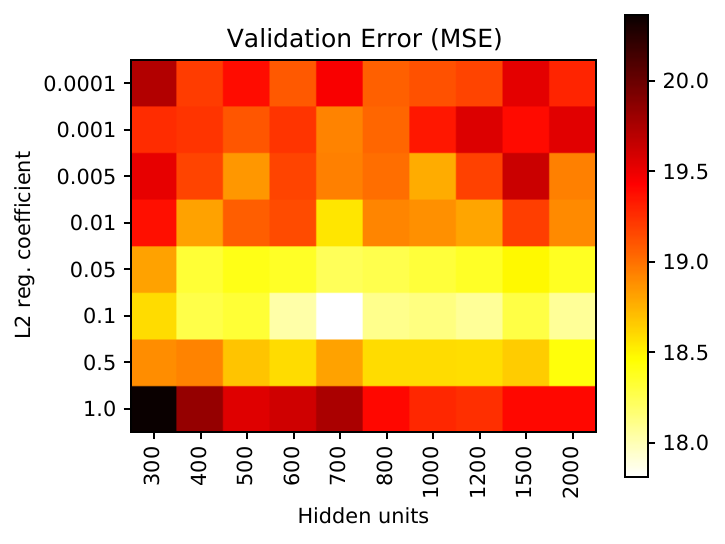}
    \includegraphics[width=0.325\textwidth]{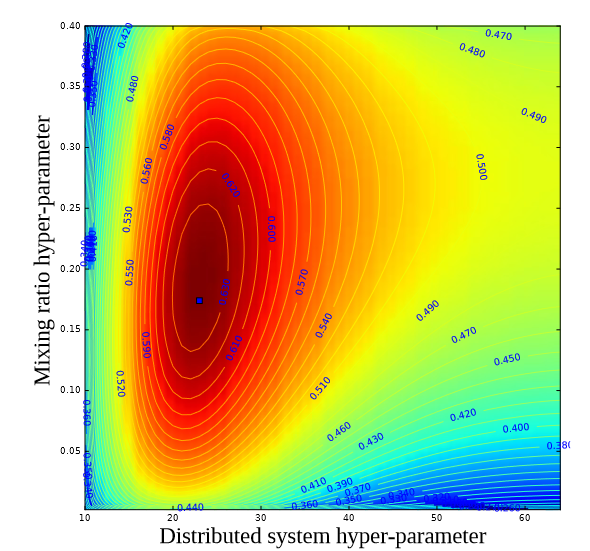}
    \caption{
    Three examples of response functions. From left to right the underlying algorithms are: ridge regression, two-layers neural network regression, Monte Carlo tree search \citep{chen_learning_2017}. 
    }
    \label{fig:2:response}
\end{figure}
Informally, a hypothesis generalizes if it ``behaves sensibly'' on
data outside $\Dtrain$.  

Formalizing this requirement satisfactorily may be a daunting task in
certain scenarios (e.g., in generative modeling), but it is easy in the
case of supervised learning, where we assume to have access to a loss function
$\taskloss:\hypothesisspace\times \mathcal{X}\times\mathcal{Y}\to \RSet$.
We can define the generalization error of $h_{\lambda}$
as the expected loss:
\begin{equation}
    \label{eq:2:gen_err}
    \mathcal{E}(h_{\lambda}) \doteq \mathbb{E}_{(x,y)\sim p} \left[\taskloss(h_{\lambda}, (x,y)) \right],
\end{equation}
where $p$ is the unknown distribution from which the training dataset
$\Dtrain$ was drawn independently. Widely used metrics such as
classification error or mean-squared-error for regression can be interpreted
as empirical estimates of Eq.~\eqref{eq:2:gen_err}:
\begin{equation}
\label{eq:2:valerr}
  \valmetric(h_{\lambda}, \Dval) =
   \frac{1}{|\Dval|}\sum_{(x,y)\in\Dval} \taskloss(h_{\lambda}, (x,y)) ,
\end{equation}
where $\Dval$ is a set of points drawn i.i.d. from $p$, dubbed
\textit{validation}, \textit{held-out} or \textit{development} set. For
example, the sentiment analysis problem addressed by
\citet{yogotama2015-emnlp15} is an instance of supervised binary
classification and the task loss is the $0$-$1$ error, namely
$\taskloss(h_{\lambda}, (x, y))=\ONE{h_{\lambda}(x)\neq y}$. Note
that in many cases, the internal structure of the learning algorithm,
$A(\mathcal{D},\lambda)$, also requires the minimization of a loss function
$\mathcal{J}$ with respect to the model's \textit{parameters}, which may or
may not coincide with $\taskloss$.
For example, in the case of classification, the
0-1 loss is intractable to minimize even for linear classifiers, and we
usually resort to a convex
surrogate such as the
cross-entropy~\citep{nguyen:2009:surrogate_loss_functions}.

Figure \ref{fig:2:response} shows some examples of other response functions.
The leftmost plot concerns a simple ridge regression algorithm to regress median housing prices using the Boston dataset \citep{harrison1978hedonic}.
The scalar hyperparameter controls the $L^2$ regularization and $\mathcal{L}(h_\lambda, (x, y))=\|h_\lambda(x) - y\|^2$. 
The plot shows the mean loss computed on the training and validation data, respectively. 
The center plot represents the response function of a two-layer neural network on the same learning scenario. 
Here the two hyperparameters are an $L^2$ regularization coefficient (real-valued, vertical axis) and the number of units in the hidden layer (integer, horizontal axis). 
The rightmost plot, reproduced from the paper by \citet{chen:2018:bayesian_optimization_alphago}, represents instead the winning probability of a Go agent introduced in Chapter \ref{ch1}, and somewhat deviates from the standard setting we just discussed.

When estimating generalization of a model, it is \textit{essential} to compute $\valmetric$ on a set of points that differs from $\Dtrain$.
In fact, as learning algorithms adapt their hypotheses through 
$\Dtrain$, evaluating $\valmetric(h_{\lambda}, \Dtrain)$ would result in a distorted estimation of Eq.~\eqref{eq:2:gen_err}; e.g., underestimation when the model is under-regularized.
This is illustrated in Figure \ref{fig:2:response}, left.
In order to obtain an unbiased estimate of the expected loss Eq.~\eqref{eq:2:gen_err} one needs to draw points from the task distribution $p$ \textit{independently} both from each other and from $h_{\lambda}$.
Otherwise, one carries over unwanted dependencies between training and validation.

In practice, one may split the entirety of the available data into two disjoint sets $\Dtrain$ and $\Dval$ and use the first as input to $A$ and hold out the second for evaluating $f$. \footnote{
Note that a third \textit{test set} is needed to estimate generalization capabilities of the hypothesis obtained after HPO!
}
More sophisticated approaches, such as \textit{cross-validation}, involve splitting the data into $K$ subsets and define the response function as a mean over $K$ evaluation of $\valmetric$, whereby at each round one subset is held out for validation and the rest of the data is used for training \citep{stone1974cross, kohavi:1995:study_cross}.
Cross-validation may reduce the variance of the estimate of Eq.~\eqref{eq:2:gen_err} at the expense of increased computational cost.

\subsection{Other Types of Response Functions}

One may also be interested in optimizing hyperparameters with respect to performance measures not necessarily linked to generalization.
For instance,  $\valmetric$ could 
measure the 
inference latency or the memory requirements of $h_{\lambda}$ when looking to deploy models for devices on the edge \citep{benmeziane2021comprehensive}.
\citet{chen:2018:bayesian_optimization_alphago} take $\valmetric$ to be an estimate of the winning rate of the AlphaGo agent that plays with hyperparameters $\lambda$ 
versus another AlphaGo agent that plays with a fixed baseline hyperparameter configuration $\lambda_0$.
Figure \ref{fig:2:response} (right) plots the response function (i.e. the winning rate) along two hyperparameter axes.

Furthermore, one may be interested in optimizing the hyperparameters of a learning algorithm simultaneously with respect to multiple measures (e.g., accuracy, inference latency and memory footprint) or in ensuring that $h_{\lambda}$
satisfies some constraints regarding one or multiple measures. 
These scenarios give rise to \textit{multi-objective} and \textit{constrained} hyperparameter optimization,  advanced topics which we will discuss in Chapter \ref{ch:adt}.

\subsection{General Algorithm Configuration}
Several algorithms, not only those employed in machine learning, require
configuration constants that may affect both the quality of the solutions
and the computational resources used to find solutions on specific
instances~\citep{schede:2022:survey_methods_automated}. While in HPO the
emphasis is on the former aspect, in general algorithm configuration (AC)
the focus is more typically on compute. A typical problem suitable for AC is
propositional satisfiability (SAT), where proper choices of parameters can
improve solving time
significantly~\citep{hutter:2009:paramils_automatic_algorithm,hutter_sequential_2011}.
Additional interesting examples can be found in the area of discrete
optimization, where several parameters may be used to configure
metaheuristics~\citep{Birattari2009TuningM}. Entering into details would be
out of scope for this survey. Still, we would like to point out a striking
similarity between AC and HPO. In both cases, optimal (hyper) parameters
depend on some unknown probability distribution, 
such as the data
distribution for HPO, or the distribution of future instances of a certain
problem in AC.
This requires resorting to data, such as a validation set or
a set of problem instances, to formulate a proxy empirical optimization
problem.

\section{Hyperparameter Optimization in the Literature}
\label{sec:examples}

Below we report a few notable milestones and examples of ``success stories'' of HPO in the literature.

\begin{itemize}
  \item The paper by \citet{bergstra_random_2012}, later extended by
        \citet{bergstra_making_2013}, brought the attention of the community
        to random search as a sensible baseline compared to manual or grid
        search which, at the time of writing, were widespread approaches to
        HPO. We cover grid, random and more advanced quasi-random search
        techniques in Chapter \ref{ch:ea}.
  \item \citet{snoek_practical_2012} used Gaussian-process based Bayesian
        optimization to optimize the hyperparameters of a 3-layer
        convolutional neural network~\citep{krizhevsky-nips12} on CIFAR-10.
        They reduced test error by $3\%$ compared to the previous
        state-of-the-art that was achieved using manual search. These
        papers sparked interest in Bayesian optimization for HPO.
        Chapter~\ref{ch:mbm} is devoted to model-based HPO and Bayesian
        optimization in the HPO context.
  \item \citet{jamieson2016non} and later \citet{li2017hyperband} introduced
        bandit-based approaches to hyperparameter optimization. These
        methods draw inspiration from the multi-armed bandit literature,
        particularly the successive halving algorithm \citep{karnin-icml13}.
        They adaptively allocate resources to promising configurations while
        terminating unpromising ones early, emphasizing the concepts of
        budget and efficient compute allocation. These works have been
        particularly influential in the development of multi-fidelity
        optimization methods, which we discuss in Chapter
        \ref{ch:multi_fidelity}.
  \item \citet{zoph2016neural} parameterized the architecture of a
        convolutional and recurrent neural networks to discover new
        architectures. To sample new architectures they trained a recurrent
        neural network controller based on reinforcement learning. This work
        marked the beginning of a sub-field of research dubbed neural
        architecture search (NAS), which we discuss in Section
        \ref{sec:nas}.
  \item As we mentioned in the introduction,
        \citet{chen:2018:bayesian_optimization_alphago} used
        Gaussian-process-based Bayesian optimization to optimize the
        hyperparameter of the Monte-Carlo Tree-Search of Alpha GO. This
        improved the win-rate of Alpha GO from $50\%$ to $66.5\%$ in
        self-play games.
\end{itemize}

\section{Practical Considerations for HPO Algorithms}
\label{sec:hpo-practical}

The general formulation of Eq.~\eqref{eq:hpo0} abstracts away specific aspects that
are practically important and that collectively set aside HPO from a generic
bilevel optimization problem. For this reason, the appraisal of an HPO
algorithm should take into account several dimensions, as summarized in
Table~\ref{tab:hpo_algo_desiderata}.

\begin{table}[ht]
    \centering
    \caption{List of desirable properties of an HPO algorithm.}
    \label{tab:hpo_algo_desiderata}
    \begin{tabular}{p{3.5cm}|p{6.9cm}}
        \toprule
         \textbf{\textit{Characteristic}} & \textbf{\textit{Brief Description} }  \\
         \midrule
         {Sample efficiency}
         & How many evaluations of $f$ does the algorithm need to find satisfactory configurations?
         \\
         {Runtime efficiency}
         & 
        How long does the algorithm take to find satisfactory configurations?
         \\
         {Parallelizability:}
         & 
         How efficiently does the algorithm exploit parallel computational resources if:
         \\
         {~~~| Synchronous}
         &
             \textbullet~~all parallel evaluations need to start at the same time?
         \\
         {~~~| Asynchronous}
         & 
         \textbullet~~evaluations can be started any time a resource becomes available?
         \\
         {Multi-Fidelity Support}
         & 
         Can the algorithm take advantage of cheaper approximate evaluations of $f$?
         \\
         {Applicability}
         & Does the algorithm have restriction on the type of search space $\Lambda$?
         \\
         {Accessibility}
         &  Which type of access does the method need to the underlying learning algorithm and/or training data?
         \\
         {Ease of use}
         & How difficult is to implement and tune the algorithm? What expertise is required to use it effectively?
         \\ 
      \bottomrule
    \end{tabular}
\end{table}

\paragraph{Efficiency and budget.} 
The response function $f(\lambda)$ is typically stochastic, depending on random effects in the computation of $A(\Dtrain,\lambda)$. This may be due for example to weight initialization or to data ordering when performing stochastic gradient descent. It is also computationally demanding to evaluate. Moreover, HPO is typically only a step in a wider iterative process of model development and refinement.
This means that in practice, the emphasis in HPO is to achieve a {\em sufficient decrease} of $f(\lambda)$ compared to a reference point (e.g., previously used default configuration) {\em within a given budget} $b$ --- expressed in time, number of function evaluations, monetary costs, etc. ---
rather than achieving an (approximate) optimum, which is the typical aim in other branches of optimization.
Such a decrease needs to be traded off with compute and runtime costs. A good HPO algorithm should be {\em sample efficient}, such that it significantly decreases $f$ with as few evaluations of $f$ as possible. More importantly, it should be {\em wall-clock time efficient}, meaning that it finds good solutions as rapidly as possible. While number of evaluations and wall-clock time are related,
they are not
equivalent. For example, decision-making (i.e., choosing the next configuration to evaluate) counts towards wall-clock time and needs to be balanced against evaluation times of $f$ for model-based HPO algorithms. When comparing different algorithms, time efficiency is captured by
plotting the best value of $f$ 
attained against wall-clock time and computing the area under this curve. Some HPO methods tend to witness rapid initial descent of $f$, but may level off at a suboptimal value, while others spend more of their initial budget in order to ultimately achieve near-optimal performance.

\paragraph{Multi-fidelity.} A powerful principle to increase time efficiency is to consider low-fidelity approximations of the response, which are cheaper to compute. For example,
if we train a neural network for 64 epochs, we can access for free to $b$ checkpoints, for all $b\leq 64$.
Multi-fidelity HPO methods operate on such a sequence of fidelities $\ffid{\lambda}{b}$, where, in this case,  $f(\lambda) = \ffid{\lambda}{64}$. By controlling the budget $b$ for each trial evaluation $\lambda$ (e.g., stopping training after $b$ epochs), they can often decrease $f(\lambda)$ with a much smaller number of full-fidelity evaluations.
These methods will be discussed in detail in Chapter~\ref{ch:multi_fidelity}.
One aspect that is worth noting when choosing an HPO algorithm is if it can leverage low-fidelity evaluation, and how the quality of such evaluation may impact the final result.

\paragraph{Synchronous and asynchronous parallelization.}
We can trade off
time efficiency against computation cost by running several
evaluations of $f$ in parallel on (say) $W$ workers. For a {\em
  synchronous} system, all $W$ workers start new jobs at the same
time. While this reduces wall-clock time, it often leads to idle time
when workers need to wait for the slowest job to finish. In an {\em
  asynchronous} system, each worker can be assigned a new job
independently of the status of the others. However, HPO
decision-making for parallel execution systems comes with additional
challenges. In the synchronous case, a batch of $W$ new configurations
has to be suggested in one go. Configurations in a batch should be
``diverse'', in that they reveal non-redundant information about the
optimum \citep{Ginsbourger:10, Desautels:14, Gonzales:16,
  Wilson2018}. While only a single new configuration needs to be
proposed in the asynchronous case, up to $W-1$ responses of running
evaluations are not yet available \citep{snoek_practical_2012}. Not
taking such pending evaluations into account runs the risk of
increased redundancy in the proposed configurations. HPO can also be
distributed across several computational nodes that may reside in
networked computers. In this case, algorithms should also be
parsimonious in terms of communication to reduce overhead as much as
possible.

\paragraph{Applicability and accessibility.}
The applicability and accessibility of an HPO algorithm refer to the requirements it may have on the type of problem or the level of access needed to the underlying learning algorithm.
Some algorithms may be restricted to certain types of search spaces, 
e.g., continuous vs. integer or categorical domains, or a mix of these.
Furthermore, some algorithms can exploit conditionality in the search space, while others cannot.
Some HPO methods like gradient-based techniques (Chapter \ref{ch:gradient-based}) presuppose a white-box access to the underlying learning algorithm in terms of computational structure and/or data processing, while others assume the learning algorithm to be a black box.
In between, there are a class of algorithms that require a gray-box access, like the ability to start and stop the execution of the learning algorithm and be able to observe intermediate statistics. 
Although HPO has been traditionally described and tackled as a black-box optimization problem, more restrictive assumptions may be met in specific cases and HPO algorithm that exploit these assumptions typically show efficiency gains.

\paragraph{Ease of use.}

This aspect refers to the complexity involved in implementing, configuring, and effectively using an HPO algorithm.
Some methods may be conceptually simple but require careful tuning of their hyperparameters or auxiliary components. 
Others may be more complex to implement from scratch or require deep expertise in the underlying techniques (e.g., Bayesian and multi-fidelity optimization) to use them effectively.

\section{Why not Tackling the Search Manually?}
\label{sec:ea-manual}

In the next chapters we will introduce the main algorithms for HPO.
One might question the need for sophisticated algorithms instead of just performing manual search.
Indeed, searching for the best hyperparameters by hand can be alluring:
getting started requires no new frameworks or assumptions,
and one may be under the impression that the search could proceed organically, driven by hunches.  

Yet, manual search can quickly become exhausting.
Experienced practitioners may intuitively narrow the search, but novices face a dizzying array of choices.
While flexible, tweaking hyperparameters without a systematic approach risks getting lost in an endless maze of configurations.
Furthermore, with each trial, insights and analysis of results can guide exploration.
While potentially beneficial to speed up the process, this risks overfitting by inadvertently exploiting patterns in the validation data.
Even for experts, it may be difficult to make valid assumptions about the
behavior of a learning algorithm, and it is not always obvious how to carry
over expertize from one application domain to another (or from a class of
models to another).
Finally, manual search also suffers from poor reproducibility, as it is hard
to track the steps a user followed to reach some final configuration.


\chapter{Elementary Algorithms}
\label{ch:ea}

In this chapter, we present simple non-adaptive, model-free and ``memoryless'' approaches to HPO, in that subsequent runs are not informed by any previous search history.
These methods correspond to basic search schemes in global optimization that are conceptually simple, widely applicable, (mostly) easy to implement and straightforward to parallelize.
Furthermore, these methods may work as initialization schemes for more advanced techniques like Bayesian optimization, as we shall see in later chapters.

We start with grid search, an approach that consists in dividing the search space into a grid of values and querying the response function on each of these values.
Next, random search swaps the grid in favor of a fixed, simple distribution over the search space, drawing configurations from such distribution until a budget is exhausted. 
Finally, we move to quasi-random search strategies, approaches rooted in low-discrepancy sequence generation, in which the simple distributions of random search are replaced by more complex schemes that seek to explore the space more evenly.
While grid search is a traditional and classic technique in HPO, 
consensus has grown in the field that random search should be considered as the baseline method for HPO \citep{bergstra_random_2012}. 
Quasi-random search may offer benefits over random search, although to date we are not aware of any theoretical result supporting this.

\section{Grid Search}

Perhaps the first formal description of grid search (GS) dates back 
to the 1930s, appearing in a foundational work by
\citet{fisher1935design} on experimental design, under the name of \emph{factorial design}.
GS consists in evaluating $f(\lambda)$ over a predefined grid of
hyperparameter configurations.
For a generic hyperparameter $\lambdacomp{i}$ with domain $\Lambdacomp{i}$, we
define the discretized domain $\barLambdacomp{i}$ as a \textit{finite}
subset of $\Lambdacomp{i}$. The search grid is the Cartesian product
$$
\bar{\Lambda} = \barLambdacomp{1} \times \barLambdacomp{2} \times \cdots \times \barLambdacomp{m} \subset \Lambda.
$$
The learning algorithm is then executed sequentially or in parallel on the resulting
set of tuples and the returned solution is
\begin{equation}
    \hat{\lambda} \in \argmin_{\lambda \in \bar{\Lambda}} f(\lambda).
\end{equation}

GS is a simple memoryless method that
can be used to explore low-dimensional hyperparameter spaces.
It requires very little implementation effort and parallelization is straightforward.
Unfortunately, if $d$ is the cardinality of the largest discretized domain, then
$|\bar{\Lambda}|=O(d^{m})$, making the method impractical for even moderate
values of $m$.
In addition, while Boolean and categorical hyperparameters naturally lend themselves to an evaluation on a predefined grid, unbounded integer and real-valued hyperparameters require manually setting
bounds and discretization. 

Extensive experimental practice has shown that some continuous hyperparameters, such as the learning rate, benefit from discretization on a logarithmic scale \citep{bengio2012practical}.
While requiring less expert knowledge than manual search, GS 
is very sensitive to 
the definition of the grid. 
To reduce the computational overhead, GS  
can be paired with manual search: the user alternates GS steps with adjustments and refinement 
of the grid values, e.g., using a finer grid \textquote{centered} around the best found configuration, or moving the search bounds of continuous hyperparameters, until a satisfactory result is found.  

\begin{figure}
	\centering
	\includegraphics[width=0.87\textwidth]{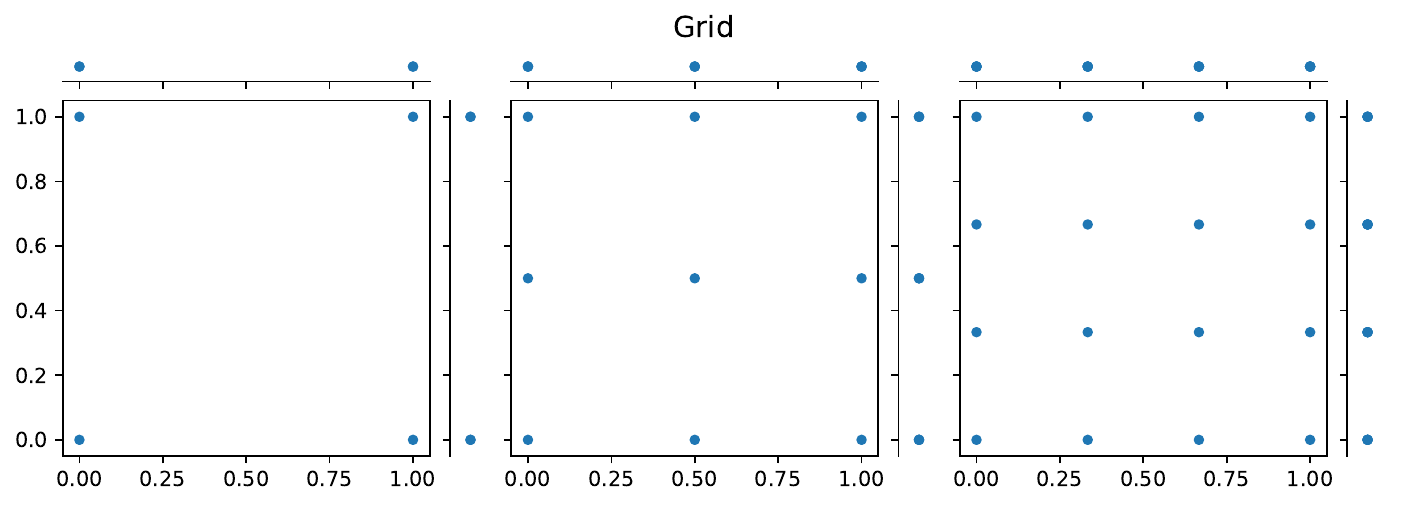}
	\includegraphics[width=0.87\textwidth]{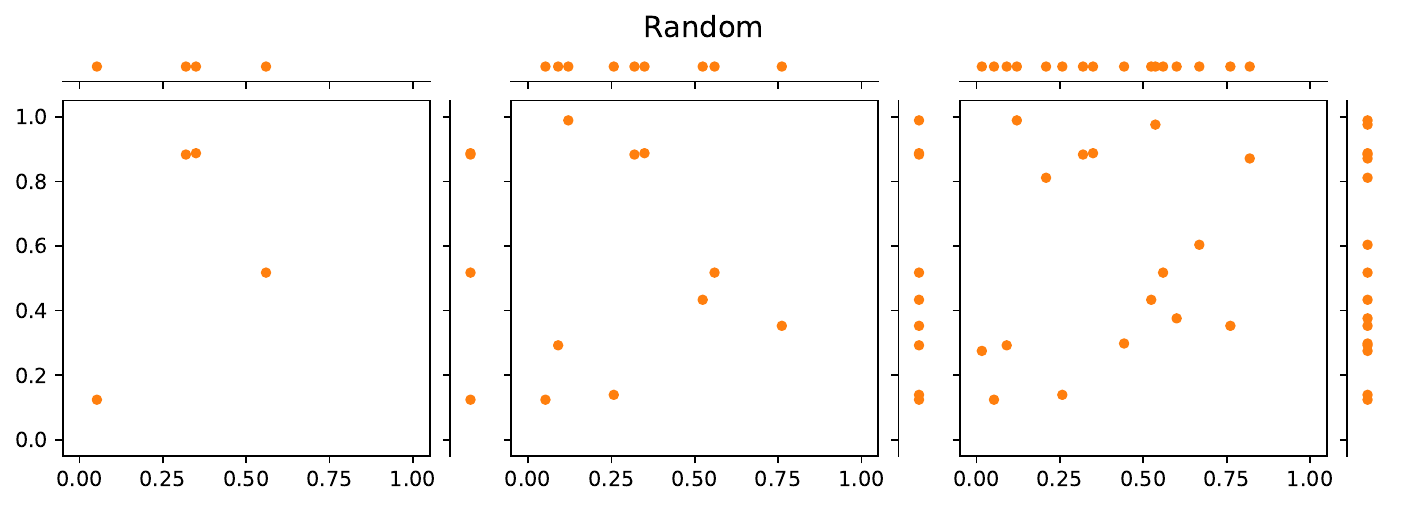}
	\includegraphics[width=0.87\textwidth]{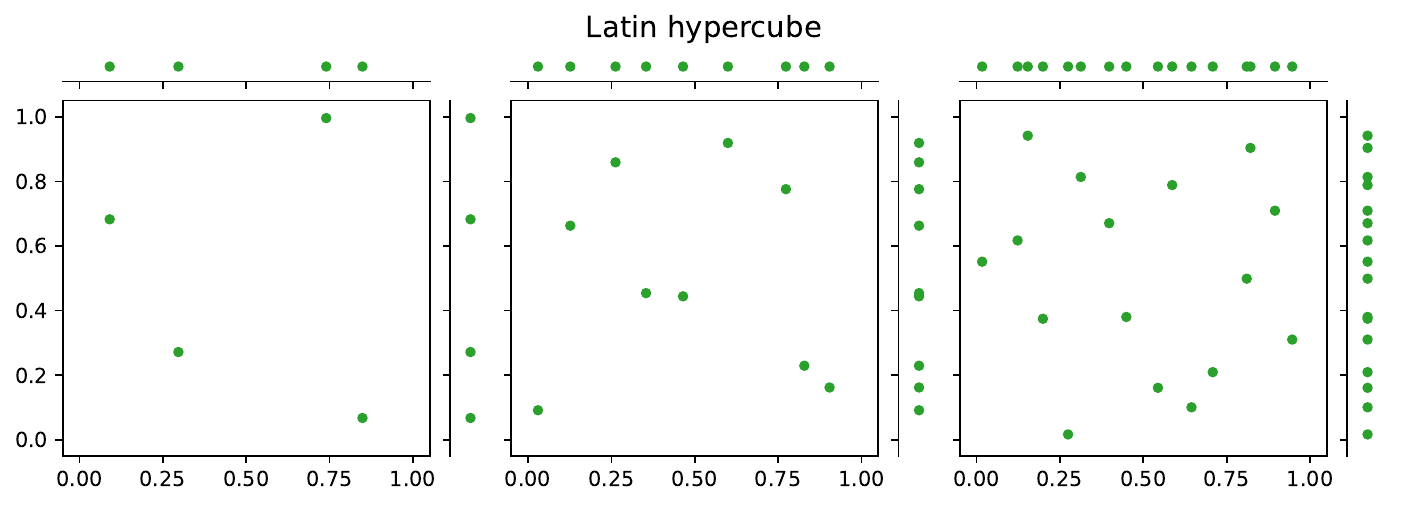}
	\includegraphics[width=0.87\textwidth]{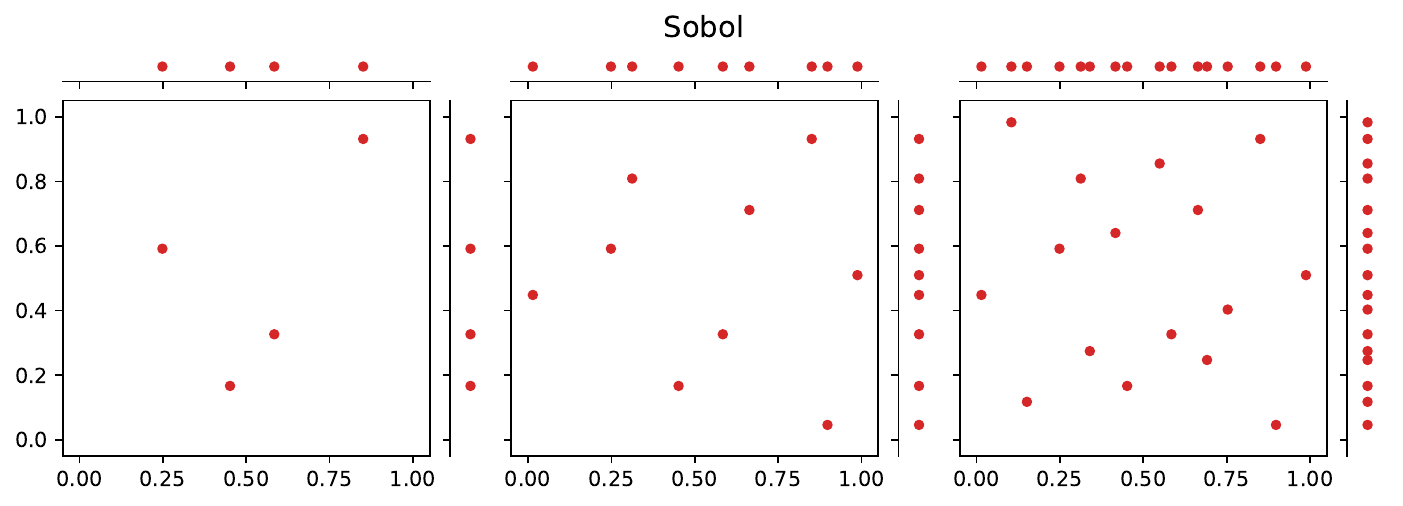} 
	\caption{
		Non-adaptive sampling methods on the unit square for different sample sizes: 4 (left), 9 (center) and 16 (right) points. 
		Each row represents a different schema. Grid and Latin hypercube sampling (LHS) require re-sampling every time we increase the sample size. Instead, for random and Sobol' we can keep on adding points, ``reusing'' previous samples. For each plot, we also show projections of the samples onto the two axes. 
		We can see how grid samples cover much fewer points on the projections w.r.t. the other sampling methods. 
		We can also observe the ``unlucky run'' phenomenon of (uniform) random sampling, where after nine draws the upper-right quadrant is still unvisited. LHS and Sobol' (with random shifts and scrambling) provide more even coverage of the space and its unidimensional projections.}
	\label{fig:enter-label}
\end{figure}

The top row of Figure \ref{fig:enter-label} depicts equally distanced grid ``samples'' with 4, 9 and 16 points, on the unit square.

\section{Random Search}
\label{sec:random_search}
Random search (RS) was
already considered in the 1950s as a method for global optimization \citep{brooks1958discussion}.
It is perhaps the conceptually simplest procedure for HPO.
RS consists in repeatedly drawing hyperparameter configurations from a predefined distribution $\hpoap$ over $\Lambda$, and it is readily parallelized.
The solution after $T$ trials is
given by
\begin{equation}
    \hat{\lambda}(T) = \argmin_{\lambda\in\{\lambda_{1}, \dots \lambda_{T}\}} f(\lambda),
\end{equation}
where the trial points $\lambda_{1},\dots,\lambda_{T}$ are drawn i.i.d. from $\hpoap\in\mathcal{P}(\Lambda)$.
While in GS the total number of trial points is predetermined, one may keep on executing RS indefinitely. 
Termination conditions might involve approximate convergence checks, 
like lack of improvements  after a certain amount of trials, or after a given budget is exhausted (see Section \ref{sec:hpo-practical}).
For instance, one may run RS for 12 hours. 

In principle, $\hpoap$ could be any distribution. However, when referring to random search for hyperparameter optimization, one typically implies the use of simple distributions (e.g., uniform or log-uniform distributions) that factorize independently along the configuration components, i.e.,
$\hpoap= \prod_{i=1}^m\hpoapcomp{i}$.
For instance, if $\lambdacomp{i}$ is a categorical choice between $N$ different
neural architectures, then $\hpoapcomp{i}$ could be a categorical uniform
distribution. Similarly, if $\lambdacomp{i}$ is the learning rate for stochastic gradient descent, then $\hpoapcomp{i}$ could be the log-uniform (or reciprocal) distribution on the interval $(10^{-5},10^{-1})$.

Using a distribution rather than a grid allows for a more natural treatment of a larger class of hyperparameters: RS does not require discretization of continuous spaces and also allows for distributions with non-compact supports.

The second row of Figure \ref{fig:enter-label} shows uniform random samples with 4, 9 and 16 points on the unit square.

\subsection{On the efficiency of random search}
In an influential article,~\citet{bergstra_random_2012} advocate for using RS as the \textquote{default} baseline method for HPO instead of manual or grid search.
They argue that while RS retains the advantages of GS, such as reproducibility, conceptual simplicity, very low implementation overhead and ease of parallelization, it is also more efficient in higher dimensional spaces, especially when the response function has a \emph{low effective dimensionality}.
The authors informally define this condition as when a subset of 
the hyperparameters 
accounts for most of the variation of the response function. 
For instance, for a bi-dimensional hyperparameter $\lambda=(\lambdacomp{1}, \lambdacomp{2})$ it could be that $f(\lambda)\approx g(\lambdacomp{1})$, for some $g:\Lambdacomp{1} \to \RSet$.

One way to reason about the success of RS is to assume that there is a
sub-region $\Lambda^* \subset\Lambda$ that leads to satisfactory performance
(i.e., where $f$ is sufficiently small). Then, we can assess the probability that
RS returns a solution in this sub-region~\citep{brooks1958discussion},
assuming that hyperparameter assignments are drawn i.i.d. from $\hpoap$:
\begin{equation}
\label{eq:bk:prob_success_rs}
\mathbb{P}\left[\hat{\lambda}(T)\in\Lambda^*\right] = 1 - \left(1 -
\mathbb{P}[\lambda\in \Lambda^{*}] \right)^T.
\end{equation}
Eq.~\eqref{eq:bk:prob_success_rs} says that the minimum number of trials needed to assure a satisfactory outcome with a given probability, say $95\%$, depends only on the probability of hitting $\Lambda^{*}$ under $\hpoap$.
This can be independent of the search space dimension if the response function has a low effective dimensionality.
This argument may explain the success of RS as a baseline in seemingly complex settings such as neural architecture search~\citep{li2019random}, where
repeated experimentation on a few well-known benchmark datasets has led to the development of well-tuned search spaces for neural architectures. 

For a general search space,
we should expect $\mathbb{P}[\lambda\in \Lambda^{*}]$ to decay exponentially in some $m'\leq m$.
For concreteness, consider the case where $\Lambda=[0, 1]^m$ is the $m$-dimensional unit hypercube and $\Lambda^*=[0, 1/2]^{m'} \times [0, 1]^{m - m'}$ is the ``good region''.
The $m'$-dimensional 1/2-hypercube represents the important hyperparameters, while the $m - m'$-dimensional unit hypercube represents the unimportant ones, as these coordinates do not matter in hitting $\Lambda^*$.
If $\hpoap$ is the uniform distribution over $\Lambda$, then
$\mathbb{P}[\lambda\in \Lambda^{*}]=2^{-m'}$.
Consequently, the number of trials needed to obtain a positive outcome with at least $95\%$ probability is 
$$
T\geq \log(0.05) / \log(1 - 2^{-m'}).
$$
For $m'=3$ one already needs $T\geq 23$ draws  to achieve a $95\%$ probability of success.
This illustrates that the number of trials needed grows exponentially in $m'$.
At the same time, this also shows that RS is much more efficient compared to
GS if $m'\ll m$.

\subsection{Unlucky runs}
One potential disadvantage of RS, due to its randomized nature, is that entire regions of $\Lambda$ could be under-explored or completely neglected.
 As a consequence, a single execution of random search may not necessarily provide sufficient information to meaningfully modify $\hpoap$ for another iteration of random search, should it be needed.
 
 Let us illustrate this phenomenon. 
 Continuing with our illustrative examples, say we divide the unit hypercube in $2^m$ 1/2 hypercubes $I_j$, for $j\in[2^m]$.
 We wish to have at least one configuration per hypercube (i.e., anywhere inside the 1/2 hypercube). 
 While GS would need exactly $2^m$ points to achieve this, 
 the probability under uniform sampling that RS hits all the hypercubes in $2^m$ draws is
 $$
 \prod_{j=1}^{2^m - 1} (1 - j/2^m).
 $$
 For example, for $m=3$ such probability is already $3/32$,.
 This means that, in expectation, RS with $2^3=8$ draws covers all the hypercubes only once every ten runs.

\section{Quasi-Random Search}

One basic requirement for a  stochastic generator of hyperparameter configuration is \textit{consistency}: we would like that $\lim_{T\to \infty} \min_{k\in[T]} f(\lambda_{k}) = f(\lambda^*)$.
RS with a uniform distribution is consistent, while GS is not. 
However, unlike GS, random search for any fixed budget may generate a set of configurations that misses large portions of the search space, as we just discussed.
Intuitively, good sequences should cover evenly the search space. 
Also, their projections onto subspaces of $\Lambda$ should be well spread, in order to speed up convergence when optimizing functions with low effective dimensionality --- a property that
GS misses. 
Quasi-random search methods seek to combine the advantages of grid and random search, with configuration-drawing strategies that combine randomness with deterministic steps.

Formalizations of these desiderata have been developed in the context of numerical integration and global optimization via the concepts of low dispersion and low discrepancy sequences \citep[see e.g.][]{sobol1979systematic, niederreiter1988low, dick2010digital}. 
\citet{bousquet2017critical} 
study the usage of such quasi-random sequences in HPO, finding that they consistently outperform random search (with a uniform distribution) on a series of deep learning tasks; see also \citep[][Chapter 7]{tuningplaybookgithub}.
The resulting methods are conceptually very similar to both GS and RS, being highly parallelizable, except that the set of hyperparameters $\lambda_{1}, \dots, \lambda_{T}$ are generated according to more complex procedures.
We review two possible choices below, assuming that the search space is an $m$-dimensional unit hypercube for simplicity. 
Note that this is typically not at all a hard limitation:  starting with samples on the unit interval, one can easily derive samples for a variety of distributions with appropriate reparameterizations 
\citep[see, e.g.][]{devroye2006nonuniform}.

\subsection{Latin hypercube sampling (LHS)}
\label{sec:latin}
LHS \citep{mckay1979comparison} is a stratified sampling scheme that directly combines grid and random search, in that it divides the search space into a grid, but samples configurations at random within regions of the grid.

LHS works as follows: choose a sequence length $n>1$, which determines the number of hyperparameter configurations generated by a run of the sampling process.
Divide each $[0,1]$ interval of the unit hypercube into $n>1$ parts, denoting by $I_s=[(s-1)/n, s/n]$ the resulting intervals. 
This gives rise to $n^m$ hypercubes of volume $n^{-m}$ each.
Then generate $m$ random permutations of
$[n]$, denoted by $\{\sigma_i\}_{i=1}^m$, $\sigma_i:[n]\to[n]$.
For $k\in [n]$, sample
hyperparameter configurations as follows:
\begin{equation}
    \lambda_{k} \sim \mathcal{U}(H_k), \;\text{ where } \;
    H_k = I_{\sigma_1(k)}\times I_{\sigma_2(k)} \times \cdots \times I_{\sigma_m(k)},
\end{equation}
where $\mathcal{U}(H_k)$ is the uniform distribution on $H_k$.
This generates $n$ trial points; one may then generate further sequences of $n$ configurations by repeating the process.
Importantly, Latin hypercube sampling ensures that the projection on each axis of the $n$-long sequences places exactly one sample per interval. 

The second last row of Figure \ref{fig:enter-label} shows Latin hypercube samples on the unit square with 4, 9 and 16 points; namely, for $n\in\{2, 3, 4\}$.

\subsection{Sobol' sequences} 
Sobol' sequences \citep{sobol1979systematic} are a collection of separate one-dimensional sequences, one for each axis: when constructing an $m+m'$ dimensional sequence one can reuse the $m$-dimensional one for the first coordinates.
Furthermore, unlike LHS, Sobol' sequences are ``extendible'', meaning that if $\{\lambda_1, \dots, \lambda_T\}$ is a Sobol' sequence, one can add new points ${\lambda_{T+1}, \dots, \lambda_{T+T'}}$ such that $\{\lambda_1, \dots, \lambda_{T+T'}\}$ remains a Sobol sequence. 
This is, in practice, a useful property shared with random search, as it allows one to avoid committing to a predefined budget $T$, offering flexibility to add more points as needed, without invalidating  previous runs.

Let $j\in[m]$ be an axis, and let $\mathbb{F}_2 = \{0, 1\}$ with integer operations modulo $2$. 
For each integer $k$, representing the iterate, let $k=\sum_{l\geq 0} a_l(k) 2^l$ be the binary expansion of $k$, where $a_l(k)\in \mathbb{F}_2$ are binary coefficients (e.g., for $k=3$, these would be $a_0(3) = 1$, $a_1(3) = 1$ and the rest $0$'s).
Denote by $a(k)$ the resulting vector.
Let $C_j$ be a specific binary (infinite) ``direction matrix'' for axis $j$. 
Then the $j$-th coordinate of the $k$-th iterate of a Sobol' sequence can be written as
\begin{equation*}
    [\lambda_{k}]_j = \sum_l \frac{[C_j \cdot a(k-1)]_l}{2^{l+1}}
\end{equation*}
where the matrix-vector multiplication $C^j \cdot a(k)$ is in $\mathbb{F}_2$ (i.e., the additions are modulo 2).
The direction matrices $C^j$ are constructed starting from primitive polynomials of $\mathbb{F}_2$ and direction numbers, which must be carefully chosen to ensure good coverage of the sequences. We refer the reader to~\citet[][Chapter 5]{Lemieux:1338331} for details.

Note that, given the directions matrices, Sobol' sequences are deterministic. 
It is common practice to obtain stochastic variants of such sequences. This is achieved by first adding random shifts, i.e. replacing $\lambda_{k}$ with $\lambda_{k} + u$ (modulo the unit hypercube), where $u$ is drawn uniformly from the interval $[0, 1]$, and by then randomly permuting the entries of the binary representation vector $a(k)$ \citep{owen1995randomly}.
The bottom row of Figure~\ref{fig:enter-label} shows 4, 9, and 16 points randomized Sobol' sequences on the unit square.

\subsection{Other schemes}
\citet{bousquet2017critical} experimented also with Halton and Hammersley sequences and found that the latter (with random permutations and shifts) performed well on the tested deep learning scenarios.
Like Sobol' sequences, both Halton and Hammersley sequences can be extended.
\citet{Lemieux:1338331} presents other quasi-random low-discrepancy sequences.
LHS, Sobol', Halton and Hammerley sampling are implemented in the \texttt{stats.qmc} module of SciPy.


\chapter{Model-Based Methods}

\label{ch:mbm}

The techniques described in the previous chapter do not attempt to model the
response function and are memoryless, i.e., the outcomes of previous
evaluations do not influence the search. In contrast, model-based techniques
use data gathered from past evaluations to
build an explicit approximation of the objective function, known as a
\textit{surrogate model}, which is the central component of the family of HPO methods described in this chapter.

In general, the surrogate model takes as input a candidate configuration and, optionally, additional contextual information, such as the performance of partially trained models or the performance of the current best configuration (the incumbent). 
It then produces a prediction about how well the configuration is expected to perform.
For example, the surrogate model might predict a distribution over performance scores.
Details on the functional form of the surrogate depend on the specific instantiation of the
model-based HPO method and will be discussed in the next sections.
Regardless of the exact implementation, a key requirement of surrogate models is that they must be much cheaper to evaluate than the original objective function.

The landscape of model-based (MB) methods is strongly dominated by techniques falling under the umbrella of \emph{Bayesian optimization} (BO), which we cover in Section \ref{sec:bk:hpo:bayesopt}.
In BO, the surrogate model is a probabilistic mapping that also captures the uncertainty of the estimation of the response function.
Common implementations include Gaussian processes (Section \ref{sec:gp}) and tree Parzen estimators (Section \ref{sec:tpe}). 
BO methods do not select the next configuration by directly minimizing the surrogate model.
Rather, they use a second mapping, called
\textit{acquisition function}, which mediates the selection process.
Typically, the acquisition function aims to balance the selection of the next configuration between two competing objectives:
\begin{itemize}
\item  \textit{Exploration}, which prioritizes evaluating the response function in regions where the surrogate model is highly uncertain, with the goal of gathering information to improve the surrogate model itself;
\item \textit{Exploitation}, which focuses on selecting configurations that the surrogate model predicts to perform well, aiming to improve the current best solution.
\end{itemize}
We examine various acquisition functions and their roles in Section \ref{sec:acq_func}, and  present a practical example that contrasts BO to random search in Section \ref{sec:mbhpo-example}.
In Section \ref{sec:other-surrogates}, we briefly touch upon other approaches to model-based optimization that do not directly fall under the BO category.
We conclude the chapter with an overall discussion of model-based methods in Section \ref{sec:mbhpo:disc}.
In particular, we discuss the parallelizability of model-based methods --- a significant practical bottleneck of MB-HPO methods --- in Section \ref{sec:parallel-mbhpo}.

\section{Model-Based HPO: A General Schema}
\label{sec:mb:general}

\begin{algorithm}
\caption{\texttt{Model-based-HPO}}
\label{alg:mbhpo}
\begin{algorithmic}[1]
\Require{Response function $f$}
\Require{Warm-up dataset $\mathcal{C}_0 = \{ (\lambda_j, f(\lambda_j)) \}_{j=1}^{n_{0}}$}
\Require{Class of surrogate functions $\hat{\mathcal{F}}$ and fitting routine}
\Require{Acquisition function $\alpha$}
\State $\beta \gets \min_{j} f(\lambda_j)$ \Comment{Initialize incumbent performance}
\State Fit $\hat{f}_0$ to $\mathcal{C}_0$ \Comment{Initialize the surrogate model}
\For{$k \gets 1$ \textbf{to} $T$}
    \State $\lambda_k \gets \argmax_{\lambda'} \alpha(\lambda',\hat{f}_{k-1}, \beta)$ \Comment{Select next hyperparameter}
    \State Evaluate $f(\lambda_k)$ \Comment{Expensive step}
    \If{$f(\lambda_k) \leq \beta$}
        \State $\beta \gets f(\lambda_k)$  \Comment{Update incumbent performance}
    \EndIf
    \State $\mathcal{C}_{k} \gets \mathcal{C}_{k-1}\cup \{(\lambda_k,f(\lambda_k))\}$
    \State Fit $\hat{f}_k$ to $\mathcal{C}_k$  \Comment{Update the surrogate model}
\EndFor
\end{algorithmic}
\end{algorithm}

The vast majority of model-based methods are, in principle, sequential. 
At each iteration, 
the surrogate model is used to select the next hyperparameter configuration to evaluate,
 typically through the optimization of an acquisition function.
Then, the response function is evaluated on the selected configuration by running the learning algorithm and assessing the resulting model performance.
Finally, the surrogate model is updated with the newly collected data, or refitted on all data collected so far, and the process is repeated.

Algorithm~\ref{alg:mbhpo} summarizes the typical steps of an MB-HPO method,
where the selection of the next configuration is guided by the optimization of an acquisition function.
The surrogate model is typically initialized by fitting it to several evaluations obtained during a warm-up stage,
where configurations are generated using methods such as random search (see Section~\ref{sec:random_search}).
We denote this initial configuration set as $\mathcal{C}_0 = \{ (\lambda_j, f(\lambda_j))\}$.
New configurations are selected and evaluated, then added to the set of points $\mathcal{C}_k = \mathcal{C}_0 \cup \{(\lambda_i, f(\lambda_i))\}_{i=1}^k$, where $k\in[T]$  denotes the iteration index and $T\in\NSet^+$ is the total number of iterations.
This set of points is then used to fit the $k$-th version of $\surrogm$, which we denote by $\surrogm_k$.
We denote the acquisition function as:
\begin{equation}
  \alpha: \Lambda \times \hat{\mathcal{F}} \times \mathcal{K} \to \RSet.
\end{equation}
Here, $\hat{\mathcal{F}}$ denotes the class of surrogate functions and
$\mathcal{K}$ optional auxiliary information that may be needed by the acquisition function,
including the validation performance of the incumbent --- denoted by $\beta $ in the algorithm, so that $\mathcal{K} = \RSet$.

One may interpret the interplay between fitting a surrogate model and choosing 
the next point to evaluate through an acquisition function 
as an iterative refinement of a sampling distribution $\nu_k \in \mathcal{P}(\Lambda)$ 
--- implicitly defined through $\surrogm_k$ and $\alpha$.
In comparison, random and quasi-random search methods discussed 
in the previous chapter maintain  a predetermined  distribution $\nu\in \mathcal{P}(\Lambda)$ that 
does not change through iterations (cf. Section \ref{sec:random_search}).

In the next section, we focus on model-based methods rooted in Bayesian optimization. 
We discuss specific implementations of surrogate models and their update strategies, 
as well as various acquisition functions.

\section{Bayesian Optimization}
\label{sec:bk:hpo:bayesopt}

BO has long been used as a method to globally optimize black-box functions,  that is,  functions for which we may only observe the value $f(\lambda)$ at any point $\lambda\in\Lambda$.
It dates back at least to the 1970s with a series of works by J. Mo{\v c}kus \citep{mockus:1975:bayesian_methods_seeking, mockus1978application}.
Although traditionally suited for low- to medium-dimensional problems, BO has proven 
to be a powerful method for optimizing expensive, non-convex, and highly multi-modal functions.
As a result, it has attracted considerable attention for its potential to tackle HPO problems, especially following the publication of three influential papers in the early 2010s \citep{bergstra2011algorithms, hutter_sequential_2011, snoek_practical_2012}.
In this section, we give an overview of the general methodology in the HPO context. 
We refer the reader to the work by \citet{jones1998efficient} for a more general perspective of the methodology and to the review by \citet{shahriari:2016:taking_human_out} for further insights into practical aspects and applications to machine learning beyond HPO.

In Bayesian optimization, the surrogate model $\surrogm$ is a probabilistic model. 
Specifically, for any configuration $\lambda\in\Lambda$, the surrogate model $\surrogm(\lambda)$ is a random
variable whose distribution captures the uncertainty about the possible values of the true objective function $f(\lambda)$.
This uncertainty has an \textit{epistemic} nature, i.e.,
due to limited number of  observations of configurations-responses pairs.
The surrogate is updated using Bayesian inference, conditioning on the collected data $\mathcal{C}_k$, to obtain a posterior distribution over functions $\surrogm_k$.
This posterior is then used to guide the selection of the next configuration via an acquisition function, as introduced in the previous section.
The acquisition function encodes a strategy to balance exploration: sampling in regions
of high uncertainty to improve the surrogate model, and exploitation: selecting configurations that are expected to yield good performance.
In this context, the surrogate model's ability to represent epistemic uncertainty is crucial for enabling effective exploration.

Next, we discuss standard choices to design and fit surrogate models in BO, including Gaussian processes and tree Parzen estimators. 
We discuss acquisition functions,  focusing on the expected improvement \citep{mockus1978application}.
A classic instantiation of BO consists in modeling the surrogate with a Gaussian process and using expected improvement as the acquisition function.
It is well-suited for the global optimization of smooth multivariate real functions,
and it provides an accessible setting to implement BO for hyperparameter optimization.

\subsection{Bayesian Linear Model}

To introduce our first surrogate model, we start with a simple linear model $\hypothesis_w$ 
in a given feature space $\mathcal{V}\subseteq \RSet^d$, to which we add independent Gaussian noise with variance $\sigma^2$:
\begin{equation}
\label{eq:bk:bo_lin1}
    \hypothesis_w(\lambda) = \varphi(\lambda)^\intercal w + \varepsilon; \qquad \varepsilon\sim\mathcal{N}(0, \sigma_{\varepsilon}^2),  
\end{equation}
where $\varphi:\Lambda\to\mathcal{V}$ is a feature map and $w\in\RSet^d$ is a weight vector.
Using linear maps directly in the input space would result in an overly simplistic model --- we are trying to approximate a response function that can be highly non-convex and multimodal.
 However, mapping configurations into a  suitable feature space allows us to capture much richer behaviors while retaining many of the benefits of the linear formulation \citep[][Ch. 14]{murphy2012machine}.
 As evaluating $f$ is expensive, we expect to work with only a few observations. 
The noise in Eq.~\eqref{eq:bk:bo_lin1} models potential errors (and inherent stochasticity) in the observations (we further discuss this aspect in Section~\ref{sec:disantangling}).
  
Suppose now that we are at the $k$-th iteration of the algorithm, after having collected the observations $\mathcal{C}_k = \{(\lambda_i, f(\lambda_i))\}_{i=1}^k$. \footnote{In the following, we disregard the warm-up set, for simplicity.} 
Let 
$$Y_k=\left(f(\lambda_1), \dots, f(\lambda_k)\right)^\intercal\in\RSet^k$$
be the vector of observed values, and denote by 
$$
X_k=\left(\varphi(\lambda_1), \dots, \varphi(\lambda_k)\right)\in \RSet^{d\times k}
$$
 the corresponding design matrix in the feature space (organized by columns). 
We are interested in obtaining a probability distribution over the output of $f$ at a query point $\lambda_*\in\Lambda$, that is, in computing
\begin{equation*}
    y_* \sim \surrogm_{k}(\lambda_*) = p(\cdot\mid x_*, X_k, Y_k),
\end{equation*}
where $x_* = \varphi(\lambda_*)$ denotes the feature representation of 
the query point $\lambda_*$.
Here, the surrogate model $\surrogm_k : \Lambda \to \mathcal{P}(\RSet)$ produces 
a predictive distribution over real-valued outcomes, modeling the response function $\lambda$.
The resulting distribution will be employed in Section~\ref{sec:acq_func}, where the acquisition function uses the surrogate $\hat{f}_k$ to guide the selection of the next configuration $\lambda_{k+1}$.

\paragraph{Modeling uncertainty away from observations.}
We could follow an empirical risk minimization approach, obtaining the weights $\bar{w}_k\in\RSet^{d}$ of the model in Eq.~\eqref{eq:bk:bo_lin1} by minimizing, for example, a (regularized) mean squared error over the dataset $(X_k, Y_k)$.
We could then directly set $\surrogm_{k}=\hypothesis_{\bar{w}_k}$.
 Such an estimate would, however, leave us with a rather meager predictive distribution 
$p(\cdot\mid x_*, X_k, Y_k, \bar{w}_k)$,
capable of capturing only our belief of the amount of noise present in the observations. As we discussed above,
we would like the stochasticity of $\surrogm$ to  (primarily) capture our uncertainty about the behavior of $f$ in unexplored regions.

To better understand the difference between these two aspects, following an argument by \citet{jones1998efficient}, consider the case where $f$ is a (non-linear) smooth deterministic function.
If we were to follow an empirical risk minimization (ERM) approach, we could either omit the noise term $\varepsilon$ in Eq.~\eqref{eq:bk:bo_lin1}, or leave it for capturing modeling errors.
However, the first choice would yield a single point estimate for any $\lambda\in\Lambda$, which is very unlikely to be accurate, especially for points far from those previously observed.
 Conversely, if we still wish to maintain a noise component, $\surrogm_{k}$ would trivially be inaccurate at $\lambda_i$ for $i\in [k]$ (since $f$ is deterministic).
 Thus, both ERM modeling options would leave us with a surrogate model that is \textit{almost surely} inaccurate.  

A way to escape this apparent stalemate is to replace the independent Gaussian noise of Eq.~\eqref{eq:bk:bo_lin1} 
 with one that depends on the evaluation point, i.e. letting $\varepsilon:\Lambda \to \RSet$ be a function of $\lambda$.
When fitting the model to the $k$ observations, this choice allows 
having $\varepsilon_k(\lambda_i)=0$ for $i \in [k]$  while still retaining $\varepsilon_k(\lambda)\neq 0$ for $\lambda\neq\lambda_i$. 
Furthermore, since we assumed $f$ to be smooth, $\varepsilon_k$ will also be smooth  (as a difference of smooth functions). 
By continuity, we then have  
$\varepsilon_k(\lambda)\to\varepsilon_k(\lambda_i)=0$ as $\lambda\to\lambda_i$. 
The Bayesian perspective offers a principled way to achieve this behavior, as we discuss next.

\paragraph{Deriving the posterior predictive distribution.} 
We can reason about all linear models in the feature space that are consistent with the 
observations, weighted by their posterior probability. In other words, we construct 
the surrogate model as a weighted average of all plausible $\hypothesis_w$ given the observed data.
Formally, we marginalize over the weights $w$ and set
\begin{equation}
    \label{eq:bk:bo:surrogm}
    \surrogm_{k}(\lambda_*) = \int p(\cdot\mid x_*, X_k, Y_k, w)p(w\mid X_k, Y_k) \, \mathrm{d}w\in\mathcal{P}(\RSet).
\end{equation}
Given our modeling assumption of Eq.~\eqref{eq:bk:bo_lin1} and the hypothesis that the noise is independent, the first distribution simply reads as
\begin{equation}
\label{eq:bk:bo:dist1}
    p(\cdot\mid x_*, X_k, Y_k, w)= p(\cdot\mid x_*, w) = \mathcal{N}(\hypothesis_w(\lambda_*), \sigma_{\varepsilon}^2) = \mathcal{N}(x_*^\intercal w, \sigma_{\varepsilon}^2). 
\end{equation}
By the Bayes' rule, $p(\cdot\mid X_k, Y_k)$ is instead given by 
\begin{equation}
    \label{eq:bk:bo:bayesw}
    p(\cdot\mid X_k, Y_k) = \frac{p_y(\cdot\mid X_k, w)p(\cdot\mid X_k)}{p_y(\cdot\mid X_k)}.
\end{equation}
Regarding the numerator of Eq.~\eqref{eq:bk:bo:bayesw}, we have that, for the same argument of Eq.~\eqref{eq:bk:bo:dist1}, 
\begin{equation*}
    p_y(\cdot\mid X_k, w) = \mathcal{N}(X_k^\intercal w, \sigma_{\varepsilon}^2 I).
\end{equation*}
This is the likelihood of the observations given the model. We must instead make a decision regarding the second term of the numerator, that is the distribution of the model parameters.
We make the very natural assumption  of independence from the design matrix
and set
\begin{equation}
    \label{eq:bk:bo:weight_prior}
    p(\cdot\mid X_k) 
    = p = \mathcal{N}(0, \Sigma),
\end{equation}
where $\Sigma$ is a positive definite covariance matrix. This is a so-called \emph{prior}.  Equipped with Eq.~\eqref{eq:bk:bo:weight_prior}, we can now also compute the denominator of Eq.~\eqref{eq:bk:bo:bayesw} by marginalizing $p_y(\cdot\mid X_k)$ over $w$. We obtain
\begin{equation}
    \label{eq:bk:bo:marginal_likelihood}
    p_y(\cdot\mid X_k) = \int p_y(\cdot\mid X_k, w) p(w)\, \mathrm{d}w = \mathcal{N}(0, X_k^\intercal \Sigma X_k + \sigma_{\varepsilon}^2 I).
\end{equation}

We recognize, now, that the feature map only appears in a quadratic form where the $(i,j)$-th entry is given by $\varphi(\lambda_i)^\intercal \Sigma \,\varphi(\lambda_j)$: a dot product in the feature space. 
We then introduce the kernel corresponding to the feature map $\varphi$ as follows:
\begin{equation}
    \label{eq:bk:bo:kernel_map}
    \kappa_{\Sigma}(\lambda, \lambda') = \varphi(\lambda)^\intercal \Sigma \,\varphi(\lambda') =  \dotprod{\varphi(\lambda)}{\varphi(\lambda')}_{\Sigma}
\end{equation}
and rewrite Eq.~\eqref{eq:bk:bo:marginal_likelihood} in terms of the Gram matrix $K=\{\kappa_{\Sigma}(\lambda_i, \lambda_j)\}_{i,j=1}^k\in\RSet^{k\times k}$ as
\begin{equation*}
    p_y(\cdot\mid X_k) = \mathcal{N}(0, K + \sigma_{\varepsilon}^2 I).
\end{equation*}
Finally, we have all the terms to compute Eq.~\eqref{eq:bk:bo:surrogm}; which is, again, Gaussian:
\begin{multline}
    \label{eq:bk:bo:fin_pred}
    \surrogm_{k}(\lambda_*) = p(\cdot\mid x_*, X_k, Y_k) = 
    \\ 
    \mathcal{N}(k_*^\intercal(K+\sigma_{\varepsilon}^2 I)^{-1}Y_k, \kappa_\Sigma(\lambda_*, \lambda_*) - k_*^\intercal(K+\sigma_{\varepsilon}^2 I)^{-1} k_*)
\end{multline}
where $k_*=\{\kappa_{\Sigma}(\lambda_*, \lambda_i)\}_{i=1}^k\in\RSet^k$ is the vector of kernel evaluations between the query and the observation points.
Eq.~\eqref{eq:bk:bo:fin_pred}, despite its apparent complexity, describes a rich
\emph{posterior} distribution over the values of $f$ that is analytically computable and whose design depends essentially only on the definition of the kernel map in Eq.~\eqref{eq:bk:bo:kernel_map}.

An example of kernel function is the squared exponential kernel: 
$$
\kappa_{\Sigma}(\lambda, \lambda') =\sigma_0 e^{-r_{\Sigma}^2(\lambda, \lambda')/2}; 
\qquad 
r_{\Sigma}^2(\lambda, \lambda') = (\lambda - \lambda')^\intercal \Sigma^{-1} (\lambda - \lambda'),
$$
where $\sigma_0>0$ is a scale parameter and $\Sigma$ a diagonal matrix with positive length scales $\{\sigma_i\}_{i=1}^m$.
Other examples include kernels from Mat{\'e}rn family, which contains mappings with various degrees of smoothness, defined as functions of $r_{\Sigma}$, introduced above  \citep{williams2006gaussian}. 
Kernels intuitively encode the concept of distance or similarity 
between data points, and allow sidestepping the definition of explicit feature maps and covariance matrices. 
The series of \textquote{free parameters} that appears in Eq.~\eqref{eq:bk:bo:fin_pred} ---
that is, $\sigma_{\varepsilon}$, $\sigma_0$ and the diagonal entries of $\Sigma$ when using the squared exponential or a Mat{\'e}rn kernel ---
 are usually called \emph{hyperparameters} in the Bayesian literature, and are not to be confused with $\lambda$.
They do not need to be fixed constants but can instead be estimated from the observed data --- see \textit{automatic relevance determination} \citep{mackay1992bayesian, williams2006gaussian,}.
Thus, the design choice left to the experimenter mainly boils down to the choice of the kernel function. 

\subsection{Gaussian Processes}
\label{sec:gp}

Instead of working with explicit feature maps and weights, 
we can directly define a model through a kernel function, which leads to the 
framework of \emph{Gaussian processes} (GPs) \citep{williams2006gaussian, 
murphy2012machine}. In fact, the Bayesian 
linear model described above converges to a GP in the limit of infinitely many features, 
provided the prior over the weights is scaled appropriately \citep{neal1996bayesian}.

A Gaussian process (GP) is a distribution over functions
whose finite-dimensional marginals are multivariate Gaussian distributions.
 A GP is fully specified by a mean function $\mu: \Lambda \to \RSet$ and a covariance function $\kappa: \Lambda \times \Lambda \to \RSet$, also called the
 kernel. 
 That is,
\[
    g \sim \mathcal{GP}(\mu, \kappa)
\]
means $g$ is a random function such that for any finite set of points $\lambda_1, \dots, \lambda_k \in \Lambda$, the random vector $(g(\lambda_1), \dots, g(\lambda_k))$ follows the multivariate Gaussian distribution:
\[
(g(\lambda_1), \dots, g(\lambda_k)) \sim 
    \mathcal{N}\left(M, \Upsilon\right)
\]
where the mean vector is given by $M_i = \mu(\lambda_i)$ and the 
covariance matrix is given by $\Upsilon_{ij} = \kappa(\lambda_i, \lambda_j)$.

In Gaussian process-based Bayesian optimization, the surrogate model is  a random function 
$
 \surrogm \sim \mathcal{GP}(\mu, \kappa)
$ 
taking values in the space of functions $\Lambda \to \RSet$. 
The update of the Gaussian process after observing $\mathcal{C}_k$ (the fitting step 
in Algorithm~\ref{alg:mbhpo}) corresponds to conditioning the prior 
on the dataset $(X_k, Y_k)$, and can be written as:
\begin{equation}
    \label{eq:bk:bo:pi2}
    \surrogm  \sim \mathcal{GP}(\mu, \kappa \mid  X_k, Y_k) = \mathcal{GP}(\mu_k, \kappa_k),
\end{equation}
where typically $\mu$ is taken to be $0$.
While the posterior GP is technically an infinite-dimensional object, in practice we 
only require its marginal distributions at 
single query points $\lambda_*$ (or small batches of points) --- which are (multivariate) Gaussians --- to compute acquisition functions and guide the search.

Our derivation of the Bayesian linear model in the previous section yields a zero-mean Gaussian process (since the prior over weights, Eq.~\eqref{eq:bk:bo:weight_prior}, has mean zero) with covariance function 
\[\kappa(\lambda, \lambda') =\kappa_{\Sigma}(\lambda, \lambda') + \sigma_{\varepsilon}^2\delta(\lambda, \lambda')
= \varphi(\lambda)^\intercal \Sigma \varphi(\lambda') + \sigma_{\varepsilon}^2\delta(\lambda, \lambda'),
\]
where $\delta(\lambda, \lambda')$ is the Dirac delta function, and the
 term $\sigma_{\varepsilon}^2\delta$ accounts for independent Gaussian observation noise.
More specifically, the model derived in Eq.~\eqref{eq:bk:bo:fin_pred} corresponds to the posterior 
predictive distribution of the GP at a single query point $\lambda_*$,  
conditioned on the observed dataset $(X_k, Y_k)$.
This marginal distribution over $\RSet$ is a univariate Gaussian and 
provides the mean 
\[
\mu_k(\lambda_*) = k_*^\intercal(K+\sigma_{\varepsilon}^2 I)^{-1}Y_k
\] and variance 
\[
\kappa_k(\lambda_*, \lambda_*) = \kappa_\Sigma(\lambda_*, \lambda_*) - k_*^\intercal(K+\sigma_{\varepsilon}^2 I)^{-1} k_*
\] required for the evaluation of acquisition functions.

\paragraph{Limitations of Gaussian processes.}
Gaussian processes also come with a number of drawbacks that may limit their applicability to real-world scenarios.
 Most importantly, computing the posterior for $k$ data points costs $\mathcal{O}(k^3)$, due to the inversion of Eq.~\eqref{eq:bk:bo:fin_pred}.
 If we are able to sample $f$ a few thousand times or more, GP-based BO starts to become intractable. It is also difficult to design kernel functions for high-dimensional $\Lambda$, or if the search space contains many integer, categorical, or conditional hyperparameters. In such situations, BO with other surrogate models can perform better, as shown in a comparative study by \citet{eggensperger_towards_2013}.

\citet{snoek_practical_2012} tackle the first problem, proposing an extension of the expected improvement that accounts for pending executions.
\citet{snoek_scalable_2015} tackles scalability issues by replacing the GP with a Bayesian neural network. \citet{hutter_sequential_2011, bergstra2011algorithms, bergstra_making_2013, perrone2018scalable, jenatton2017bayesian}  focus  on scalability and applicability to discrete and conditional search spaces by using random forests and tree Parzen estimators, which we will discuss in  Section~\ref{sec:mbm-tpe}).
More recent advances in the field \citep{springenberg2016bayesian, wu2017bayesian, klein2017fast} focus on replacing GPs with  more practical and scalable approaches, exploiting additional information besides function evaluations (thus \textquote{dropping} the black-box assumption for $f$).
The additional information could be, for example, the performance of $\learningalgo$ on a small subsets of $D$ \citep{klein2017fast} or (possibly noisy) gradients \citep{wu2017bayesian}.
Finally, \citet{falkner2018bohb} propose hybridizing BO with bandit-based methods (discussed in the next chapter)
 to improve performance in the first stage of HPO, 
 when the surrogate model may still be  unreliable.

\subsection{Acquisition Functions}\label{sec:acq_func}

We now come to the second central component of a Bayesian optimization method: the acquisition function (AF). 
Suppose we have fitted our surrogate model $\surrogm_k$ to the current set of evaluations $\mathcal{C}_k = \{ ( \lambda_j, f(\lambda_j))\}_{j=1}^{k}$. 
To decide which point to evaluate next, BO optimizes an \textit{acquisition function} $
\alpha(\lambda,  \surrogm_k)$:  
\begin{equation}
    \label{eq:bk:bo:pi1}
    \lambda_{k+1} = \arg\max_{\lambda\in\Lambda} \, 
    \alpha(\lambda,  \surrogm_k).
\end{equation}
In this section, we primarily think of the surrogate model as a $\Lambda \to \RSet$-valued 
random function modeled by a Gaussian process, as described above.
However, most AFs discussed here can be applied to other models as well. 
As we will see, most common AFs depend on the posterior distribution of $\surrogm_k$ only through its marginal mean and variance
\[
  \mu_k(\lambda) = \mathbb{E}\left[\surrogm_k(\lambda)\right],\quad \sigma_k(\lambda)^2 = \mathbb{V}\left[ \surrogm_k(\lambda)\right],
\]
given by Eq.~\eqref{eq:bk:bo:fin_pred}, where $\mathbb{E}$ and $\mathbb{V}$ denote expectation and variance, respectively.

A useful AF should guide
the choice of the next candidate
to gain as much new information about high-performing hyperparameters as possible, given the data $\mathcal{C}_k$ collected so far. 
In particular, it needs to balance the trade-off between exploration (i.e.,
looking
away from points in $\mathcal{C}_k$ in order to increase coverage) and exploitation (i.e., refining the search by probing near current optima in $\mathcal{C}_k$). We can gain intuition by considering two extremes. First, we could sample at our current best guess of where the minimum should be:
\[
  \alpha_{\text{exploit}}(\lambda \mid  \surrogm_k) = -\mu_k(\lambda).
\]
Maximizing $\alpha_{\text{exploit}}$ constitutes an extreme form of exploitation, which typically gets stuck in suboptimal local optima. On the other extreme, we could sample where we know least about $f$:
\[
  \alpha_{\text{explore}}(\lambda, \surrogm_k) = \sigma_k(\lambda).
\]
Maximizing $\alpha_{\text{explore}}$ is entirely focused on exploration and may waste many evaluations in regions of poorly performing hyperparameters. However, a linear combination of the two,
\[
  \alpha_{\text{LCB}}(\lambda,   \surrogm_k) = -\mu_k(\lambda) + \eta \sigma_k(\lambda),\quad \eta>0,
\]
is a competitive and frequently used AF, called \emph{lower-confidence bound} (LCB) \citep{Srinivas2009}.
The choice of $\eta$ explicitly determines the trade-off between exploration and exploitation. Choosing $\eta$ depends on the application and, unfortunately, there is no good default rule of thumb to set the value.

\paragraph{Expected improvement.}
A classic AF, introduced in the seminal work by \citet{mockus1978application} and frequently used in HPO libraries \citep{snoek_practical_2012} and services \citep{AMT2020}, is the \emph{expected improvement} (EI), defined as
\begin{equation*}
    \alpha_{\text{EI}}(\lambda,   \surrogm_k, \beta) = \expectation \left[\max\left\lbrace \beta - \surrogm_k(\lambda), 0\right\rbrace\right].
\end{equation*}
Here, $\beta\in\RSet$ is a baseline, which is typically chosen as the
minimum value of $f$ observed so far: $\beta=\min_{k\in [K]} f(\lambda_k)$. 
In
contrast to LCB, EI does not come with a free parameter to be set. 
EI makes an implicitly explore-exploit trade-off that has been reported to result in good performances in practice. A generalized version of EI replaces $0$ by $\eta > 0$, which adjusts this trade-off with a cut-off hyperparameter.
In
addition, it can be shown to have a simple closed-form expression in terms of
$\mu_k(\lambda)$ and $\sigma_k(\lambda)$:
\begin{equation}
  \alpha_{\text{EI}}(\lambda, \surrogm_k,  \beta) = \sigma_k(\lambda)\nu(\lambda)\Phi\left(\nu(\lambda)\right)+\sigma_k(\lambda)\phi\left(\nu(\lambda)\right)
  \label{eq:EI}
\end{equation}
where
$$
 \nu(\lambda) = \frac{\beta - \mu_k(\lambda)}{\sigma_k(\lambda)}
$$
and where $\Phi$ and $\phi$ denote the cumulative distribution function and the
probability density function of the standard Gaussian distribution,
respectively.

\begin{figure}
  \centering
  \begin{subfigure}{0.49\textwidth}
    \includegraphics[width=0.99\textwidth, height=4cm]{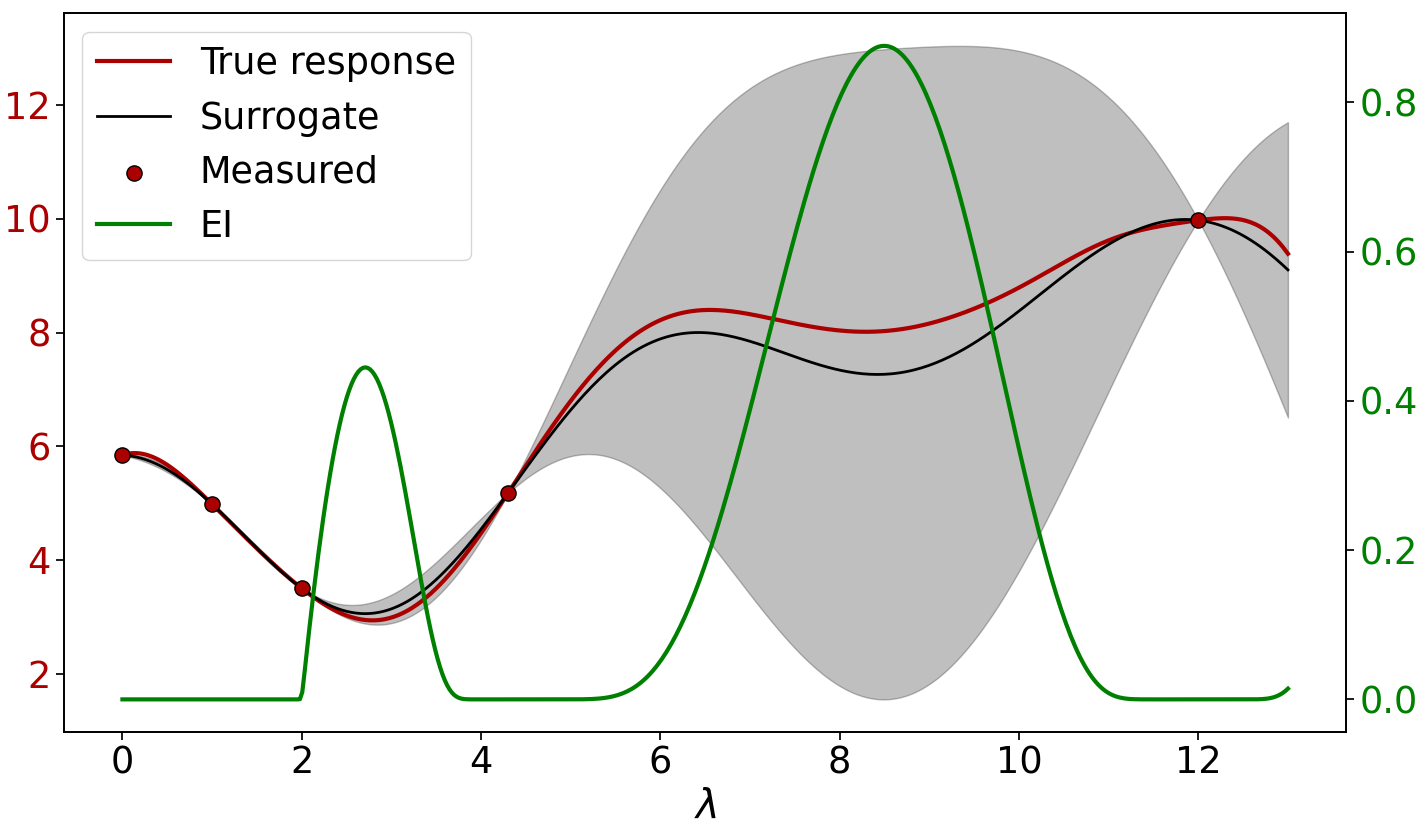}
  \end{subfigure}
  \begin{subfigure}{0.49\textwidth}
    \includegraphics[width=0.99\textwidth, height=4cm]{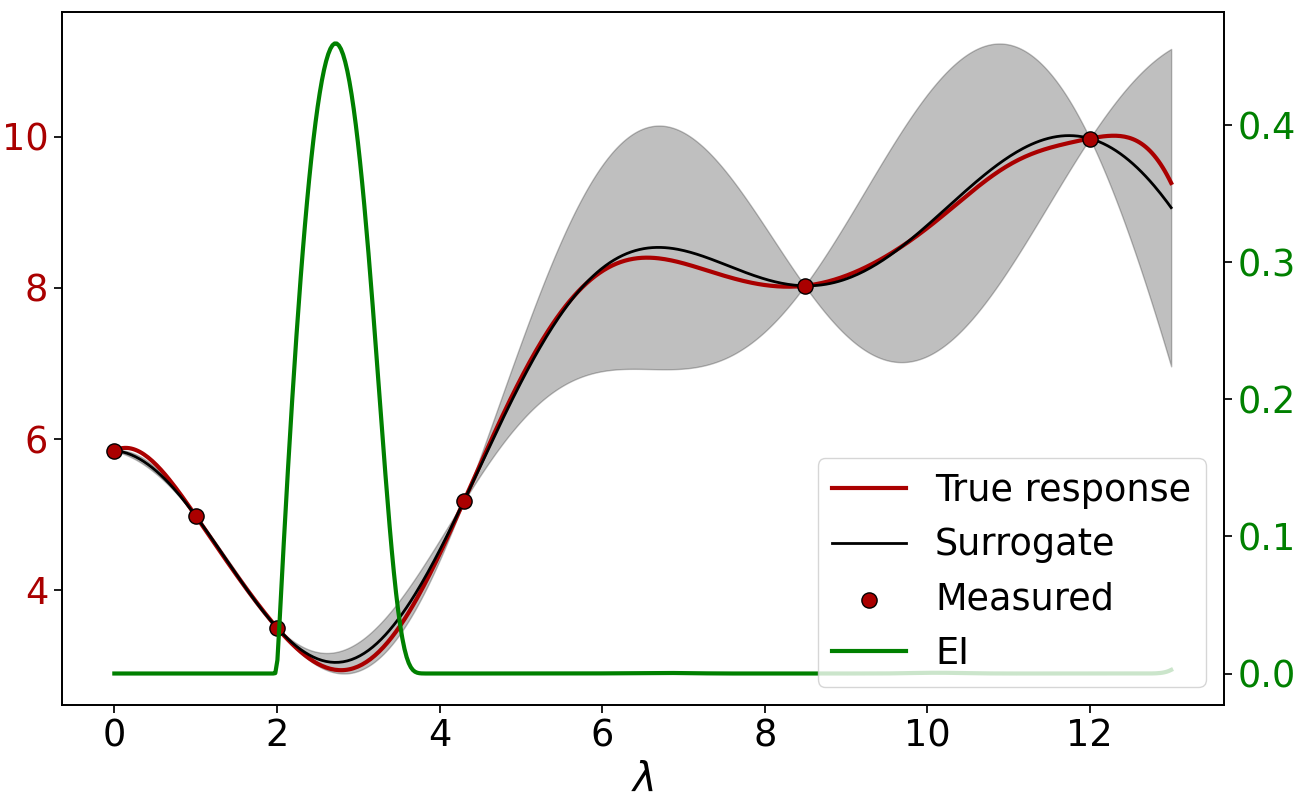}
  \end{subfigure}
  \caption{Left: Suppose we have measured the response function (plotted in red) on a small set of points. Fitting a Gaussian process on these points yields a mean function (plotted in black) with and a variance function ($\pm$ one standard deviation is filled in gray). If $\lambda$ falls between 7 and 11, the uncertainty is very high, which yields the right bump in the expected improvement (plotted in green), allowing us to \textit{explore} a region where potentially a smallest response might be found. The left bump of the expected improvement falls in an \textit{exploitation} region, which in this example looses against exploration. Hence, the next candidate is $\lambda= 8.5$. Right: once the response is measured at the new candidate, the Gaussian process is trained again. Now uncertainty shrinks significantly and the new expected improvement has just one bump around 2.7, which is already very close to the minimum of the true response. This example takes inspiration from Figure 11 in \citep{Jones1998:Efficient}.
  }
  \label{fig:ei_demo}
\end{figure}

Eq.~\eqref{eq:bk:bo:pi1}, where $\alpha$ is the expected improvement, and Eq.~\eqref{eq:bk:bo:pi2} specify a particular simple instantiation of a GP-based BO algorithm that may be used for HPO.
 As an illustrative example, two steps of the Bayesian optimization algorithm using EI as the acquisition function are shown in Figure~\ref{fig:ei_demo}.

\paragraph{Other acquisition functions.}
An earlier and simpler variant of EI is the probability of improvement \citep{Kushner1964:New}, given by
\begin{equation}
  \alpha_{\text{PI}}(\lambda, \surrogm_k, \beta) = \mathbb{P}\left[\surrogm_k(\lambda)<\beta\right] = \Phi(\nu(\lambda)).
  \label{eq:PI}
\end{equation}
Another popular AF is known as \emph{Thompson sampling} (TS) \citep{Thompson1933, Hoffman2014}. The idea is to draw a sample path from our surrogate model and to search for its minimum point. Exact TS is intractable for a GP surrogate model, sample paths of which do not have a finite representation. In practice, we can draw a joint sample over a finite set of points
or approximate the GP posterior by a Bayesian linear model \citep{Hernandez-Lobato2014}.

Other relevant families of AFs rely on information-theoretic criteria \citep{Villemonteix2009,Hennig2012,Hernandez-Lobato2014,wang2017max}.
These typically are more complex to implement, come with higher computational cost, and require biased approximation, 
but have compelling theoretical foundations and often perform better than standard choices such as EI or LCB, 
achieving lower sample complexity.
Given that the choice of the best AF can depend on the target function to optimize, one can also resort to a multi-armed bandit strategies to pick the most suitable AF from a portfolio of choices over the course of the optimization \citep{hoffman2011portfolio, shahriari:2016:taking_human_out}.

\paragraph{Optimizing acquisition functions in practice.}
The optimization of acquisition functions can itself pose a challenge.
As a result, Eq.~\eqref{eq:bk:bo:pi1} is rarely solved exactly.
Techniques for maximizing AFs include grid, random, and quasi-random search,  discussed in the previous chapter. 
Because surrogate models are cheap to evaluate, it is typically feasible to sample the acquisition function at many candidate points. 
This allows even naive search strategies to perform reasonably well in practice.
More sophisticated approaches rely on gradient-based optimization, 
which uses the derivatives of the acquisition function to locate local maxima more efficiently.
This is especially effective when the acquisition function is smooth and differentiable --- such as in the case of GP-based posteriors where gradients can be computed analytically or via automatic differentiation.
However, it is important to note that thorough optimization of the acquisition function is rarely essential.
 This is because the surrogate model is only an approximation of the true objective, and the acquisition function itself is a heuristic derived from it. 
Approximate maximization is typically sufficient to drive effective search progress for the overall HPO algorithm \citep{Wilson2018}.

\subsection{Aleatoric and Epistemic Uncertainty}
\label{sec:disantangling}

The noise term $\varepsilon$ introduced in Eq.~\eqref{eq:bk:bo_lin1} 
accounts for the inherent stochasticity of the learning algorithm—for example, due to 
random initialization when training a neural network. This kind of randomness gives rise 
to \emph{aleatoric uncertainty}, which reflects intrinsic noise in the observations 
and may depend on the specific hyperparameter configuration.

However, as we saw in the previous sections, Bayesian optimization fundamentally relies 
on a surrogate model that not only predicts performance but also quantifies uncertainty. 
In particular, acquisition functions such as expected improvement and lower-confidence bound use the predictive variance of the surrogate model to guide exploration. This 
variance primarily captures \emph{epistemic uncertainty} --- uncertainty due to limited data.
Crucially, this is the kind of uncertainty that can be reduced by collecting more 
observations, and that the acquisition function should leverage when choosing new points to evaluate.

Failing to distinguish between epistemic and 
aleatoric uncertainty can be problematic. If 
the surrogate model attributes too much of 
its predictive variance to aleatoric noise (which cannot be reduced), the acquisition 
function may over-prioritize regions where 
further evaluation is uninformative. 
Conversely, underestimating aleatoric noise can lead to overconfident predictions that neglect areas of genuine variability.
To address this, heteroscedastic models—which explicitly account for input-dependent noise—can be employed. 
For example, \citet{griffiths:2021} train two 
distinct Gaussian processes aiming to predict the unknown
response and the noise separately.

\subsection{Tree Parzen Estimator}
\label{sec:mbm-tpe}
\label{sec:tpe}

We now return to surrogate models and describe an implementation particularly 
well-suited for structured search spaces: the tree Parzen estimator (TPE). 
Unlike the Gaussian process framework, 
which models the conditional distribution of the response function given a configuration,
TPE \citep{bergstra2011algorithms, bergstra_making_2013} adopts a different approach.
It models two conditional densities over configurations, given a performance threshold: 
one for configurations with good performance, and the other for those with poor performance.
Formally, in the TPE framework the surrogate model $\surrogm: \Lambda \times \RSet \to \RSet^2$ is defined as
\begin{equation}
\label{eq:tpe}
\surrogm(\lambda, \beta) = \left( l_\beta(\lambda),\, g_\beta(\lambda) \right),
\end{equation}
where $l_\beta$ and $g_\beta$ are conditional density estimates over $\Lambda$.
The threshold $\beta$ is typically set to the $\gamma$-quantile of the observed responses at iteration $k$, for some $\gamma \in (0, 1)$, e.g., $\gamma = 0.15$. 
This induces a $(\gamma, 1 - \gamma)$ partition of the data $\mathcal{C}_k$, where the set of good-performing configurations is defined as 
\[
\{ (\lambda, f(\lambda)) \in \mathcal{C}_k \mid f(\lambda) < \beta \}\subset \mathcal{C}_k,
\]
based on the available observations and the corresponding partition induced by $\gamma$.
At each iteration $k$, the densities $l_\beta$ and $g_\beta$ are fitted separately to the two subsets of $\mathcal{C}_k$, so that
\[
l_\beta(\lambda) \approx p(\lambda \mid f(\lambda) < \beta), \quad 
g_\beta(\lambda) \approx p(\lambda \mid f(\lambda) \geq \beta).
\]

TPE uses as acquisition function the ratio of the two densities:
\[
\alpha_{\mathrm{TPE}}(\lambda, \hat{f}_k, \beta) = \frac{l_{\beta}(\lambda)}{g_{\beta}(\lambda)},
\]
which assigns higher values to configurations that the surrogate model 
deems more likely to be good-performing than poor-performing.
The ratio is monotonically related to the expected improvement acquisition function with threshold $\beta$. 
\footnote{
Note that unlike in GP-based BO, we cannot set $\beta$ to the minimum observed value, 
as this would imply $\gamma = 0$ and yield an empty set for fitting $l_{\beta}$.
}
In practice, to approximate the maximization of $\alpha$, TPE samples a set of candidate configurations from $l_{\beta}$ and selects the one that maximizes the ratio $l_{\beta}(\lambda) / g_{\beta}(\lambda)$. 
This generative process of sampling $\lambda$ from $l_{\beta}$ allows TPE to flexibly accommodate conditional or hierarchical search spaces by employing appropriate factorizations of $l_{\beta}$ -- for example, tree-structured models.

TPE is conceptually simple, naturally parallelizable, and handles complex, conditional search spaces well. 
Subsequent work has shown that the same acquisition objective can be approximated 
via probabilistic binary classification, which bypasses density estimation and 
leverages standard machine learning tools \citep{Tiao:21}.

\subsection{Bayesian Optimization Vs. Random Search: An Example}
\label{sec:mbhpo-example}

\begin{figure}[t]
  \centering
  \includegraphics[width = 0.8\textwidth]{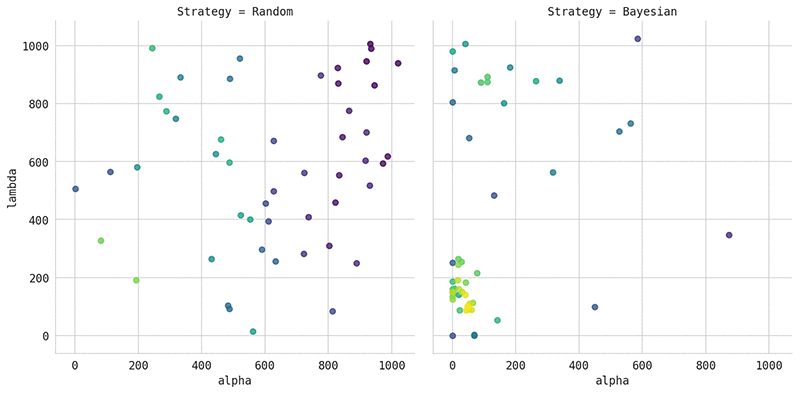}
  \includegraphics[width = 0.8\textwidth]{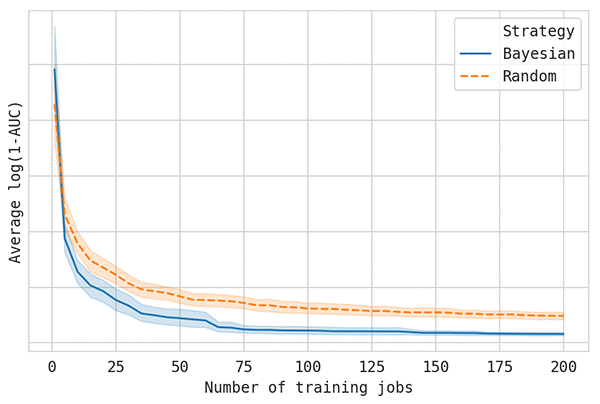}
  \caption{\texttt{Top}: Hyperparameters suggested by random search (left) and BO (right). The $x$-axis and $y$-axis are the hyperparameter values for \texttt{alpha} and \texttt{lambda} and the color of the points represent the quality of the hyperparameters (yellow: better AUC; violet: worse AUC).
    \texttt{Bottom}: Best model score obtained so far ($y$-axis, where lower is better) as more hyperparameter evaluations are performed ($x$-axis).}
  \vspace{-0.3cm}
  \label{fig:random-vs-bayes}
\end{figure}

We illustrate the difference between model-based HPO and random search 
in a concrete example of tuning regularization parameters of a machine learning model to prevent overfitting. 
We tuned the L1 and L2 regularization terms, respectively alpha and lambda, of the XGBoost algorithm \citep{chen2016xgboost} on the UCI direct marketing binary classification dataset from Kaggle. 
The goal was to minimize the AUC. Figure~\ref{fig:random-vs-bayes} presents a comparison between random search and BO (model-based), which models the objective function through a GP and uses the expected improvement as the acquisition function.

Each experiment was conducted using 50 different random seeds, and the results were reported as the average and standard deviation. The left and middle plots reveal that BO consistently suggests better-performing hyperparameters than random search. The right plot further demonstrates that BO consistently outperforms random search across all evaluated hyperparameter configurations.

While BO is particularly effective in a sequential setting, random search remains a valuable approach, especially in distributed environments where many configurations can be evaluated in parallel, significantly reducing wall-clock time.

\section{Other Approaches to Model-Based HPO}
\label{sec:other-surrogates}

Not all model-based methods follow a Bayesian optimization framework.
\citet{ilievski2017efficient} use a deterministic radial basis function model for $\surrogm$ which they interpolate on observations of the response function.
 \citet{hazan2017hyperparameter} focus on Boolean hyperparameters and use a sparse Fourier basis model for $\surrogm$, 
in conjunction with heuristics to estimate the most sensitive hyperparameter entries. They then use a base global optimizer to perform HPO on a restriction of the original hyperparameter space.
The works by \citet{lorraine2018stochastic} and \citet{mackay2019self}, 
related to gradient-based hyperparameter optimization (covered in
Chapter~\ref{ch:gradient-based}),
can also be interpreted as attempts to learn a surrogate of the entire
learning algorithm (applied to a specific dataset), 
i.e., learning a mapping 
$\surrogm: \Lambda \to \hypothesisspace$ that maps hyperparameters to trained models (cf. Eq.~\eqref{eq:algodef}).

Other approaches blend meta- and transfer-learning to model-based HPO by training a surrogate model on multiple datasets and tasks.
In particular, \citep{feurer2015initializing}
extracts meta-features from multiple datasets, including descriptive statistics (number of instance, classes, mean, variance and higher moments) and ``landmarks'' (i.e. the performances of a few quick-to-evaluate learning algorithms on the dataset).
They then use this information to train a surrogate model that initializes a BO process, thereby leveraging prior knowledge to accelerate optimization.
\citet{fusi2018probabilistic} follow instead a probabilistic matrix factorization approach akin to collaborative filtering in recommender systems, 
modeling the performance of configurations
on datasets -- therefore, users and items are ``mapped to configurations \textit{and} datasets).
The learned probabilistic representations are then used to guide the search of next configurations to try via the expected improvement acquisition function.

More sophisticated strategies to exploit parallel compute resources, either with random search or BO, are described in the next chapter.

\section{Discussion}
\label{sec:mbhpo:disc}

Model-based methods tend to be somewhat more complex than other HPO techniques, both conceptually and from an implementation standpoint. They require fitting a (probabilistic) surrogate model with relatively few function evaluations, and to update this fit online whenever new observations become available. 
Moreover, the optimal decision on where to evaluate $f$ next needs to be approximated via an AF depending on the surrogate model, as we saw in Section \ref{sec:acq_func}.
Thus, while there is a thriving ecosystem of free academic software, it is not rare that model-based HPO techniques are implemented and proposed as commercial products \citep{sigopt-web-page, golovin2017google, AMT2020}.
The development of model-based techniques requires several design choices to be made, such as the concrete representation of $\surrogm$. Some configurations may be left to be set by the final user. While authors often provide \textquote{default values}, the presence of many such parameters may make some model-based techniques less accessible. See, e.g., the list of required inputs for the algorithms proposed by \citet{hazan2017hyperparameter} or  \citet{falkner2018bohb}.

While the foremost representatives of these type of methods --- tied to BO --- are classically black-box procedures, model-based techniques may greatly benefit from direct access to the underlying learning algorithm. For instance, several works \citep{swersky2014freeze, klein2017fast, falkner2018bohb} use information gathered from partial execution of $\learningalgo$ to stop unpromising evaluations, which may be additionally resumed at a later stage \citep{swersky2014freeze} as we will discuss in more details in the next chapter.
Before doing so, we discuss what is possibly the biggest practical limitation of conventional model-base methods: parallelizability.

\subsection{Decision Making Under Parallelism}
\label{sec:parallel-mbhpo}

In its original formulation BO is sequential, which is unsuitable for many real-world scenarios where parallel computation resources are readily available in order to increase wall-clock time efficiency. A number of extensions of BO to parallel evaluation settings have been proposed. Although more time efficient, these techniques tend to be less sample efficient than sequential BO. The extensions we discuss aim thus to find a diverse set of configurations to preserve sample efficiency as much as possible.

For simplicity, assume a fixed set of $W$ workers available for parallel evaluations. Recall from Section~\ref{sec:hpo-practical} that systems may either schedule evaluation jobs synchronously or asynchronously. In the synchronous case, we need to suggest batches of $W$ configurations to be evaluated next. A simple way to do this would be to pick the $W$ top-ranked configurations when optimizing our AF. However, this strategy risks suggesting very similar configurations, and their parallel evaluation may reveal little more information than any single one of them. A heuristic approach to suggest a more diverse batch is to run Thompson sampling $W$ times with independently drawn sample paths \citep{Kandasamy:18}, but this can be quite expensive. Another idea is to select batch members greedily one at a time. For each configuration placed in the batch, the AF is modified by a local penalty function, ensuring that subsequently selected points are at least some distance away \citep{Gonzales:16}.

Independent of how configurations are suggested, synchronous job scheduling tends to be substantially less efficient than asynchronous. Namely, the time to process a batch is determined by the slowest one, while up to $W-1$ workers are idle waiting. Such differences in training times are particularly pronounced when searching over parameters determining the model size. In an asynchronous system, whenever a worker becomes available, it can be assigned a new configuration. While BO needs to make single suggestions only (as opposed to batches), this has to be done in the presence of up to $W-1$ pending evaluations, i.e. jobs which have been started for previously suggested configurations, but did not return a metric value yet. Ignoring pending evaluations runs the same risk of selecting highly redundant configurations in subsequent rounds. Fortunately, dealing with pending evaluations is quite related to ensuring diversity in batch selection, and local penalty methods apply just as well \citep{Gonzales:16, Alvi:19}. Simpler approaches start from the observation that in sequential BO, diversity is enforced by conditioning on observed points. The resulting very low posterior variance shrinks the AF to small values near such points. In the kriging believer heuristic \citep{Ginsbourger:10,Desautels:14}, we impute the response for each pending evaluation by its current posterior mean and condition on this pseudo-datapoint. A somewhat more principled approach is fantasizing \citep{snoek_practical_2012,Wilson2018}, where pending target values are integrated out by Monte Carlo: they are sampled several times from the joint posterior over pending configurations, and the AF is averaged over these samples.


\chapter{Multi-Fidelity Methods}
\label{ch:multi_fidelity}

One of the main challenges of HPO is that training and validating a machine learning algorithm each time the objective function is evaluated can be expensive. This is particularly true for modern machine learning algorithms applied to large datasets, such as deep neural networks, which can often take hours or days to train.
In this chapter, we consider how to obtain and exploit low-fidelity signals that serve as approximations of the objective function. 
These auxiliary metrics are faster to evaluate and often provide a strong enough signal to guide the optimization process.

We begin with an overview of different types of fidelities in Section~\ref{sec:mf_hpo}. Section~\ref{sec:mf_rs} explores methods for multi-fidelity optimization based on randomly sampling configurations. Finally, in Section~\ref{sec:mf-modelbased}, we examine multi-fidelity approaches that leverage probabilistic models to guide the search.

\section{Multi-Fidelity Optimization}
\label{sec:mf_hpo}

As noted in Section~\ref{sec:ea-manual}, an important aspect of manual search and hyperparameter tuning is a fast turnaround time. To achieve this, practitioners typically work with a subset of the training data during their initial explorations.
They also do not wait for training to fully converge but instead check validation performance after a few iterations or epochs. While these early metrics are not entirely reliable, they are useful for identifying promising regions of the search space.
In this section, we discuss how these manual practices can be transformed into algorithmic tools that enable automated early stopping of underperforming runs.

Multi-fidelity HPO~\citep{li2017hyperband,klein2017fast,swersky2014freeze} extends the HPO formulation from Section~\ref{sec:the_HPO_problem} by taking the budget that we spend to train a hyperparameter configuration into
explicit account. 
We assume access to inexpensive approximations of the true objective function $f(\lambda)$, denoted as $\ffid{\lambda}{b}$, where the fidelity level is determined by the budget $b \in [\bmin, \bmax]$.

If we set the budget to its maximum value, $b = \bmax$, we recover the original objective function, $\ffid{\lambda}{\bmax} = f(\lambda)$.
From this perspective, standard HPO can be seen as a special case of multi-fidelity HPO.
Additionally, we assume that the approximation of $f(\lambda)$ improves as $b$ increases. However, the cost $c(\lambda, b)$ associated with evaluating $\ffid{\lambda}{b}$—such as training time—also increases with $b$.

While we can freely choose $b$ during the optimization process, our primary interest lies in the performance at the full budget $\bmax$. More formally, our goal is (still) to find $\lambda^* \in \argmin_{\lambda \in \Lambda} \ffid{\lambda}{\bmax}$, which coincides with the formalization of Chapter~\ref{ch:hpo}.
Next, we explore commonly used fidelity measures that can be leveraged to accelerate the optimization process.

\subsection{Iterations}

For iterative machine learning algorithms, such as learning in neural networks, we can interpret the number of iterations as the budget and use the performance at each iteration as an approximation of the final performance~\citep{swersky2014freeze,li2017hyperband,jamieson2016non}. In the case of neural networks, we typically consider epochs as iterations rather than individual mini-batch updates to reduce the overhead of validation error computations.

Figure~\ref{fig:lc} (left) illustrates the validation loss after each epoch of training for different hyperparameter configurations of a multi-layer perceptron.
Even after just a few epochs, we can visually distinguish between poorly performing and well-performing configurations. However, to reliably identify the best configuration, it is often necessary to evaluate the top-performing candidates for the full number of epochs.

\subsection{Training Dataset Subsets}

\begin{figure}
  \centering
  \includegraphics[width=0.49\textwidth]{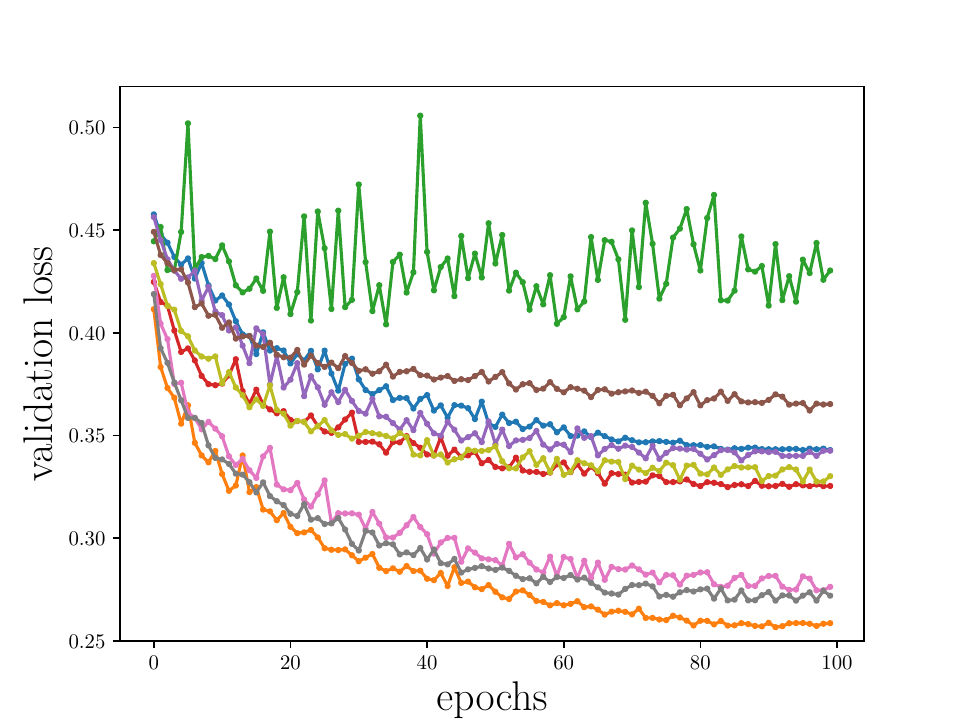}
  \includegraphics[width=0.49\textwidth]{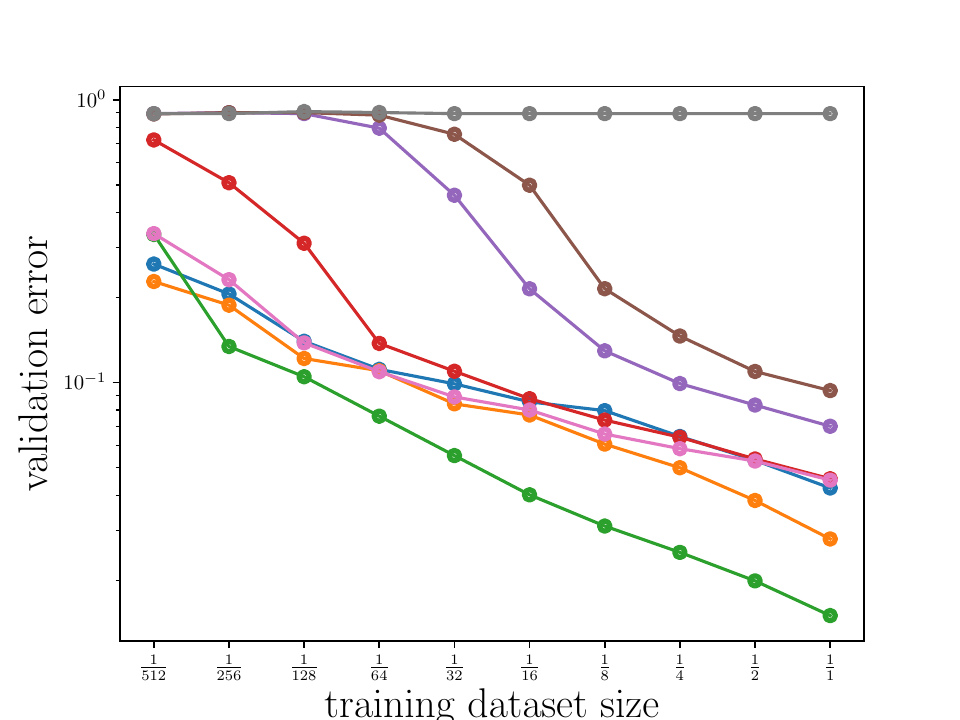}
  \caption{Left: Validation loss of different neural networks evaluated after each epoch of training. Right: Validation error of a support vector machine trained on different subsets of the training data. Each color represents a different hyperparameter configuration.}
  \label{fig:lc}
\end{figure}

The runtime of most machine learning algorithms scales linearly or even super-linearly with the number of training data points. Figure~\ref{fig:lc} (right) shows the validation error of a support vector machine trained on increasing subsets of the MNIST dataset, with each color representing a different hyperparameter configuration. As expected, the validation error decreases as the training dataset size increases. More interestingly, however, the ranking of hyperparameter configurations remains relatively consistent across dataset subsets~\citep{bornschein-icml20}.
In the context of deep neural networks, this observation is related to recent research on scaling laws~\citep{kaplan-arxiv19,hoffmann-arxiv22}, which empirically demonstrate that the training loss of autoregressive transformer models~\citep{vaswani_attention_2017} often follows a power-law function with respect to the size of the training dataset.

In multi-fidelity optimization, we can exploit this phenomenon by interpreting the budget as the training dataset size~\citep{klein2017fast,swersky_multi-task_2013}. To evaluate $\ffid{\lambda}{b}$, we vary only the training dataset size while keeping the validation set fixed to avoid introducing additional noise. Most methods resample the training dataset uniformly at random for each evaluation of $\ffid{\lambda}{b}$, though more sophisticated strategies, such as importance sampling, can also be used to select dataset subsets~\citep{ariafar-aistats21}. Importantly, global statistics such as the class distribution must be preserved to avoid biasing performance estimation.

\subsection{Cross-validation}

As discussed in Section~\ref{sec:generalization}, cross-validation (CV) is a widely used technique for estimating the generalization error of a machine learning method and reducing the risk of overfitting.
Rather than relying on a single training/validation split, we divide the dataset into $K$ subsets or folds, using each fold for validation in turn and averaging our target criterion across them.
The main drawback of CV is its computational cost, which is $K$ times higher than using a single fold. However, we can approximate it using a multi-fidelity approach by treating the number of folds, $b = 1, \dots, K$, as the budget.

Unlike the previous two fidelities, where performance typically improves with increasing budget, CV does not necessarily exhibit this behavior. Instead, the key assumption is that the variance of the generalization performance estimate decreases as the number of folds increases.

\section{Methods Based on Random Search}\label{sec:mf_rs}

In this section, we review algorithms that leverage multiple fidelities and propose new configurations through random sampling.
Random search based multi-fidelity methods were first introduced in the multi-armed bandit (MAB) literature.
We discuss how a simple MAB approach can dynamically halt the evaluation of poorly performing configurations.
Additionally, we discuss how these methods can be adapted for efficient execution in large-scale distributed settings.
The methods covered here assume that the evaluation costs, $c(\lambda, b)$, for any fixed $\lambda$ are linear (or affine linear) in $b$.

\subsection{Successive Halving}
\label{sec:sh}

Successive halving (SH)~\citep{karnin-icml13,jamieson2016non} is an MAB strategy that aims to identify the best hyperparameter configuration from a finite set of configurations.
Given the maximum and minimum budgets, $b_{max}$ and $b_{min}$, for evaluating a single hyperparameter configuration, we first define a set of rung levels:

$$
\mathcal{R} = \{b_{min}\eta^k \mid k\in [T] \}.
$$

We assume, for simplicity, that $b_{max} / b_{min} = \eta^T$, where $T \in \NSet$. Here, $ \eta \geq 2$ defines the halving constant and determines how aggressively SH  prunes configurations.

\begin{algorithm}[t]
\caption{Successive-Halving-HPO}
\label{algo:sh}
\begin{algorithmic}[1]
\Require{Multi-fidelity response function $f_b$}
\Require{Initial budget $b_{min}$, halving constant $\eta \geq 2$, number of rungs $T$}
\Require{Set of $n$ candidate configurations $\mathcal{C}_0 = \{ \lambda_1, \lambda_2,\dots \lambda_n\}$}
\For{$k \gets 1$ \textbf{to} $T$}
\State $r_k \gets b_{min}\eta^k$ \Comment{Define rung level}
\State $\mathcal{F} \gets \{\ffid{\lambda}{r_k} \mid \lambda \in \mathcal{C}_{k-1}\}$ \Comment{Evaluate configurations on the rung}
\State $s \gets \left\lfloor \vert \mathcal{C}_{k-1}\vert / \eta \right\rfloor$ \Comment{Number of configurations to keep}
\State $\mathcal{C}_k \gets \underset{\lambda\in\mathcal{C}_{k-1}}{\arg\mathrm{top}\text{-}s} \ \mathcal{F}$  \Comment{Keep $s$ best performing configurations}
\EndFor
\end{algorithmic}
\end{algorithm}

\begin{figure}
  \centering
  \includegraphics[width=0.9\textwidth]{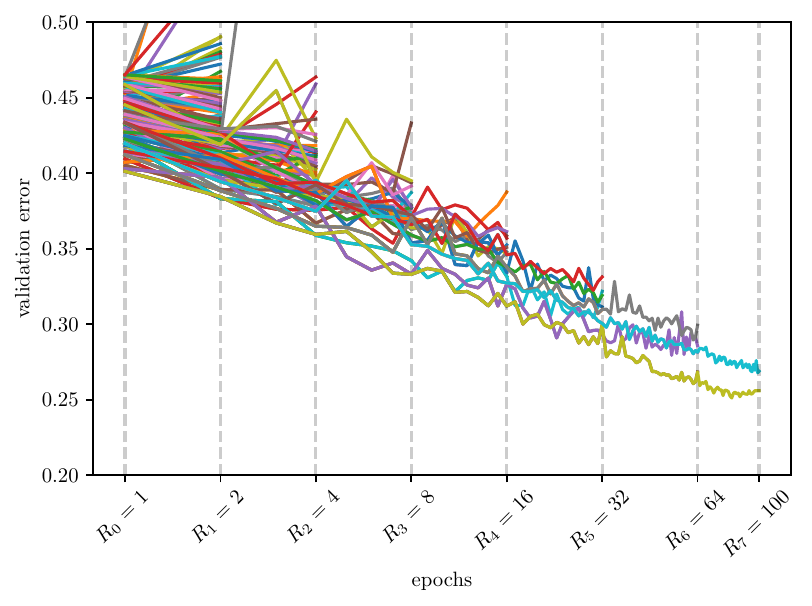}
  \caption{Successive halving splits the maximum budget $\bmax$ into a set of exponentially increasing rungs. Then, in each iteration, successive halving evaluates all configurations on the current rung and promotes the top $1/\eta$ configurations to the next rung.
    Here, half of the configurations are discarded at every rung, and their learning curve is interrupted.
    Only two configurations are evaluated for the full budget $\bmax=100$.}
  \label{fig:sh}
\end{figure}

Algorithm~\ref{algo:sh} shows pseudocode for SH. Given a set $\mathcal{C}_0$ of $n$ hyperparameter configurations, usually drawn uniformly at random, we evaluate them at the first rung level, i.e., each configuration is evaluated with a budget of $b_{min}$ (Line 2). For convenience, the number of initial configurations is often set to $n = \eta^T$, but this can also be arbitrary. 
Afterwards, SH promotes $\frac{1}{\eta}$ of the configurations, sorted by their validation performance, to the next higher rung level $b_{min} \eta$ (Line 4). 
These steps are iterated until we reach the highest rung level, $b_{max}$. See Figure~\ref{fig:sh} for a visualization.

While conceptually simple, SH has demonstrated strong performance in the literature and has even outperformed more sophisticated Bayesian optimization techniques that operate only on the full budget, in terms of achieving the same performance in less wall-clock time~\citep{jamieson2016non,li2017hyperband}.
However, its performance relies on a high correlation between the ranking of hyperparameter configurations across rung levels. In other words, the performance of a configuration at the first rung level should be representative of its performance at the highest rung level. Otherwise, configurations that perform best at $b_{max}$ 
may be discarded too quickly.

\subsection{Hyperband}

The minimum budget $b_{min}$ allocated to each configuration is an important meta-parameter of SH. If $b_{min}$ is too small, we may filter out good configurations based on unreliably low fidelities, while if $b_{min}$ is too large, our advantage over standard random search may diminish. Hyperband~\citep{li2017hyperband} addresses this issue by running different instances of SH in sequence. The $l$-th instance (or bracket) has rung levels 
$$
 \mathcal{R}_l = \{ b_{min} \eta^k \mid k \in \{ l, \dots, T\} \}, 
$$ 
so $l = 0$ corresponds to SH with $T + 1$ rung levels, while $l = T$ corresponds to random search with a single level $b_{max}$. Each bracket has its own number of initial configurations $n_l$, which is chosen such that the total budget spent in each bracket is roughly the same. The risk of choosing too small a value for $b_{min}$ is mitigated at a modest extra expense.

\subsection{Synchronous Successive Halving}

Recall the notation of synchronous versus asynchronous parallelization from Section~\ref{sec:hpo-practical}, and assume that $W$ workers are available.
For synchronous scheduling, we can start $ \min(W, M) $ jobs in parallel, where $M$ is the number of free slots remaining in the currently processed rung.
Here, $W - M$ workers remain idle, which is particularly wasteful for the higher rungs.
For example, if $n = \eta^K$, the highest rung has only a single slot, so $W - 1$ workers will be idle for the maximum budget $b_{max}$.
The same issue arises for asynchronous scheduling. A remedy will be discussed in Section~\ref{sec:mf-asha}.

Hyperband inherits the idle worker issue noted above for synchronous scheduling. However, since brackets are independent of each other, we can work on several of them in parallel, which eliminates idle time for asynchronous scheduling.

\subsection{Asynchronous Successive Halving (ASHA)}\label{sec:mf-asha}

As discussed above, neither SH nor Hyperband can fully exploit parallel computation because they require synchronization among workers evaluating hyperparameter configurations for a specific run.  

This also means that if a configuration is clearly superior to all others on a rung \( r_i \in \mathcal{R} \), it may take a long time before it is promoted to the next rung level \( r_{i+1} \). In fact, all configurations on the current rung \( r_i \) must be evaluated before any promotion decisions can be made.  

\begin{figure}
      \centering
      \includegraphics[width=0.9\textwidth]{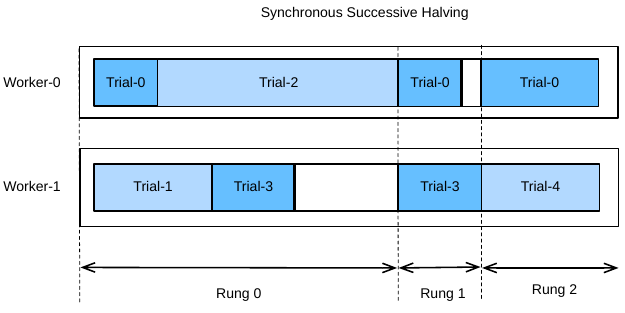}
	 \caption{Scheduling behavior of Synchronous SH with two workers. A darker shade of blue indicates better-performing trials. Some trials may take significantly longer than others, resulting in idle workers until all trials in the current rung are completed.}
      \label{fig:sync_sh}
 \end{figure}

For example, consider Figure~\ref{fig:sync_sh}. In a distributed setting, we refer to the evaluation of a hyperparameter configuration as a trial, and in this scenario, we have two workers evaluating four different trials on Rung 0. If Trial-2 takes significantly longer than the others, worker 1 remains idle until all trials on Rung 0 are completed, delaying progress to Rung 1.

These shortcomings are addressed by Asynchronous Successive Halving (ASHA) \citep{Li:19}.  
In this algorithm, rungs do not have a predefined size but grow dynamically over time.  
Additionally, promotion decisions are made more aggressively.  

\begin{figure}
      \centering
      \includegraphics[width=0.9\textwidth]{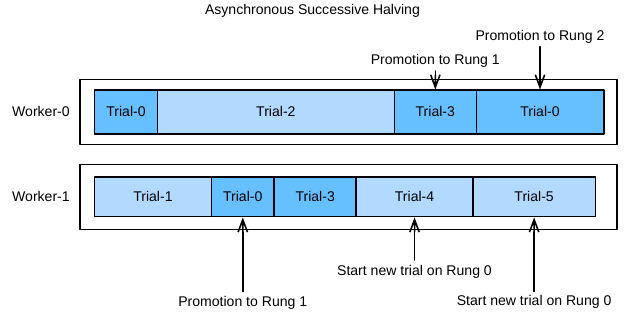}
	 \caption{Scheduling behavior of ASHA. As in Figure~\ref{fig:sync_sh}, a darker shade of blue indicates better-performing trials. ASHA eliminates idle time by immediately promoting a trial to the next rung as soon as it meets the promotion criteria, rather than waiting for all trials in the current rung to be completed. }
      \label{fig:asha}
 \end{figure}

For example, assume that \( \eta = 2 \); a configuration is immediately promoted to the next rung if it ranks in the top \( 50\% \) of the current rung.  
In other words, configurations are promoted as early as possible: once a rung has two entries, one of them advances to the next level.  
Figure~\ref{fig:asha} illustrates this process: as soon as Trial-1 is completed, evaluation of Trial-0 continues on Rung 1—assuming Trial-0 performs better than Trial-1 — rather than evaluating Trial-3 on Rung 0.  

After evaluating Trial-3, no configurations can be promoted because, at this point, only two configurations have been evaluated on Rung 0, and one on Rung 1. As a result, the evaluation proceeds with a new trial at the lowest level.

In ASHA, all available computational resources are utilized immediately, ensuring that workers are never idle. Additionally, superior configurations are quickly promoted toward \( \bmax \) without the need to wait for synchronization points.  
In practice, ASHA often significantly outperforms synchronous SH or Hyperband
in terms of wall-clock time~\citep{Klein:20}.

\section{Model-Based Methods}\label{sec:mf-modelbased}

So far, the multi-fidelity methods we have examined sample hyperparameter configurations purely at random. As we saw in Chapter~\ref{ch:mbm}, more sophisticated methods that use a probabilistic model to sample configurations typically require fewer function evaluations to approach the global optimum compared to random search-based approaches. In this section, we will discuss multi-fidelity strategies that leverage one or multiple probabilistic models to guide the search.

\subsection{Fabolas}

Fabolas~\citep{klein2017fast} models the performance $ \ffid{\lambda}{b} $ of a hyperparameter configuration $ \lambda $ across different subsets $ b \in [\bmin, \bmax] $ of the training data.

Compared to classical Bayesian optimization (see Chapter~\ref{ch:mbm}), we cannot simply optimize standard acquisition functions as described in Section~\ref{sec:acq_func}.  
For example, expected improvement without any further modifications tends to evaluate mostly where \( b = \bmax \), since, as discussed above, here we expect the validation error to be lower than for smaller \( b\).
As a result, this approach would lead to over-exploitation, primarily evaluating configurations at the highest budget.

Instead, Fabolas is based on the entropy search method by \citet{Hennig2012}, which models the probability 
$$
p_{\min}(\lambda^* \mid \mathcal{C}) = \mathbb{P}\left[\lambda^* \in \argmin_{\lambda \in \Lambda} \ffid{\lambda}{}\right]
$$
of $ \lambda $ being the global optimum, where $\mathcal{C}$ are the collected observations (see Section~\ref{sec:bk:hpo:bayesopt}).
The key idea of entropy search is to select, in each iteration, the configuration that provides the most information about $ p_{\min} $, which we quantify by computing the KL divergence
$KL\big[p_{\min}(\lambda\mid\mathcal{C})||\mathcal{U}(\lambda)\big]$ between $ p_{\min} $ and a uniform distribution $ \mathcal{U}(\lambda) $, which serves as a prior assuming that each point $ \lambda \in \Lambda $ is equally likely to be the optimum.
In other words, entropy search selects, in each iteration, the candidate point $ \lambda $ that shifts $ p_{\min} $ from a uniform distribution to a highly peaked distribution around the global optimum.

For Fabolas, we model $ p_{\min} $ only at the highest budget $ b=\bmax $, since we aim to find
$ \lambda^* \in \argmin_{\lambda \in \Lambda} \ffid{\lambda}{\bmax} $.
To balance information gain with the cost of evaluating $ \lambda $, Fabolas maintains a second Gaussian process $ p(c\mid \lambda, b) $ to model the wall-clock time $ c(\lambda, b) $ required to evaluate $ \ffid{\lambda}{b} $.
In each iteration, Fabolas optimizes the following acquisition function:
$$
\alpha(\lambda, b) =
\frac{KL\big[p_{\min}(\lambda\mid\mathcal{C})||\mathcal{U}(\lambda)\big]}{\mu_c(\lambda, b)} 
$$
to select both the configuration $ \lambda $ and budget $ b $ that provide the most information about $ p_{\min} $ while remaining cost-effective to evaluate, where
$ \mu_c(\lambda, b)=\mathbb{E}_{p(c\mid \lambda, b)}[c(\lambda, b)] $ is the expected cost of evaluating $ \lambda $ using budget $ b $.

\subsection{BO-HB: Bayesian Optimization Hyperband}

To combine the sample efficiency of Bayesian optimization with the strong anytime performance of Hyperband, \citet{falkner2018bohb} fit an independent TPE model (see Section~\ref{sec:mbm-tpe}) on each rung level $ b_i \in \mathcal{R} $ to sample new configurations.

More specifically, once the proposed method BO-HB observes \( N_{\min} \) data points at the rung level \( b_i \), it trains two kernel density estimators—\( l(\lambda) \) and \( g(\lambda) \)—to model the probability of a configuration \( \lambda \) achieving a validation error lower or higher than a certain threshold, as defined in Eq.~\eqref{eq:tpe}. 
However, the density estimators are only trained on all data points at the given rung level.

New configurations are then sampled according to the density ratio $ l(\lambda) / g(\lambda) $ instead of being chosen uniformly at random. This process continues until $ N_{\min} $ observations have been collected at the highest rung level $ \bmax $, after which new configurations are sampled exclusively from the model trained at this level.

While the original implementation of TPE by \citet{bergstra2011algorithms} used a univariate kernel density estimator, \citet{falkner2018bohb} proposed using a multivariate kernel density estimator to better capture interactions between hyperparameters.
To fit such a model, at least $ N_{\min} = m + 1 $ data points are required, where $ m $ is the dimensionality of the search space.
Follow up work also considered more sophisticated modelling approaches, such as different GP variants~\citep{klein-arxiv20,wistuba-neurips22} or conformalized quantile regression~\citep{salinas-icml23}.

\begin{figure}
  \centering
  \includegraphics[width=0.9\textwidth]{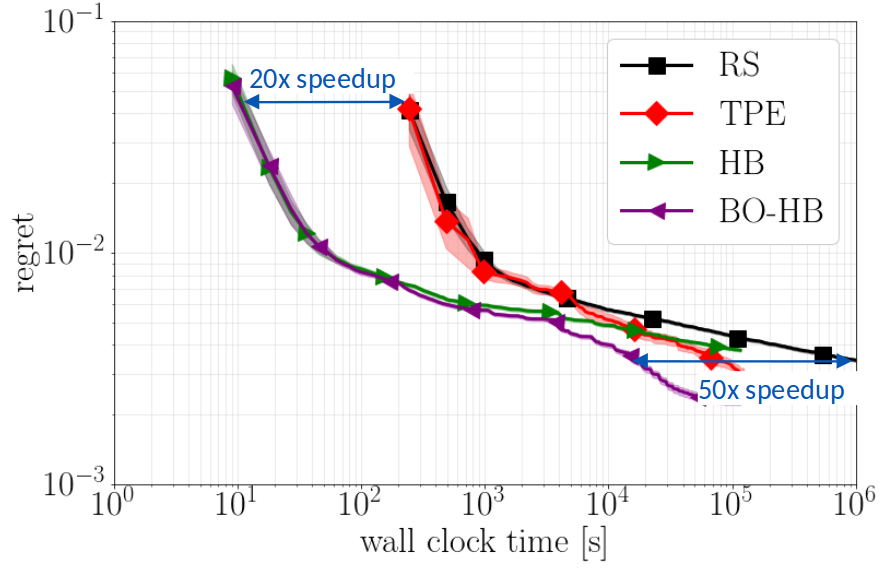}
  \caption{Comparison of random search (RS), TPE, Hyperband (HB) and BO-HB over time from \citet{falkner2018bohb}. BO-HB keeps the strong anytime performance of Hyperband but because of the model samples better configurations over time.
  }
  \label{fig:bohb}
\end{figure}

Figure~\ref{fig:bohb} presents a comparison by \citet{falkner2018bohb} of BO-HB with other approaches for optimizing the hyperparameters of a neural network. Hyperband initially outperforms other methods since it returns configurations much earlier than non-multi-fidelity methods such as random search or TPE. However, compared to model-based approaches like TPE, Hyperband eventually converges toward random search, as it samples configurations uniformly at random and, in later stages, requires evaluating them on larger budgets to improve performance. BO-HB, on the other hand, retains the early speed-up advantage of Hyperband while achieving a performance level similar to TPE in the later stages of the optimization process.

\chapter{Population-Based Methods}
\label{ch:popM}

In this chapter, we discuss a class of randomized methods for global optimization that maintain a population of candidate solution, similar to multi-fidelity  methods discussed in the previous chapter.
Whereas successive halving and related techniques are rooted in the bandit problem and linked to hedging strategies, the methods we explore here draw inspiration from processes of natural evolution  \citep{back1997handbook, simon2013evolutionary}.
Evolutionary algorithms date back to the beginning of computer science \citep{neumann1966theory} and have been used to tackle non-smooth and non-convex problems across many domains.
Methods in this class typically rely on function evaluations only. 
Their application to hyperparameter optimization includes methods like CMA-ES \citep{hansen1996adapting}, population-based training (PBT) \citep{jaderberg2017population} and particle swarm optimization \citep{lorenzo2017particle}, which we review below. 
Despite the lack of theoretical 
backing, population-based methods, especially evolutionary algorithms, 
have enjoyed continued attention in HPO, due to their adaptability to complex 
search spaces and their strong empirical performances \citep{loshchilov2016cma}.

\section{Evolutionary Algorithms}

Traditionally, evolutionary algorithms (EAs) distinguish between evolutionary strategies (ES) and genetic algorithms (GA).
GA refers to a class of methods that deal with binary problems and discrete objective functions while ES is concerned with real-valued, continuous objectives.

The terminology of EA techniques is borrowed from biology: the set of points is called population and configurations are called individuals, chromosomes or genotypes. 
Iterations are called generations and the objective function is called fitness. 
Typically, at the end of each generation, the fitness of all individuals is observed, making EAs inherently parallel algorithms --- traditional implementations usually requiring synchronization. 
After observing the fitness of the population, a series of transformations  is applied to the individuals.
Common types of transformations, which collectively implement explore/exploit strategies in this context, include the following one: 
\begin{itemize}
    \item \textit{Selection}, global transformations that prime a new generation, typically by discarding individuals with low fitness and selecting a subset of fitter individuals to become ``parents'' for the next generation; 
    \item \textit{Recombination} or \textit{crossover}, global transformations that consist in generating new individuals combining slices (e.g. subsets of hyperparameter coordinates) of two or more parents;
    \item \textit{Mutation}, local modifications designed to escape local optima that introduce random changes in the individual, for instance adding independent Gaussian noise.
\end{itemize}
Other strategies like \textit{elitism} (i.e., keep the best individuals unchanged) and \textit{niching} (i.e., maintain a diverse population) are often used in conjunction to the explore/exploit strategies outlined above.

We outline this class of methods in Algorithm \ref{alg:pbh}.
We can think of individuals and generations as draws from $N$ distributions $\{p_k^j\}$, where $N$ is the population size.
Here, the superscripts $j\in [N]$ index individuals and the subscripts $k\in [T]$ index generations, where $T$ is the number of generations.
These distributions implement the strategies described above, and evolve at each generation using global information from the current generation. 
For instance, elitism can be implemented by setting $p_k^j=\delta_{\lambda_k^j}$ for the best performing configurations, where $\delta$ is a Dirac delta distribution.
In contrast to random or quasi-random search methods covered in Chapter \ref{ch:ea}, 
population-based methods keep in general $N$ distinct distributions rather than a global one that are updated iteratively rather than fixed.

\begin{algorithm}
\caption{\texttt{Population-Based-HPO}}
\label{alg:pbh}
\begin{algorithmic}[1]
\Require{Response function $f$}
\Require{Population size $N$ and number of generations $T$}
\Require{Initial distributions $\{p^j_0\}_{j\in [N]}$}
\For{$k \gets 1$ \textbf{to} $T$}
    \For{$j \gets 1$ \textbf{to} $N$} \Comment{Typically executed in parallel}
        \State $\lambda_k^j\sim p^j_{k-1}$  \Comment{Draw an individual}
        \State Evaluate $f(\lambda_k^j)$		\Comment{Evaluate its fitness (if needed)}
    \EndFor
    \State Update $p^j_k$ from $p^j_{k-1}$ using $\{(\lambda^i_k, f(\lambda^i_k))\}_{i\in [N]}$  \Comment{``Evolve''}
\EndFor
\end{algorithmic}
\end{algorithm}

Various implementations of ES and GA have been proposed in HPO e.g.\ for tuning hyperparameters of support vector machines~\citep{friedrichs2005evolutionary}, for neural architecture search~\citep{real2019regularized, miikkulainen2024evolving} (see Section~\ref{sec:nas} for a discussion on neural architecture search), parameters of games~\citep{kunanusont2017ntuple} and even entire machine learning algorithms~\citep{real2020automl}.
In the next sections, we review three particular implementations that have received notable attention.

\section{Covariance Matrix Adaptation Evolution Strategy (CMA-ES)}

CMA-ES \citep{hansen1996adapting, hansen2016cma, loshchilov2016cma} maintains a global multivariate Gaussian 
distribution from which it samples  individuals for the new generations. 
That is, the generation at iteration $k+1$ is drawn according to
\begin{equation*}
    \lambda_{k+1}^j \sim p_k^j =  \mathcal{N}(\mu_k, \sigma_k^2\Sigma_k)  \quad \text{for } j \in [N]
\end{equation*}
where $N\in\NSet^+$ is the population size, $\mu_k$ and $\Sigma_k$ are the distribution parameters and $\sigma_k>0$ is an overall standard deviation, akin to a step size.
Referring to Algorithm \ref{alg:pbh}, this means in CMA-ES $p^j_k=\mathcal{N}(\mu_k, \sigma_k^2\Sigma_k)$ for all $j\in [N]$.

The parameters of the distribution are evolved at each iteration following a selection and recombination principle. 
Suppose $\{\lambda_k^j\}_{j=1}^N$ are sorted in increasing objective value, i.e. $f(\lambda_k^1)\leq f(\lambda_k^2)\leq \dots \leq f(\lambda_k^N)$. 
\footnote{We recall we seek to minimize $f$, hence the first points are the best points.}
Then the mean is updated as a convex combination of the best $1\leq M \leq N$ points: 
\begin{equation*}
    \mu_{k+1} = \sum_{i=1}^M w_i \lambda_{k}^i,
\end{equation*}
where the weights $w_i$ are positive and sum to 1. The threshold $M$ and the recombination weights are parameter of the method. \citet{hansen2016cma} suggests $M=N/2$ and $w_i \propto M - i + 1$ as reasonable defaults.

The covariance matrix is adapted similarly by reestimating it from the $M$ best points of the previous generation. That is:
\begin{equation}
\label{eq:cma-est}
    \Sigma_k = \sum_{i=1}^M w_i (\lambda_k^i - \mu_k)(\lambda_k^i - \mu_k)^\intercal
\end{equation}
where the $w_i$ are the recombination weights. In practice the update of Eq.~\eqref{eq:cma-est} could be unreliable especially for small populations and \citet{hansen2003reducing} proposes to maintain a moving average for the covariance matrix. 
Current implementation of CMA-ES incorporate other strategies to improve robustness and efficiency of the algorithms like adaptation of the step-size $\sigma_k$ and generalizations of Eq.~\eqref{eq:cma-est}  \citep[see][]{hansen2019pycma}.

\section{Population-Based Training (PBT)}

\label{sec:pbt}

\begin{figure}[t]
\centering
\includegraphics[width=0.75\textwidth]{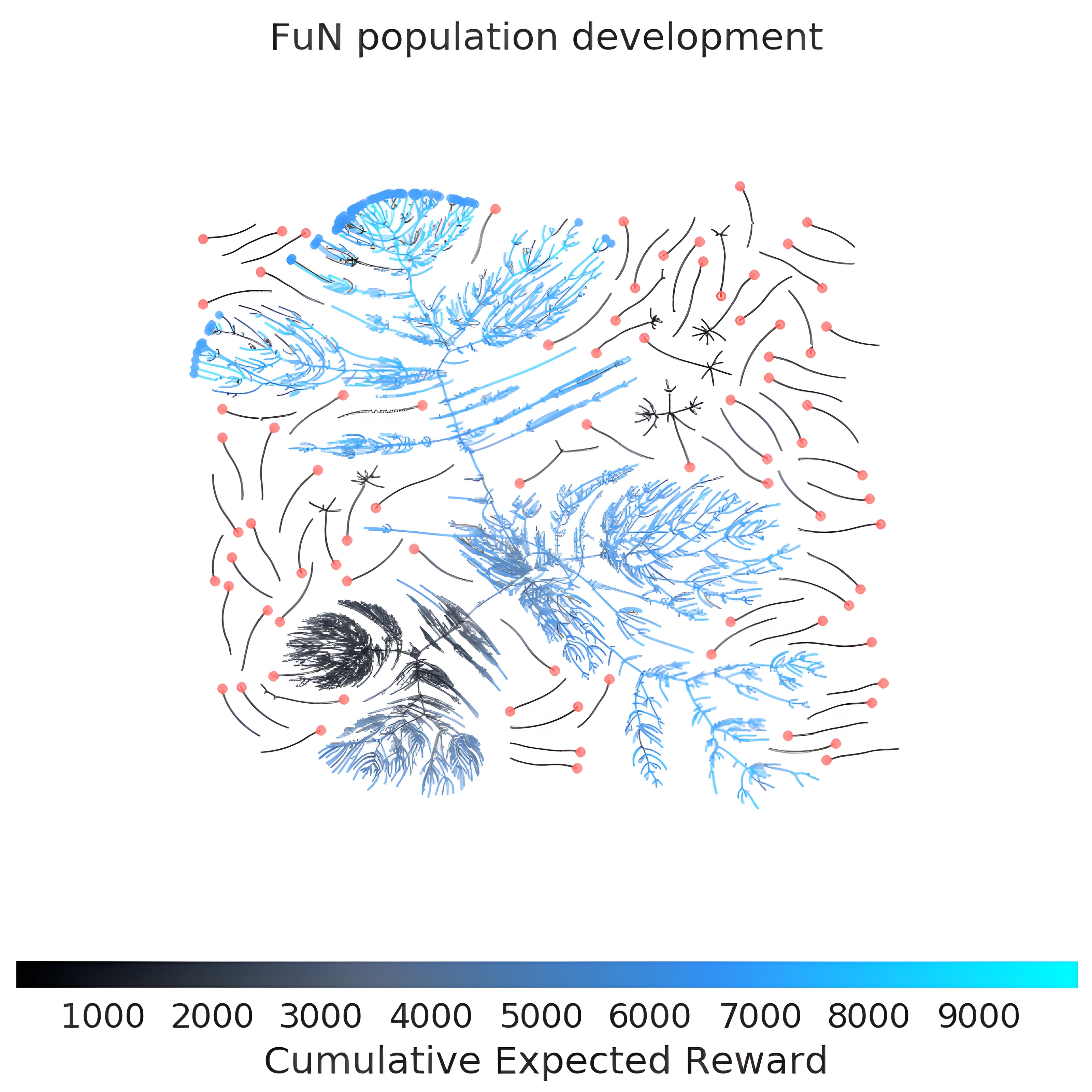}
\caption{Evolution of models trained with PBT for Atari game-playing policies. Red dots represent initial points in the parameter and hyperparameter spaces. Paths represents progressions via parameter training. Branching points represent applications of the exploit and explore operations, namely, a copy of the parameters and hyperparameters of better performing individuals onto worse performing one and a subsequent perturbation of their hyperparameter configurations. The performance score is color-coded, with light blue color representing better performing models. Reproduced from \citep{jaderberg2017population}.}
\label{fig:pbt}
\end{figure}

\citet{jaderberg2017population} propose a parallel method that maintains a population of model and hyperparameter pairs that are evolved as the models are trained. 
The procedure, devised in the context of learning deep neural networks in challenging scenarios like image generation or reinforcement learning, 
consists in interleaving neural network training (by gradient descent) with applications of evolutionary transformations of  both the parameters of the models and the hyperparameter of the learning algorithm. 
PBT is an online hyperparamter optimization algorithm, as hyperparameters are updated alongside model parameters.  
In other words, in contrast to many techniques we discussed so far,  PBT yields a hyperparameter sequence (or schedule) rather than a single hyperparameter configuration. 
\footnote{
We discuss in more generality hyperparameter schedules and online HPO in Section~\ref{sec:online-hpo}.
}

Suppose one can ``decompose'' the learning algorithm of interest into a series of stages: $\learningalgo=\learningalgo_T\circ \dots \circ \learningalgo_1$, where each map $\learningalgo_t$ represents a stage of the learning process that can be warm-started given the outcome of the previous stage.
This is particularly straightforward for neural network training: given a space of weights $\mathcal{W}$, each $\learningalgo_t:\mathcal{W}\times \Lambda \to \mathcal{W}$ could be, for instance, 
an epoch of gradient-based optimization of the network weights.
Let $N\in\NSet$ be a population size and $\{(\theta_0^j, \lambda_0^j)\}_{j\in[N]}$ be a set of initial model weights and hyperparameters, where $\theta_0^j\in\mathcal{W}$ and $\lambda_0^j\in\Lambda$.
These pairs represent the ``individuals'' in PBT and they can be initialized drawing from suitable distributions. 

Generations in PBT correspond to stages of the learning algorithm. At each generation $t\in [T]$ the weights are updated as follows:
\begin{equation*}
    \tilde{\theta}_{t}^{j} = \learningalgo_t(\theta_{t-1}^j, \lambda_{t-1}^j) \quad \text{for } j\in[N].
\end{equation*}
Then, evolutionary transformations are applied to both parameters and hyperparameters before proceeding to the subsequent learning stage, on the basis of a fitness function $\valmetric: \mathcal{H} \to \RSet$ (e.g. a validation error, cf. Section~\ref{sec:the_HPO_problem}) evaluated on the current model weights.
\footnote{Note that here we cannot explicitly refer to the response function $f$, as hyperparameters may change throughout iterations; see also the discussion on hyperparamter schedules in Section \ref{sec:online-hpo}.}
Suppose we sort individuals in ascending order so that $\valmetric(\tilde{\theta}_t^{1}) \leq \valmetric(\tilde{\theta}_t^{2}) \leq \dots \leq \valmetric(\tilde{\theta}_t^N)$.
\citet{jaderberg2017population} propose two types of operations:
\begin{itemize}
    \item \textit{Exploit}: a selection strategy that involves copying the parameters and hyperparameters of better performing individuals onto worst performing ones; for instance, one may set $(\theta^N_t, \lambda^N_t) = (\tilde{\theta}^1_t, \lambda^1_{t-1})$. \citet{jaderberg2017population} suggest to replace the worst-performing 20\% with individuals randomly drawn from the best-performing 20\%; 
    \item \textit{Explore}: a mutation transformation that brings over the parameter form the last learning stage, but either resamples the hyperparamters from the starting distribution, or perturbs them, e.g. by setting $(\theta_t^j, \lambda_t^j) = (\tilde{\theta}_t^j, \lambda_{t-1}^j + \varepsilon)$ for a perturbation $\varepsilon$ such that  $\lambda_{t-1}^j + \varepsilon \in \Lambda$. 
\end{itemize}
PBT further uses elitism, leaving the top performing individuals unchanged; that is, setting for some $M\in[N]$, $(\theta_t^i, \lambda_t^i) = (\tilde{\theta}_t^i, \lambda_{t-1}^i)$ for $i\in[M]$.
Figure \ref{fig:pbt} present a visualization of a phylogenetic tree 
of game-playing policy functions for Atari games. 
The performance metric is the expected episodic reward, and the models deep neural networks trained using the Feudal Networks (FuN) framework \citep{vezhnevets2017feudal}.

 \citet{tao2020learning} combine PBT with gradient-based optimization of the hyperparameters.
 The authors propose an ``hyperparameter mutation'' algorithm that learns a dynamic trade-off between local search via gradient-based hyperparameter optimization and global exploration via PBT.
We discussed gradient-based HPO methods in the next chapter.

\section{Particle Swarm Optimization}

Particle swarm optimization (PSO) methods draw inspiration from the social behavior of animals, like ant colonies or bird flocks \citep{kennedy1995particle}, as well as from the evolution of particle trajectories in a physical system.
The general idea is that each candidate solution evolves according to a simple equation that weights a momentum term, or velocity, and one or more attractor terms to steer the trajectory toward better performing points.

Assume for simplicity that $\Lambda=\RSet^m$, and let $\{(\lambda_1^j, v_1^j)\}_{j=1}^N$ be an initial population of hyperparameter and velocity pairs, both drawn randomly from suitable distributions (e.g.,  uniform distributions).
\citet{lorenzo2017particle} propose an application of PSO to hyperparameter optimization that uses the following equation:
\begin{align}
    \lambda_{k+1}^j & = \lambda_k^j + v_k^j, 
\label{eq:pso-state}    
    \\
    v_{k+1}^j & = \alpha_1 v_k + \alpha_2 r_{l} \circ (\bar{\lambda}_k^j - \lambda_k) + \alpha_2 r_g \circ (\lambda^* - \lambda_k), \label{eq:pso-velocity}
\end{align}
where $\alpha_1, \alpha_2, \alpha_3\in \RSet^+$ are three coefficients, $r_l, r_g\in\RSet^m$ are two randomly drawn vectors and $\circ$ is the Hadamard product. The configuration $\bar{\lambda}_k^j$ and $\lambda_k^*$ are, respectively,  the best particle and population points found so far,
that is,
\begin{equation*}
	\bar{\lambda}_k^j = \argmin_{\lambda \in \bar{\Lambda}^j_k}f(\lambda) \quad \text{for } j \in[N],
\end{equation*}
where $\bar{\Lambda}^j_k=\left\lbrace \lambda^j_t \in \Lambda \mid t\in [k-1]\right\rbrace$, and
\begin{equation*}
\lambda^*_k = \argmin_{\lambda \in \Lambda^*_k} f(\lambda),
\end{equation*}
where $\Lambda^*_k = \left\lbrace \lambda_t^j \in \Lambda \mid t\in [k-1], \, j\in [N]\right\rbrace$.
The second term in the velocity update of Eq.~\eqref{eq:pso-velocity} can be interpreted as a local memory attractor that pushes the current configuration toward the individual best configuration found so far, while the third term is a global attractor that pushes toward the global best configuration found so far. 
The randomness introduced by $r_l$ and $r_g$ serves to encourage diversity throughout the search.
Referring to Algorithm \ref{alg:pbh}, each individual distribution is implicitly modeled by the Eqs.~\eqref{eq:pso-state}-\eqref{eq:pso-velocity}.

\section{Discussion}

Despite the lack of theoretical guarantees, population-based methods enjoy broad applicability and strong empirical performances.
In addition, this class of methods is designed to readily scale to parallel and distributed compute systems. 
Although traditional EA and PSO implementation assume synchronicity, one may easily derive asynchronous versions of many algorithms. 
For instance in PSO, one may replace the global best configuration $\lambda^*$ with the best known solution to each  individual $j$. 
However, population-based algorithms require a considerable amount of customization due to the presence of several method parameters,  task-specific heuristics and design choices.
For instance, it might be challenging to define meaningful mutation or perturbation strategies, 
or implement effective niching. 
These considerations may undermine the applicability of these methods to novel learning algorithms and HPO problems, and may preclude 
accessibility to non-expert users.
Besides, they typically require a considerable computational power, benefiting usually from large population sizes.

\chapter{Gradient-Based Optimization}
\label{ch:gradient-based}

In some cases, the response function, $f(\lambda)$, is differentiable
with respect to $\lambda$, enabling the application of optimization
algorithms that require the gradients of their objective. 
Assuming $f(\lambda)$ to be differentiable excludes the direct optimization of
discrete
hyperparameters such as discrete choices, for example, which nonlinear
activation function should be used in a layer of a neural network, or
integer-valued hyperparameters, such as the dimension of a
layer.
There is, however, a distinct advantage of gradient-based
methods for HPO, namely the possibility of tuning a (very)
high-dimensional hyperparameter vector $\lambda$, which would be rather
difficult to do with model-based algorithms. This opportunity allows
us to completely rethink when a variable should be considered as a
parameter or as a hyperparameter.
\citet{maclaurin2015gradient} suggested considering the
initial weights of a neural network as a large hyperparameter vector,
an approach that in their own words effectively ``blurs the
distinction between learning and meta-learning.'' Indeed,
gradient-based methods reveal some deep connections between HPO and
some forms of meta-learning, which we expand on in Section~\ref{sec:meta-learning}.
For example, the idea of tuning
initial weights from several learning tasks was
successfully exploited by \citet{finn_model-agnostic_2017} in their
model-agnostic meta learner.

The crux of gradient-based HPO lies in computing approximate gradients of the response function w.r.t. hyperparameters. 
In this chapter, we cover three main technical approaches: implicit differentiation (Section \ref{sec:implicit_hypergradients}), iterative differentiation (Section \ref{sec:gbhpo:iterative}) and fixed-point iterations (Section \ref{sec:gbhpo:fixed}). 
Next,  we explore some use cases in Section \ref{sec:gbhpo:use}.
We touch on extensions that allow for the gradient-based optimization 
of discrete hyperparameters in Section \ref{sec:gbhpo:discrete}, and conclude with a
 discussion centered on computational considerations in Section 
 \ref{sec:gbhpo:discussion}.

\section{Preliminaries}

At a high level, gradient-based HPO methods start from an initial hyperparameter configuration  $\lambda_0 \in \Lambda$ and iteratively update such configuration as follows:
\begin{equation}
    \lambda_r = \mathrm{Proj}_{\Lambda}\left[\lambda_{r-1} - \nu \hat{\nabla} f(\lambda_{r-1})\right] \qquad r=1, 2,\dots
\end{equation}
where $\hat{\nabla}f$ informally denotes some approximation of the gradient of the HPO objective $f$, $\nu>0$ is a step-size and $\mathrm{Proj}_{\Lambda}$ is the projection onto $\Lambda$. 
Gradient-based optimization of the loss function inside the learning
algorithm $A$ is extremely common, especially in deep
learning.
Hence, to avoid confusion, we will call $\hat{\nabla}f$  the \emph{hypergradient}.

Not every HPO problem is suitable for
gradient-based optimization.
Recalling the general formulation of the HPO problem in Section \ref{sec:the_HPO_problem}, 
one underlying assumption shared by all gradient-based HPO methods is that the performance metric $\valmetric$ be differentiable w.r.t. the model output of $A$.
In this chapter, to simplify the notation we take the output of $A$ to be a parameter vector $w \in \weightspace\subseteq\mathbb{R}^d$ and let the performance metric take values in the weight space directly, that is  $\valmetric:\weightspace \times \setofdata\ \to \mathbb{R}$.
Hence,
\begin{equation}
    f(\lambda) = \valmetric(w(\lambda), \Dval) = \valmetric(A(\Dtrain, \lambda), \Dval).
\end{equation}
In the following, it will be useful to explicitly write $w$ as a function of $\lambda$, but we will drop $\Dtrain$ and $\Dval$ to simplify the notation.

In order to compute hypergradients, it is also necessary to specify at
least some details of the learning algorithm $A(\mathcal{D},\lambda)$
that appears in (\ref{eq:hpo0}) and that we have mostly treated as an opaque 
box until now. 
Gradient-based HPO methods are distinguished mainly by the assumptions they make about $A$, which then translate into different computational schemes for $\hat{\nabla} f$. 
Next, we review the main approaches.

\section{Hypergradients via Implicit Differentiation}
\label{sec:implicit_hypergradients}
Many learning algorithms may be conceptually expressed as solutions to
optimization problems:
\begin{equation}
    \label{eq:gbhpo:algo}
    A(\Dtrain, \lambda) \in \argmin_{u\in \weightspace} \mathcal{J}(u, \lambda),
\end{equation}
where $\mathcal{J}:\weightspace\times \Lambda\to\mathbb{R}$ 
is an objective function (e.g., the training loss) that depends on the data,
and $\weightspace\subseteq \mathbb{R}^d$ is a parameter vector.
We have seen a concrete example in Eq.~\eqref{eq:logreg}.
If $\mathcal{J}$ is a smooth function w.r.t. $u$ then, the first-order optimality condition implies that 
\begin{equation}
    \label{eq:fw:implicit_eq}
    \nabla_{u} \mathcal{J}(u, \lambda) = 0.
\end{equation}
Notably, neural networks and many related deep learning
techniques are formulated in these terms.
If $\mathcal{J}$ is twice differentiable w.r.t. $u$ at a point $\bar{\lambda}\in\Lambda$ and $\partial_w^2 \mathcal{J}(u, \lambda)\in\mathbb{R}^{d\times d}$ is invertible, we can
invoke the implicit function theorem \citep[see, e.g., ][]{krantz2012implicit} to Eq.~\eqref{eq:fw:implicit_eq} which assures us that (locally) there exist a function $w:U\subset \Lambda \to \weightspace$ such that $\nabla_{u} \mathcal{J}(u, \lambda)_{|u=w(\lambda)} = 0$ for all $\lambda\in U$, and $A(\Dtrain, \lambda)=w(\lambda)$. We can then locally express the HPO objective as the nested function $f(\lambda)=\valmetric(w(\lambda), \Dval)$. Now, if $\valmetric$ is differentiable, we have that
\begin{equation}
    \label{eq:fw:implicit:gradf}
    [\nabla f(\lambda)]^\intercal  = \partial_{\lambda} f(\lambda) =  
    \partial_{w} \valmetric\left(w(\lambda), \mathcal{V}\right)
    \partial_{\lambda} w(\lambda)  \qquad \forall \lambda \in U, 
\end{equation}
where, again due to the implicit function theorem,  
\begin{equation}
    \label{eq:fw:implicit:dw}
    \partial_\lambda w(\lambda) =  - \left[ \partial_u^2 \mathcal{J}(u, \lambda)_{|u=w(\lambda)} \right]^{-1} 
     \partial^2_{u, \lambda} \mathcal{J}(u, \lambda)_{|u=w(\lambda)}.
\end{equation}
Equations \eqref{eq:fw:implicit:gradf} and \eqref{eq:fw:implicit:dw} give us the analytic expression of the exact gradient of the HPO objective $f(\lambda)$.
Such expression is, however, only implicit.
In fact, we almost never have an explicit formula for the map $w(\lambda)$.

Let us denote by $w_T$ an approximate solution of the inner problem, e.g. found with an iterative optimization algorithm. 
Then, after substituting $w(\lambda)$ with $w_T$ and plugging Eq.~\eqref{eq:fw:implicit:dw} into Eq.~\eqref{eq:fw:implicit:gradf}, one can set $q^\intercal = - \partial_u \valmetric \left[\partial_u^2 \mathcal{J}\right]^{-1}\in\mathbb{R}^d$
and approximately solve the linear system 
\begin{equation}
    \label{eq:fw:linsys}
    \partial^2_u \mathcal{J}(u, \lambda)_{|u=w_T} q = -\nabla_{u} \valmetric(u)_{|u=w_T}. 
\end{equation}
Call $q_{T, K}$ an approximate solution of Eq.~\eqref{eq:fw:linsys}. Then an approximate gradient of the outer objective is given by 
\begin{equation}
    \label{eq:fw:implicit_approx_hypergrad}
    \nabla_{T, K} f(\lambda) = 
     \left[\partial^2_{u, \lambda} \mathcal{J}(u, \lambda)_{|u=w_T}\right]^{\intercal} q_{T, K}.
\end{equation}
The two possible sources of approximations stem from ($i$) the solution of the inner problem and ($ii$) of the linear system in Eq.~\eqref{eq:fw:linsys}. 
A particular implementation of this computation scheme depends on how Eq.~\eqref{eq:fw:linsys} is solved.
One classic procedure is the conjugate gradient (CG) method \citep{hestenes1952methods}, 
which features a linear rate of convergence that depends on the conditioning number of the Hessian of the inner objective.

This computational approach has been studied by \citet{larsen_design_1996, larsen1998adaptive, bengio_gradient-based_2000,  chapelle2002choosing, seeger2007cross}. More recently,
\citet{pedregosa_hyperparameter_2016} proposed an approximation scheme whereby solutions to the inner problems and to the linear Eq.~\eqref{eq:fw:linsys} become more precise as the number of HPO iterations $r$ grows, proving convergence for algorithms which feature strongly convex (inner) objectives.

\section{Hypergradients via Iterative Differentiation}
\label{sec:gbhpo:iterative}

Another approach to computing hypergradients involves explicitly accounting for an iterative training dynamics.
Methods based on 
iterative
differentiation further assume that solutions to the learning algorithm problem  of Eq.~\eqref{eq:gbhpo:algo} are sought (with approximation)
by iterating an optimization dynamics for $T$ steps:
\begin{equation}
  \label{eq:dynamics}
  s_t = \Phi_t(s_{t-1},\lambda,\mathcal{D}) \ \ t=1,\dots,T.
\end{equation}
We call $s_t=\transpose{[\transpose{w}_t,\transpose{v}_t]} \in\mathbb{R}^d$
 the ``state'' of the optimizer, which contains the current
parameters, 
$w_t$, and a (possibly empty) set of auxiliary variables $v_t$ such as
velocities and moments for algorithms like Adam and RMSProp \citep{ruder:2017:overview_gradient_descent}.
This process 
yields a solution $w_T=w_T(\lambda)$.
The transition
maps
$\Phi_t : \mathbb{R}^d \times \mathbb{R}^m \times \setofdata \rightarrow \mathbb{R}^d$
are assumed to be differentiable and describe the actual operations
associated with a specific optimizer.
Typically, $\Phi_t$ depends explicitly on the
current minibatch $\mathcal{D}_t$ (e.g., for stochastic gradient descent) and it internally requires the
computation of the inner objective gradients:
\[
g_t \doteq \nabla_w \bigl. \mathcal{J}(w,\lambda, \mathcal{D}_t) \bigr|_{w=w_{t-1}}.
\]

Note that $\Phi$ depends on $\lambda$ both explicitly and implicitly via
$s_{t-1}$. Also, $\lambda$ might contain quantities related to the optimizer
itself, such as the learning rate.
As an example, the
dynamics (\ref{eq:dynamics}) for stochastic gradient descent with
momentum are
\begin{equation}
  \label{eq:momentum}
  \begin{array}{lcl}
    v_t & = &\mu v_{t-1} + g_t \\
    w_t & = & w_{t-1} - \eta v_{t-1}
  \end{array} 
\end{equation}
where $\mu$ and $\eta$ are the momentum term and the learning rate,
and may be included in $\lambda$ if desired.

In a nutshell, iterative differentiation schemes for gradient-based HPO 
rely on the application of automatic differentiation (AD) to the 
composite map $\valmetric \circ \Phi_T \circ \dots \circ \Phi_1$. 
We briefly survey AD next, and refer the reader to the book by \citet{Griewank08:_evaluat_deriv}
for an in-depth account.

\subsection{Automatic Differentiation}
Automatic (or algorithmic) differentiation
(AD)~\citep{Griewank08:_evaluat_deriv,baydin2017automatic} is the
standard tool for computing gradients in deep learning.  Suppose we
have a generic differentiable function $G:\mathbb{R}^N\mapsto\mathbb{R}^M$ with
$(\eta_1,\dots,\eta_M) = G(\xi_1,\dots,\xi_M)$ and we are interested
in computing the Jacobian matrix, with entries
$\frac{\partial \eta_i}{\partial \xi_j}$.  In AD, we begin by
describing $G$ as a computational graph $(V,E)$, which is a directed
acyclic graph whose nodes $V$ include inputs $\xi_j$, outputs
$\eta_i$, and intermediate variables that are computed by applying
elementary functions to inputs or other intermediate
variables, that can be scalars, vectors or tensors. Elementary functions are typically atomic operations such
as (tensor) sum, (Hadamard) product, and (elementwise) transcendental functions, whose partial
derivatives are readily known.  Edges $E$ in the graph describe
functional dependencies (i.e., they link intermediate variables to the
elementary functions they are fed into). Forward propagation, following
a topological sort of the graph, allows us to compute
$G(\xi_1,\dots,\xi_M)$.  Two AD strategies are available, forward and
reverse mode, which are based on the definition of local derivatives
and a message-passing procedure based on the standard chain rule of
calculus.

In forward mode AD, the local derivative of $v_k$ is defined as
\[
\dot{v}_k \doteq \frac{\partial v_k}{\partial \xi_j}
\]
and can be updated by the recursion
\begin{equation}
\dot{v}_k = \sum_{\ell \in \pi_k}\dot{v}_\ell \frac{\partial v_k}{\partial v_\ell}
  \label{eq:forward}
\end{equation}
initialized as $\dot{\xi_j}=1$, where $\pi_k$ are $v_k$'s parents, and
iterated following a topological sort of $(V,E)$.
In the above equation, if $v_k$ and $v_\ell$ are vector-valued,
$\frac{\partial v_k}{\partial v_\ell}$ denotes a Jacobian matrix.
The
desired derivatives can be finally read as $\dot{\eta}_i$. The same
propagation allows us to compute the $j$-th column of the Jacobian and
to obtain the whole Jacobian, 
where the propagation needs to be repeated for
each input $\xi_j$. Assuming elementary functions can be computed in
$O(1)$, the total computational cost is $O(N|E|)$. Forward mode is thus
advantageous for functions with few inputs and many outputs. If on the
other hand $N=O(|E|)$, as it happens for example for the objective
function of neural networks, forward mode is as inefficient as
computing derivatives numerically. It is therefore not surprising that
forward mode is never used to train neural network models (although
there is a remarkable exception when training recurrent networks on
data streams~\citep{williams_learning_1989}, a setting where reverse
mode cannot be applied).

In reverse mode, the strategy used by the well known backpropagation
algorithm~\citep{rumelhart1986learning}, the local derivative of $v_k$,
usually called \textit{adjoint}, is defined as
\[
\bar{v}_k \doteq \frac{\partial \eta_i}{\partial v_k}
\]
and can be updated by the backward recursion

\begin{equation}
  \bar{v}_k = \sum_{\ell \in \gamma_k}\bar{v}_\ell \frac{\partial v_\ell}{\partial v_k}
  \label{eq:backward}
\end{equation}
initialized as $\bar{\eta_i}=1$, where $\gamma_k$ are $v_k$'s
children, and iterated following a reversed topological sort of
$(V,E)$. The desired derivatives can be finally read as
$\bar{\xi}_j$. The same propagation allows us to compute the $i$-th
row of the Jacobian and to obtain the whole Jacobian, propagation
needs to be repeated for each output $\eta_i$, for a total
computational cost of $O(ME)$. Reverse mode is thus advantageous for
functions with many inputs and few outputs. However, since the
computation proceeds in a backward way, the memory used to store the
values of intermediate nodes $v_k$ cannot be freed until the end of
the propagation, yielding a memory cost of $O(V)$.

\subsection{Computing Hypergradients}

\begin{figure}
  \centering
  \includegraphics[width=0.99\textwidth]{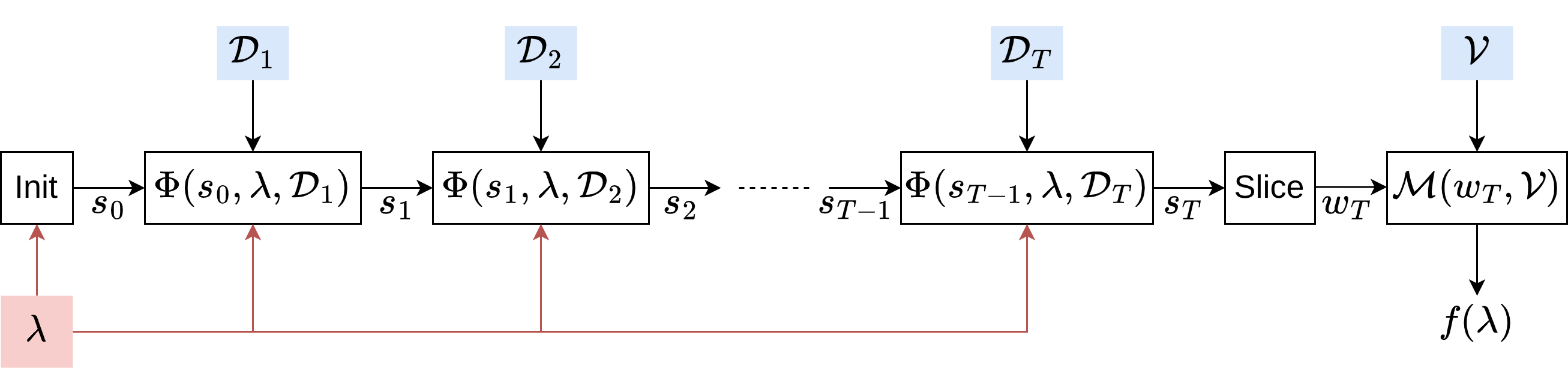}
  \caption{Computational graph of the response function for forward and reverse mode hypergradient calculation.}
  \label{fig:computational_graph}
\end{figure}

We are now ready to apply the principles of AD to the differentiation 
of the validation error with respect to $\lambda$.
A simplified view of the computational graph mapping $\lambda$ into
the response $f(\lambda)$ is shown in
Figure~\ref{fig:computational_graph}, where minibatches
$\mathcal{D}_t$ and initial weights $w_0$ are supplied as given
constants\footnote{Note that $w_0$ could be equally well included in
  $\lambda$ but here we keep it distinct for clarity.}.  Here the
number of inputs $M$ is the dimension of the hyperparameter vector and
there is a scalar ($N=1$) output. Forward and reverse calculations are
presented in the work of \citet{franceschi2017forward} and repeated below.

In the case of forward mode, the local variable is $\dot{s}_t$ and is
udpated by the recursion

\begin{equation}
  \dot{s}_t \doteq \frac{\partial s_t}{\partial \lambda} =
  \left.\frac{\partial \Phi(a,\lambda,\mathcal{D}_t)}{\partial a} \dot{s}_{t-1}\right|_{a=s_{t-1}}
  + \left. \frac{\partial \Phi(s_{t-1},b,\mathcal{D}_t)}{\partial b}\right|_{b=\lambda}
  \label{eq:hg_forward}
\end{equation}
where $\frac{\partial \Phi(a,\lambda,\mathcal{D}_t)}{\partial a}$ and
$\frac{\partial \Phi(s_{t-1},b,\mathcal{D}_t)}{\partial b}$ are
$d\times d$ and $d\times m$ Jacobian matrices, respectively. The base
step is $\dot{s}_0 = \frac{\partial s_0}{\partial \lambda}$ which is
zero unless the hyperparameters directly affect the initial weights
(e.g., their range, if randomly initialized, or even their values as
in~\citet{maclaurin2015gradient}). The desired gradient is finally
obtained at the last node:
\[
  \nabla f(\lambda) = \left.\frac{\partial
  	\mathcal{M}(u, \mathcal{V})}
  {\partial u}\right|_{
  	u=w_T(\lambda)} \dot{s}_T
\]
where $w_T$ is the proper slice of $s_T$.

In the case of reverse mode, the adjoint $\bar{s}_t$ is initialized
as
\[
  \bar{s}_T=\left.\frac{\partial
  	\mathcal{M}(u, \mathcal{V})}
  {\partial u}\right|_{
  	u=w_T(\lambda)}
\]
and updated as
\[
  \bar{s}_t \doteq \frac{\partial f}{\partial s_t} =
  \left.\bar{s}_{t+1}\frac{\partial \Phi(a,\lambda,\mathcal{D}_t)}{\partial a}\right|_{a=s_{t-1}}.
\]
Since the variable $\lambda$ is a parent of all $s_t$,
Eq.~(\ref{eq:backward}) requires us to sum all the contributions:
\begin{equation}
  \nabla f(\lambda) = \sum_{t=1}^T \bar{s}_t
  \left. \frac{\partial \Phi(s_{t-1},b,\mathcal{D}_t)}{\partial b} \right|_{b=\lambda}.
  \label{eq:hg_backward}
\end{equation}

It must be remarked the striking similarity between hypergradient
calculations and gradient calculations in recurrent neural networks
(RNN). Indeed, the computational graphs are identical, where $s_t$
matches the ``state'' of an RNN, $\lambda$ the weights, and $\Phi$ the
state update function. Equations (\ref{eq:hg_forward}) and
(\ref{eq:hg_backward}) closely resemble real-time recurrent
learning~\citep{williams_learning_1989} and backpropagation through
time~\citep{werbos1990backpropagation}, respectively.

\section{Fixed-Point Methods}
\label{sec:gbhpo:fixed}

Another possible approach
that sits
in between the iterative and the implicit differentiation schema is given by the so-called fixed-point method. 
The method is based on the observation that one can rewrite the condition of Eq.~\eqref{eq:fw:implicit_eq} as a fixed point equation of an appropriate contractive dynamics 
\begin{equation}
    \label{eq;fp1}
    w(\lambda) = \Phi(w(\lambda), \lambda), 
\end{equation}
which could be, for instance, gradient descent $\Phi(w, \lambda) = w - \eta \nabla \mathcal{J}$. \footnote{As in the iterative case, the dynamics may be defined on an augmented state, for example when using gradient descent with momentum.}
Then, the application of the implicit function theorem to Eq.~\eqref{eq;fp1} yields to
\begin{equation*}
    \frac{d w(\lambda)}{d \lambda} =  \left( I - \frac{\partial \Phi(w(\lambda), \lambda)}{\partial w} \right)^{-1} \frac{\partial \Phi(w(\lambda), \lambda)}{\partial w \partial \lambda}
\end{equation*}
and, repeating the procedure described above, Eq.~\eqref{eq:fw:linsys} becomes
\begin{equation}
    \label{eq:fw:linsys_fp}
    \left( I - \partial_w \Phi(w_T, \lambda)\right)^\intercal q = \nabla \valmetric(w_T).
\end{equation}
Starting from any point $q_{T,0}\in\mathbb{R}^d$, one can iterate
\begin{equation}
    \label{eq:fw:iter_fp}
    q_{T, k} = \partial_w \Phi(w_T, \lambda)^\intercal  q_{T, k-1} +  \nabla \valmetric(w_T) \qquad k\in[K]
\end{equation}
to find an approximate fixed-point of Eq.~\eqref{eq:fw:linsys_fp} and then plug $q_{T, K}$ into Eq.~\eqref{eq:fw:implicit_approx_hypergrad} to obtain a hypergradient.

This computational scheme emerges in more general cases where $\Phi$ is not necessarily an optimization dynamics: $\Phi$ may represent a stable recurrent neural network \citep{miller2019stable}, a graph neural network \citep{scarselli2009graph} or an equilibrium model \citep{bai2019deep}. 
In these contexts, the fixed-point algorithm is usually known as recurrent backpropagation \citep{almeida1987learning, liao2018reviving}. 
Warm-starting $q_{T, 0}$ with the vector obtained at the previous optimization iteration (of $\lambda$) may speed up the convergence of Eq.~\eqref{eq:fw:iter_fp}.

\section{Use Cases}
\label{sec:gbhpo:use}

Gradient-based approaches have mostly found application in academic
contexts. Their flexibility, however, allows us to tune very
high-dimensional hyperparameter vectors, as showcased in several papers.

One possibility is to attach a hyperparameter to every weight in a deep network.
We have already mentioned the idea proposed by~\citet{maclaurin2015gradient} 
regarding the initial values for the weights as hyperparameters.
In the same paper, the authors also suggest a closely related opportunity
where a regularization penalty is associated to every individual weights.
The meta-learning technique presented by \citet{franceschi2018bilevel} can
be interpreted as treating all weights in the first layers of a deep network
as hyperparameters, and optimize the loss on new tasks with respect to the
weights in the classification head.

Another direction that has been studied is to attach hyperparameters to
data. \citet{maclaurin2015gradient} demonstrated an extreme experiments
where image pixels themselves are hyperparameters. In a more realistic
scenario, hypergradients have been used to tune the hyperparameters of
differentiable data augmentation procedures, yielding significant speedups
without performance loss compared to other data augmentation searching
methods~\citep{lin:2019:online_hyperparameter_learning,%
  hataya:2020:faster_autoaugment_learning}. A different opportunity is to
assign a cost weight to each example in a supervised learning setting so
that the overall loss for a training set of $n$ examples is
$\sum_{i=1}^n \lambda^{(i)} \mathcal{J}(h(x^{(i)}),y^{(i)})$. Here we have $n$
hyperparameters $\lambda^{(i)}$. \citet{franceschi2017forward} show how this
idea can be exploited to filter out from the training set examples with
noisy (or perhaps poisoned) labels. In this regard, it is interesting to
note that (although not directly related to HPO) the problem of building
training set attacks on linear models can be formulated in terms of bilevel
optimization~\citep{mei:2015:using_machine_teaching} and has commonalities
with gradient-based HPO in its mathematical structure.

We finally note here that tuning the whole learning rate schedule has also
been tackled with hypergradient approaches~\citep{baydin_2017_online,%
  donini2019opt,%
  micaelli2021gradient,%
  sharifnassab:2024:metaoptimize_framework_optimizing}. This specific
problem is covered in more detail in Section~\ref{sec:online-hpo}.

\subsection{Probabilistic Reparameterization of Discrete Hyperparameters}
\label{sec:gbhpo:discrete}

The presence of discrete hyperparameters in the response function --- and therefore in the training objective
or dynamics ---  prevents a straightforward
application of gradient-based HPO techniques we just described.
For some of such hyperparameters, however, it might make sense
to reformulate  them
as random variable that follow some discrete distribution. 
For instance, consider a discrete binary choice $\lambda\in \{0, 1\}$ (e.g., indicating whether to use an $L_1$ or $L_2$ regularizer in Eq.~\eqref{eq:logreg}).
Then, one may define a discrete binary random variable $\iota$ as follows:
\begin{equation}
	\iota \sim \mathrm{Ber}(\theta), \qquad \theta \in [0, 1]
\end{equation}
and reinterpret $\lambda$ as a drawing of $\iota$, where $\mathrm{Ber}(\theta)$ is 
the Bernoulli distribution with success parameter $\theta$. 
Similarly, one may reformulate hyperparameters encoding multiple choices via categorical distributions.
Crucially, the distribution parameter $\theta$ is now continuous and one 
may consider optimizing the response function 
\begin{equation}
	\label{eq:gbhpo-reparam}
	g(\theta) = \mathbb{E}_{\iota \sim \mathrm{Ber}(\theta)} [f(\iota)] = \theta f(1) + (1-\theta) f(0)
\end{equation}
instead of $f$,  having that $\min_\theta g(\theta) = \min_\lambda f(\lambda)$.
Although Eq.~\eqref{eq:gbhpo-reparam} might look like a circular argument, the value of such reformulation is unlocked when one computes (approximate) hypergradients via appropriate continuous relaxation or discrete gradient estimators
\citep{niculae2023discrete}.
These techniques may be combined with the procedures described in this chapter, making it possible to avoid the exponential cost of computing 
$f$ on all configurations.
This is still an active area of research: successful examples include the optimization of graph structures in transductive learning \citep{franceschi2019learning} and topological order of nodes in DAG learning \citep{zantedeschi2023dag}, deep learning architectures \citep{liu2018darts} and dropout coefficients \citep{gal2017concrete}.

\section{Discussion}
\label{sec:gbhpo:discussion}

Implicit differentiation methods are \textquote{oblivious} to the way the approximate solution $w_T$ has been found. 
This means that they have a computational advantage in memory over the reverse-mode differentiation as they do not require storing the entire dynamics. 
If implemented using algorithmic differentiation, they have a run-time advantage over forward-mode differentiation since they only deal with vector quantities (the $q_{T, K}$'s). 
However, for this very reason, they do not directly allow computing the gradient of hyperparameters related to the optimization of the inner problem, such as the initialization mapping or the learning rate. 
Furthermore, they are also quite sensitive to the proprieties of the inner problem (e.g., for the fixed-point method, the dynamics must be a contraction), and therefore are applicable to a narrower class of problems.
A comparative theoretical study between the three main approaches covered in this chapter is presented in the work by \citet{grazzi2020iteration}.

Computational disadvantages of plain iterative reverse mode are related to memory requirements.
Indeed, the memory complexity is $O(TW)$, with $W$ the
number of weights, which can quickly become prohibitive when tuning the
hyperparameters associated with modern networks. The problem has been
addressed in different ways. \citet{maclaurin2015gradient} suggested
to completely avoid the storage of $s_t$, recomputing it in the
backward pass by reversing the learning
dynamics of Eq.~\eqref{eq:dynamics}. Since the operation is numerically
unstable, reconstructed initial weights may be very different from the
original ones. As a remedy, they proposed an effective numerical
strategy capable of overcoming the loss of information due to finite
precision arithmetic. A general technique for reducing the memory
requirements in reverse mode AD is
\textit{checkpointing}~\citep[Ch. 12]{Griewank08:_evaluat_deriv},
where only a subset of the possible checkpoints is stored, namely
$w_{jC}$ for $j=1,\dots\lceil T/C\rceil$, being $C$ a given window
size.  In so doing, one must redo the forward pass from $jC$ to
$(j+1)C-1$ to recover the missing $w_t$, effectively doubling the
computational effort. It is easy to see that the best window size is
$O(\sqrt{T})$, which allows us to complete hypergradient calculations
with a memory size of $O(W\sqrt{T})$. This technique has been
considered by \citet{shaban2019truncated}, together with a more
radical truncation strategy similar to truncated backpropagation
through time~\citep{DBLP:journals/neco/WilliamsP90}, which however
introduces further bias in the computation of hypergradients.


\chapter{Further Topics}
\label{ch:adt}

In the last five chapters we covered the main classes of algorithms for hyperparameter optimization.
Here, moving away from particular techniques,  we discuss horizontally  advanced topics in HPO such as multi-objective and constrained optimization (Section \ref{sec:multi-objective-hpo}), online HPO and hyperparameter schedules (Section \ref{sec:online-hpo}) and neural architecture search (Section \ref{sec:nas}). 
Next, we dedicate Section \ref{sec:meta-learning} to meta-learning, a closely related topic in the literature, discussing similarities and differences with the typical hyperparameter optimization setting. 
We conclude the chapter with a brief discussion on transfer of hyperparameters across model sizes, a recent topic that 
is relevant especially for the tuning of large foundational models 
(Section \ref{sec:hp-transfer-scale}).







\section{Optimizing for Multiple Objectives}
\label{sec:multi-objective-hpo}
In many real world settings there is not just one, but multiple, potentially conflicting, objectives of interest. 
For example, when working with hardware-constrained systems one desires predictive models that are accurate and have low resource consumption. Assume one needs to decide between two solutions $\lambda_1, \lambda_2$. Each individual objective induces a distinct ordinal relationship between the candidates, but while $\lambda_1$ might dominate $\lambda_2$ in one objective, it might be worse in another. Because there is rarely a single best solution, the multi-objective hyperparameter optimization (MO) problem is to identify the Pareto front --- the set of all non-dominated solutions --- to help users make an informed trade-off decision. 
More formally, consider a function $f:\Lambda\to\outputspace$ mapping points in domain $\Lambda$ to a space $\outputspace$, which in the following we consider to be an $n$-dimensional objective space that we aim to minimize. 
For two points $\lambda_1, \lambda_2 \in \Lambda$, we say $\lambda_1$ weakly dominates $\lambda_2$ if $f(\lambda_1)_i \leq f(\lambda_2)_i$ for all objectives $i \in [n]$. We write $\lambda_1 \succeq \lambda_2$ to denote weak domination.  
Furthermore, $\lambda_1$ is said to dominate $\lambda_2$ if $\lambda_1 \succeq \lambda_2$ and $f(\lambda_1)_i < f(\lambda_2)_i$ for some objective $i$. We write $\lambda_1 \succ \lambda_2$ for domination. The Pareto front $\mathcal{P}_f$ contains all nondominated points in $\Lambda$, formally: 
$$\mathcal{P}_f = \{\lambda \in \Lambda \mid \nexists \lambda' \in \Lambda : \lambda' \succ \lambda\}.$$ 

Since the Pareto front is often infinite, multi-objective optimizers seek a finite approximation set $B \subset \outputspace$ of objective vectors near the Pareto front. The quality of $B$ can be measured by the dominated hypervolume indicator $\mathcal{H}$~\citep{Zitzler1998:Multiobjective}. Given a reference point $r$, $\mathcal{H}(B)$ equals the hypervolume dominated by $B$:
\begin{align*}
\mathcal{H}(B) = \text{Vol} \left(\{y \in \outputspace \subseteq \RSet^n \mid \exists y' \in B: y' \succeq y \wedge y \succeq r\}\right).
\end{align*}
Computing $\mathcal{H}(B)$ involves partitioning the space into hypercubes, an operation that scales exponentially with the number of objectives $n$. This exponential scaling poses a key bottleneck for hypervolume-based multi-objective optimizers.

\paragraph{Examples.}
A popular example of an MO scenario is the case of algorithmic fairness which aims to train accurate models subject to fairness criteria. Prior work in the field has proposed several fairness criteria, whose suitability depends on the application domain, and it has been shown that criteria, commonly referred to as acceptance-based and error-based, can conflict with each other~\citep{friedler2016possibility,fairness2018verma,kleinberg2018inherent}. However, obtaining solutions that satisfy various definitions to some extent and expose empirical trade-offs between various definitions can still be important from a societal perspective.
Another way to tackle MO scenarios is the case of gradient-free optimization with stochastic constraints. For example, when enhancing the accuracy of a machine learning model, considerations may include constraints on model memory or prediction latency. This occurs in the practical application of deploying machine learning models on mobile devices, where memory cannot exceed a certain pre-specified threshold, or when models power devices with tight requirements on inference latency. The objective and constraint functions in such cases are real-valued and observed together. Alternatively, in tuning a deep neural network (DNN), one may aim to prevent training failures due to out-of-memory errors. Here, the constraint feedback is binary, indicating whether the constraint is violated, and the objective is not observed under constraint violation. It is common to treat constraints with equal importance to objectives, employing a surrogate model as discussed in Chapter 4. Unfeasible evaluations incur a cost, such as wasted resources or compute node failure, necessitating minimization of their occurrence. 
However, avoiding the unfeasible region should not compromise convergence to optimal hyperparameters. Often, optimal configurations lie at the boundary of the feasible region, presenting a challenge in balancing the costs of unfeasible evaluations and the need for exploration.

\subsection{Multi-Objective Optimization}
Initial efforts towards MO optimization for HPO~\citep{Jin2006:Multi} were inspired by evolutionary techniques  \citep{Yao1999:Evolving, Nolfi2002:Evolution, Igel2005:Evolutionary, friedrichs2005evolutionary}.
To mitigate the large computational cost of evaluating a single hyperparameter configuration (e.g., training a DNN), sample-efficient Bayesian optimization (BO) techniques have been introduced~\citep{hutter_sequential_2011, snoek_practical_2012, bergstra2011algorithms}.

Multi-objective Bayesian optimization (MBO) techniques can be categorized into three classes. First, \emph{scalarization}-based MBO methods map the vector of all objectives to a scalar $V$~\citep{Knowles2006:Parego, Zhang2009:Expensive, Nakayama2009:Sequential, Paria2019:Flexible, Golovin2020:Random} and then use conventional single-objective BO techniques. The scalarization is defined by using a weight vector $w \in \RSet^n$.
Some common choices for $V$ are:
\begin{itemize}
\item Inner product between objective and weight vectors (also known as random weights): $V_{\text{RW}}(f(\lambda), w) = f(\lambda)^\top w$;
\item The scalarization used in~\cite{Knowles2006:Parego}, called ParEGO:
$$V_{\text{ParEGO}}(f(\lambda), w) = \max_{j \in \{1, \dots, n\}}(w_j f(\lambda)_j) + \rho (f(\lambda)^\top w);$$
\item The scalarization used in~\cite{Golovin2020}:
$$ V_{\text{Golovin}}(f(\lambda), w) = \min_{j \in \{1, \dots, n\}} (\max (0, f(\lambda)_j / w_j))^n.$$
\end{itemize}
Note that by optimizing different scalarizations, different solutions along the Pareto front can be obtained. The choice of $V$ influences the subset of the Pareto front that is found, and the right selection might change depending on the task at hand.

A second class of MBO techniques builds on a performance measure of Pareto front approximations, namely the \textit{dominated hypervolume}~\citep{Zitzler1998:Multiobjective}. Both the expected hypervolume improvement (EHI)~\citep{Emmerich2011:Hypervolume} and probability hyper-improvement (PHI)~\citep{Keane2006:Statistical} operate by extending their single-objective BO counterparts --- expected improvement~\citep{Mockus1978:Application, Jones1998:Efficient} and probability of improvement~\citep{Kushner1964:New}, respectively. Other methods include step-wise uncertainty reduction (SUR)~\citep{Picheny2015:Multiobjective}, smsEGO~\citep{Ponweiser2008:Multiobjective, Wagner2010:Expected} and expected maximin improvement (EMMI)~\citep{Svenson2010:Multiobjective}.

Finally, \emph{information-theoretic} MBO approaches aim to select points that reduce uncertainty about the location of the Pareto front. Methods in this class tend to be more sample efficient and scale better with the number of objectives. PAL~\citep{Zuluaga2013:Active} iteratively reduces the size of a discrete uncertainty set. PESMO~\citep{Hernandez2016:Predictive} adapts the predictive entropy search (PES)~\citep{Hernandez2014:Predictive} criterion. Pareto-frontier entropy search (PFES)~\citep{Suzuki2019:Multi} is suitable when dealing with decoupled objectives. MESMO~\citep{Belakaria2019:Max} builds on the max-value entropy-search criterion~\citep{Wang2017:Max} and enjoys an asymptotic regret bound. Building on MESMO, two recent works have proposed MF-OSEMO~\citep{Belakaria2020:Multi} and iMOCA~\citep{Belakaria2020:Information}, two multi-fidelity based information-theoretic MBO techniques which internally use multi-fidelity Gaussian processes.

\subsection{Constrained Hyperparameter Optimization}

When optimizing for multiple targets, some objectives may have predefined constraints on their values. For example, one might seek to optimize a given objective function while enforcing a strict memory constraint. In this section, we focus on this scenario.

Formally, the goal in a constrained optimization scenario is to minimize the target response function $f(\lambda)$, subject to a constraint $c(\lambda) \le \delta$. Here, we limit our attention to modeling the feasible region by a single function $c(x)$. As the latent constraints are pairwise conditionally independent, an extension to multiple constraints is straightforward, yet notationally cumbersome. 
Both $f(\lambda):\Lambda \rightarrow \RSet$ and $c(x): \Lambda \rightarrow \RSet$ are unknown and need to be queried sequentially. The constrained optimization problem is defined as follows:
\begin{equation}\label{eq:ystar}
   \min_{\lambda\in\Lambda}  f(\lambda), \quad \text{s.t. } \; c(\lambda) \leq \delta 
\end{equation}
where $\delta\in\RSet$ is a threshold on the constraint. For instance, in the case of BO, the latent functions $f(\lambda)$ and $c(\lambda)$ can be conditionally independent in our surrogate model, with different Gaussian process priors placed on them.

We consider two different setups, depending on what information is observed about the constraint. Many existing works \citep{Gardner14, Gelbart14, Lobato15a} assume that {\em real-valued feedback} is obtained on $c(\lambda)$, just as for $f(\lambda)$. In this case, both latent functions can be represented as Gaussian processes with Gaussian noise.
Unfortunately, this setup does not cover practically important use cases of constrained hyperparameter optimization. For example, if training a deep network fails with an out-of-memory error, we cannot observe the amount of memory requested just before the crash. Rather, we only know that an out-of-memory error occurred, which is a binary outcome.
Covering such use cases requires handling {\em binary feedback} on $c(x)$, even though this is technically more difficult. 
We can assume an evaluation returns $z_f\sim \mathcal{N}(z_f | f(\lambda), \alpha_f^{-1})$ and $z_c\in\{-1, +1\}$, where $z_c = -1$ for a feasible, $z_c = +1$ for an unfeasible point. Here, $z_f$ is the observed value of the objective function, while $z_c$ is the observed value of the constraint. We never observe the latent constraint function $c(\lambda)$ directly. We assume $z_c\sim \sigma(z_c c(\lambda))$, where $\sigma(t) = \frac{1}{1 + e^{-t}}$ is the logistic sigmoid, but other choices are possible. We can then rewrite the constrained optimization problem (\ref{eq:ystar}) as follows:
\[
   y_{\star} = \min_{\lambda\in\Lambda} \left\{ f(\lambda) \mid \mathbb{P}[z_c = 1 | \lambda] =
  \sigma(c(\lambda)) \le \sigma(\delta) \right\}.
\]
This formulation is similar to the one proposed by \citet{Gelbart14}. The parameter $\sigma(\delta)\in (0,1)$ controls the size of the (random) feasible region for defining $y_{\star}$. Finally, note that in the example of out of memory errors training failures, the criterion observation $z_y$ is obtained only if $z_c = -1$: if a training run crashes, a validation error is not obtained for the queried configuration.
Apart from an experiment by \citet{Gelbart14}, the case of constrained BO work covering the binary feedback case is covered by \citet{perrone_cmes}.

The most established technique to tackle constrained BO is constrained EI (cEI) \citep{Gardner14, Gelbart14, snoek_scalable_2015, letham2019}.
A separate regression model is used to learn the constraint function $c(\lambda)$, typically a Gaussian process, and EI is modified in two ways. First, the expected amount of improvement of an evaluation is computed only with respect to the current \emph{feasible} minimum. Second, hyperparameters with a large probability of satisfying the constraint are encouraged by optimizing
$cEI(\lambda) = \mathbb{P}[ c(\lambda)\le \delta] \cdot EI(\lambda)$, where $\mathbb{P}[ c(\lambda)\le \delta]$ is the posterior probability of $\lambda$ being feasible under the constraint model, and $EI(\lambda)$ is the standard EI acquisition function, see Section \ref{sec:acq_func}. %

Several issues with cEI have been noted by \citet{Lobato15a}.
First, the current feasible minimum has to be known, which is problematic if all initial evaluations are unfeasible. 
A workaround is to use a different acquisition function, initially focused on finding a feasible point \citep{Gelbart14}. In addition, the probability of constraint violation is not explicitly accounted for in cEI. A confidence parameter was introduced by \citet{Gelbart14}, but it is only used to recommend the final hyperparameter configuration. %
Another approach was proposed by \citet{Lobato15a}, where PES is extended to the constrained case. Constrained PES (cPES) can outperform cEI and does not require the workarounds mentioned above.
However, it is complex to implement, expensive to evaluate, and unsuitable for binary constraint feedback. Drawing from numerical optimization, different generalizations of EI to the constrained case were developed by \citet{Picheny2016} and by \citet{Ariafar2019}. The authors represent the constrained minimum by way of Lagrange multipliers, and the resulting query selection problem is solved as a sequence of unconstrained problems.
An alternative option that does not require expensive computations is the extension of another acquisition function to the constrained case, namely Max-value Entropy Search \citep{perrone_cmes}. This acquisition function handles binary constraint feedback, does not require numerical quadrature~\citep{Picheny2016}, and does not come with a large set of extra hyperparameters~\citep{Ariafar2019}.


\section{Hyperparameter Schedules and Online HPO}
\label{sec:online-hpo}

The vast majority of learning algorithms do not produce a hypothesis in one shot, but rather  refine it incrementally over the course of many iterations. 
In the previous chapters we already discussed algorithms designed around this observation,
such as those leveraging multi-fidelity approaches (Chapter \ref{ch:multi_fidelity}), and hypergradients (Section \ref{sec:gbhpo:iterative}), as well as some evolutionary algorithms like population-based training (Section \ref{sec:pbt}).
When this is the case, 
it may be sensible to consider hyperparameter schedules rather than fixed values. 
One of the most prominent examples in the literature are the ubiquitous learning rate schedules when training neural nets 
\citep{bengio2012practical, loshchilov2017sgdr, goyal2017accurate}.
In this section we overview such setup and briefly discuss how one may tackle the HPO problem as an instance of online optimization. 

Mimicking the notation of Section \ref{sec:gbhpo:iterative}, let us consider an algorithm $\learningalgo:\Lambda\to 
\hypothesisspace$ as a composition of iterates  $\Phi_t : \hypothesisspace \times \Lambda \rightarrow \hypothesisspace$. \footnote{
We disregard the dependence of the maps on data, for notational simplicity.
}
Such transitions maps need not be necessarily differentiable, although several methods that exploit the iterative view rely on this assumption.
This time, we let the hyperparameter configuration also vary as a function of the step. That is, given an initial model $h_0\in\hypothesisspace$:
\begin{equation}
    \label{eq:adv:scheudles}
    A(\lambda) = \Phi_T(\cdot, \lambda_T) \circ \Phi_{T - 1}(\cdot, \lambda_{T-1}) \circ  \dots \circ \Phi_1(h_0, \lambda_1),
\end{equation}
for some $T>0$, where we view $\lambda\in\Lambda^T$ as a hyperparameter sequence.
The iterations represented by the iterates $\Phi_t$ can correspond to different granularities or spans of operations within the training process, such as single stochastic gradient descent steps, entire epochs, or any other sequence of operations.

Hyperparameter iterates may be left ``free to float'', in which case one can formulate a related HPO problem following the general setup of Section \ref{sec:the_HPO_problem}, but this time searching for configurations in $\Lambda^T$ rather than in $\Lambda$. 
This is seldom done in practice due to the complications arising from an expanded search space, few exceptions being \citep{maclaurin2015gradient, wu2018learning}.
More commonly, the iterates are expressed as a function of the step $t$ and a set of other fixed hyperparameters $\lambda'\in \Lambda'$. 
Learning rate schedules are a primary example. For instance, a triangular schedule (i.e. warmup followed by linear decay) is given by:
\begin{equation*}
    \eta_t = 
    \left\lbrace
    \begin{array}{ll}
        \eta_{\mathrm{max}}(t/T_{\mathrm{max}}) 
        & \text{ for } \, t \geq T_{\mathrm{max}}  \\
        \eta_{\mathrm{max}} (1 - (t - T_{\mathrm{max}})/(T-T_{\mathrm{max}}))  & \text{ for } \, T_{\mathrm{max}} < t \leq T
    \end{array}
    \right.
\end{equation*}
where $\lambda'=(\eta_\mathrm{max}, T_{\mathrm{max}})$. 
Once again, when defining a related HPO problem, one may fall back to the setting of Section \ref{sec:the_HPO_problem}, optimizing over $\Lambda'$ rather than $\Lambda$. 
Along these lines, one may even let the hyperparameter iterates be a function of a state (e.g. the weights of the trainee model at step $t$), 
an approach that has been explored e.g. by \citet{hansen2016using} using reinforcement learning to fit a learning rate adjustment policy or \citet{lorraine2018stochastic} using hyper-networks and gradient-based HPO.
 
\paragraph{Online HPO.}
One main advantage of explicitly taking an iterative view is that it enable recasting the HPO problem as an instance of online optimization \citep{shalev2012online}, albeit a non-standard one.
Online HPO techniques attempt to speed up the HPO procedure by finding a performing hyperparameter schedule in one go, during a single execution of the learning algorithm.
A key difficulty is that one does not have access to the true objective $\valmetric$. 
Indeed, being able to compute (an unbiased estimate of)  $\valmetric$ requires terminating the learning algorithm, which would defeat the purpose of online HPO.
In this sense, these techniques do not -- or rather, cannot -- strictly solve Problem \eqref{eq:hpo0}-\eqref{eq:hpo1}, but rather rely on approximations, heuristics or surrogate objectives. 
This is akin to what we discussed in Chapter \ref{ch:multi_fidelity}, however multi-fidelity techniques only use early observations to take decisions or fit models of the response surface, and do not change hyperparameters during execution.  
Examples of online HPO include \citep{orabona2016coin, baydin_2017_online, luketina2016scalable, jaderberg2017population, lorenzo2017particle, mackay2019self, donini2019opt, vicol2021unbiased, micaelli2021gradient, chandra2022gradient}, with gradient-based techniques, and especially forward iterative differentiation, being proposed by various authors.

Another stream of recent research focuses on tuning hyperparameters of online learning algorithms, such as contextual bandit \citep{li2010contextual}
methods like UCB \citep{auer2002finite}
or Thomson Sampling \citep{agrawal2013thompson}.
In these algorithms, hyperparameters affect the agent’s exploration-exploitation trade-off and must adapt in real time. 
\citet{bouneffouf2020hyper} treat this as a nested bandit problem, 
tuning exploration parameters online via meta-bandits. \citet{ding2022syndicated} generalize this idea with Syndicated Bandits, 
a multi-layer optimization scheme that tunes several hyperparameters simultaneously,
with favorable regret guarantees. \citet{kang2024online} further extend this line by framing tuning as a continuum-armed bandit problem, enabling efficient search in continuous spaces. 
These methods depart from gradient-based HPO discussed above by tightly coupling hyperparameter adaptation with decision-time feedback,
offering principled and scalable alternatives tailored to the online nature of bandit learning.
Online HPO remains an area of active research.


\section{Neural Architecture Search}\label{sec:nas}

Deep learning has revolutionized the traditional approach of hand-crafting features by enabling automatic feature learning during training.  
However, the design of neural network architectures still relies heavily on manual effort, involving substantial trial and error.

Following the problem definition described in Section~\ref{sec:the_HPO_problem}, we can extend the idea of hyperparameter optimization (HPO) to automate the design of neural architectures --- a process known as neural architecture search (NAS) (see also \citet{white-arxiv23, elsken2019neural} for an overview).  
NAS is closely related to HPO but focuses specifically on hyperparameters that define the architecture. At the same time, NAS introduces unique challenges not typically encountered in standard HPO tasks. Its search spaces tend to be higher-dimensional, entirely discrete, and often exhibit highly conditional structures (e.g., the number of units in a layer is only relevant if that layer exists, as determined by another hyperparameter).  
Furthermore, the high computational cost of training modern deep learning models makes NAS especially expensive and calls for even more efficient optimization techniques.

\subsection{Search Space}
A key question in NAS is how to define a search space to parameterize a neural network architecture. 
An expressive search space might lead to entirely new and innovate architecture but might be also prohibitively expensive to explore.
On the other hand, a too restrictive space will lower the chance to improve upon manually design architectures.
We discuss two popular approaches to define NAS search spaces~\citep{white-arxiv23}: Macro Search Spaces and Cell Search Spaces.

\paragraph{Macro search spaces} define the design of neural network architectures at a high level, focusing on configurable choices across the overall structure of the model rather than fine-grained details.
Instead of searching over every possible low-level configuration, they assume a fixed architectural topology --- such as a residual network or transformer --- and explore how the major components within that structure can be customized.

These design decisions might include, for instance, whether to apply layer normalization before or after an operation, how many attention heads to use in each transformer layer, or what type of convolution is used in different stages of a CNN. 
The goal is to balance architectural flexibility with enough structure to keep the search space manageable.
For example, MNASNet \citep{tan-cvpr19} divides the architecture into blocks and tunes high-level choices like convolution type and kernel size per block.

\paragraph{Cell search spaces} focus on smaller computational units  ---  called cells  ---  which are repeatedly stacked to form the full network.
Unlike macro search spaces that configure existing architectures, cell-based approaches define a compact building block whose internal structure is optimized and then reused throughout the model allowing to find new topologies.

Typically, each cell is represented as a directed acyclic graph (DAG) where nodes correspond to intermediate feature representations and edges denote operations (e.g., convolutions, pooling).
The search algorithm then determines the operations and how they are connected within the cell.
This abstraction significantly reduces the size of the search space while retaining enough flexibility to discover high-performing architectures.

For instance, \citet{zoph:2018:learning_transferable} introduced a cell-based NAS approach where each node in the DAG is assigned a specific operation, such as a 1x1 or 3x3 convolution.
To manage complexity and enable spatial downsampling, they distinguished between two types of cells: normal cells, which preserve input dimensions, and reduction cells, which downsample feature maps using larger strides.

\subsection{Neural Architecture Search as Hyperparameter Optimization}

We can phrase NAS as a special type of hyperparameter optimization problem $\min_{\lambda\in\Lambda} f(\lambda)$ (see also Section \ref{sec:the_HPO_problem}), where the inputs $\lambda$ define the architecture and $f(\lambda)$ represents the validation error after training the corresponding network.

While early work on automatically discovering neural network architectures dates back some time (for example~\citep{kenneth-hyperneat}), \citet{bergstra_making_2013} were the first to tackle this problem definition using TPE (see Section~\ref{sec:mbm-tpe}) to automatically design architectures for convolutional neural networks.  
Arguably, \citet{zoph2016neural} coined the term neural architecture search.
In their pioneering work, they proposed a controller based on a recurrent neural network to suggest new architectures for convolutional and recurrent neural networks.  
In each iteration, a set of candidate architectures is sampled and evaluated to collect observations, which are then used to update the controller using reinforcement learning.  
While they were able to discover architectures that outperformed the state-of-the-art at the time, their proposed method was prohibitively expensive, requiring up to 800 GPUs concurrently.
\citet{real2019regularized} proposed an evolutionary approach that maintains a population of architectures (see Chapter \ref{ch:popM}).  
New candidates are sampled uniformly at random from the population and mutated along a single dimension.  
Afterward, the population is updated by removing the last entry.

Inspired by the success of Bayesian optimization in HPO, some NAS approaches proposed Bayesian optimization variants to tackle the often high-dimensional and structured search spaces of NAS.  
\citet{kandasamy-neurips18} proposed a Gaussian process variant of Bayesian optimization that measures similarities across architectures using optimal transport approaches.  
\citet{ru-iclr21} argued for combining a Weisfeiler-Lehman graph kernel with a Gaussian process surrogate model instead, which allows for better interpretability of the final architectures.  
Similar to HPO, these Bayesian optimization approaches usually tend to be faster in terms of convergence speed than evolutionary algorithms such as regularized evolution~\citep{ying-icml19,ru-iclr21}.

While these approaches led to architectures outperforming state-of-the-art models at the time, the original methods required substantial computational resources since each network was trained from scratch.  
Nevertheless, these black-box approaches can be quite effective for moderately sized neural networks and smaller search spaces. Furthermore, they can be easily extended to a multi-objective setting, for example, to simultaneously optimize energy consumption and validation performance.
Building on recent work on HPO to accelerate the search process, in this setting we can also use more sophisticated approaches, such as multi-fidelity optimization (see Section~\ref{ch:multi_fidelity}) or transfer learning~\citep{feurer2015initializing,wistuba-ieee15,salinas-icml20}.  
However, due to the ever-growing size of these networks, even these approaches quickly become infeasible.

\subsection{Weight-Sharing Based Neural Architecture Search}\label{subsec:ws_nas}

A more radically different approach to accelerate NAS is the so-called weight-sharing NAS.  
The idea is to define a single super-network that contains all possible architectures in the finite search space.  
Each path through the super-network represents a single architecture.  
In the original work by \citet{pham-icml18}, this method accelerated the search by more than a factor of 1000.

In a similar spirit to gradient-based HPO (see Chapter~\ref{ch:gradient-based}), \citet{liu2018darts} defined continuous auxiliary parameters for each architectural choice, which are learned by computing the gradient of the validation loss.
This resembles the bi-level optimization problem described in Section~\ref{sec:the_HPO_problem}; however, here the outer objective optimizes the auxiliary architectural parameters instead of other hyperparameters, and the inner objective optimizes the parameters of the neural network.
To reduce computational overhead, \citet{liu2018darts} proposed DARTS --- a gradient-based NAS method --- which alternates after each step between the inner and outer objectives. 
After training the super-network, the best architecture is selected based on the shared weights and then re-trained from scratch.  
However, several papers \citep{li-uai20,yang-iclr20} reported that this formulation heavily relies on the search space and does not yield better results than just randomly sampling architectures.

Two-stage weight-sharing NAS~\citep{cai-iclr20,yu-eccv20} first trains the super-network by updating individual subnetworks in each iteration.  
After training the super-network, we compute the validation error $f(\lambda)$ by performing a single pass over the validation data.  
This substantially reduces the computational cost and allows us to apply hyperparameter optimization techniques, such as Bayesian optimization (see Chapter~\ref{ch:mbm}) or evolutionary algorithms.  
For example, \citet{cai-iclr20} proposed the once-for-all model, which provides a single super-network that allows selecting the optimal architecture with weights for a given hardware, minimizing both latency on the target device and validation error.

Compared to black-box approaches, weight-sharing methods can be applied to higher-dimensional search spaces and neural networks with a larger parameter count. However, they require more substantial changes to the neural network implementation and therefore cannot be straightforwardly applied to existing codebases. Additionally, they are not easily applicable to more sophisticated cell search spaces defined using a context-free grammar~\citep{ericsson-neurips24}.
While weight-sharing based methods are popular in the NAS literature, it is not straightforward to apply these techniques to HPO problems, because they only control the capacity of the network and not the training process itself.






\section{Meta-Learning}
\label{sec:meta-learning}
We discuss here meta-learning since, from a certain point of view, it shares
some interesting similarities to HPO. This section only focuses on this
point of view, in the case of supervised learning. For a more general
overview of meta-learning, see for example the survey by
\citet{hospedales:2021:metalearning_neural_networks} and references
therein.


\subsection{Definition}
\begin{figure}
  \centering
  \includegraphics[width=0.89\textwidth]{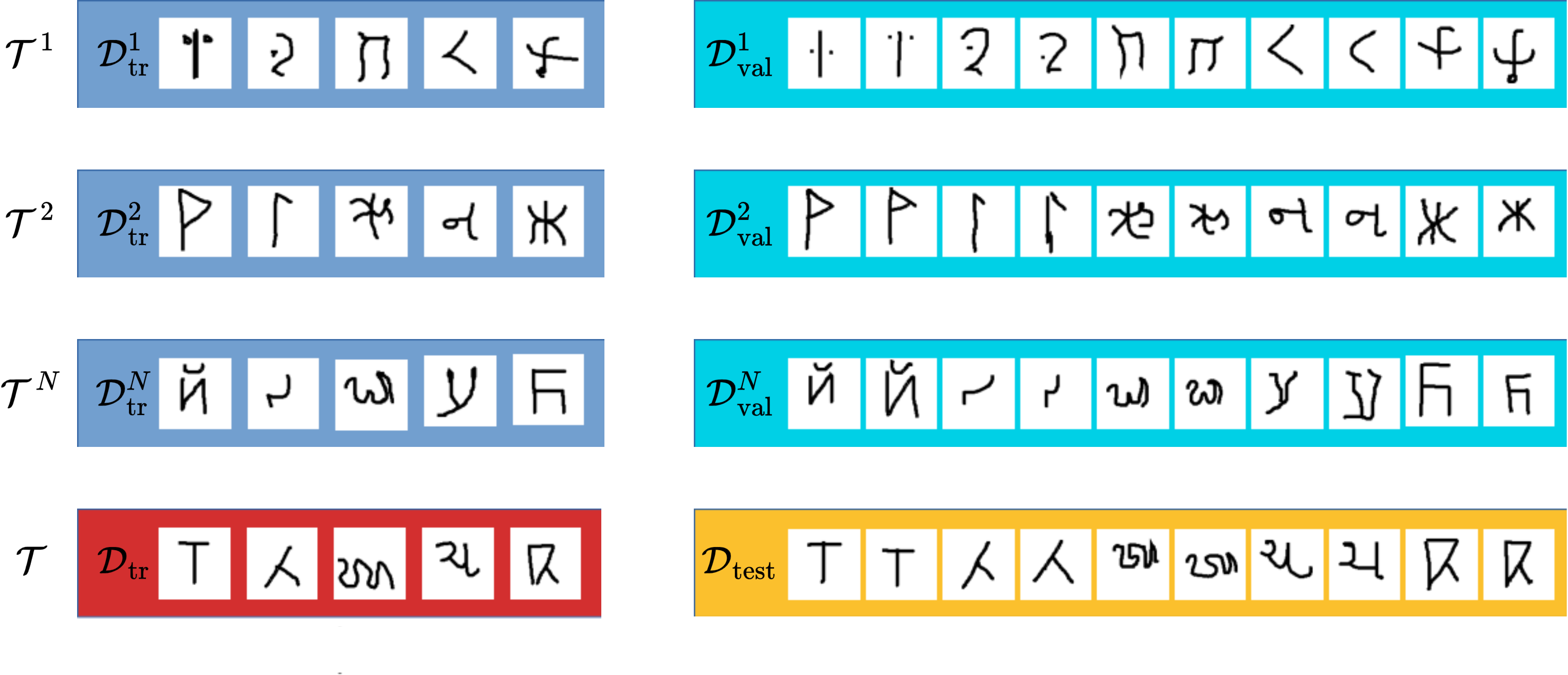}
  \caption{Tasks and datasets for meta-learning on
    Omniglot~\citep{lake:2019:omniglot_challenge_3year}. At the end of
    the meta-training phase, a small training set for the new task
    $\mathcal{T}$ is provided and performance is evaluated on test
    data $\mathcal{D}_{\text{test}}$. A similar figure was shown
    in~\citep{ravi_optimization_2016} for tasks derived from
    Imagenet.}
  \label{fig:meta}
\end{figure}

In meta-learning (also called learning-to-learn), we have a sequence of tasks
$\mathcal{T}^1,\mathcal{T}^2,\dots,\mathcal{T}^N$, where each task
$\mathcal{T}^j=(p^j,\taskloss^j)$ consists of a distribution $p^j(x,y)$,
$x\in\mathcal{X}$, $y\in\mathcal{Y}^j$ over data points and a loss
function $\taskloss^j$. Note that all tasks share the same input space, but
their label spaces may differ.  Tasks are sampled from a meta
distribution $\mathcal{P}$ over tasks. Unlike a traditional
supervised algorithm (which, given a dataset, will produce a
prediction function $h$), a meta-learning algorithm outputs a learning
algorithm that is hoped to learn a good prediction function when
presented with a new (unseen) task $\mathcal{T}=(p,\taskloss)$ sampled from
the same meta distribution $\mathcal{P}$. The meta-training phase
consists in computing the learning algorithm
$A = ML(\mathcal{D}^1,\dots,\mathcal{D}^N)$ where
$\mathcal{D}^j\sim p^j$ and $ML$ is the meta-learning algorithm. The
meta-test phase, in turn, consists in computing the prediction function using
the new task: $h=A(\mathcal{D})$, with $\mathcal{D}\sim p$.
The concept is illustrated in Figure \ref{fig:meta}.

Meta-learning has its own specificity when compared to other settings
dealing with several tasks. Compared to multitask
learning~\citep{Caruana94,Baxter2000,ruder:2017:overview_multitask_learning},
the set of tasks is not fixed in advance. Compared to transfer learning or
domain
adaptation~\citep{pan2009survey,torrey2010transfer,kouw2018introduction},
there are several ``source'' tasks. Typically, we are interested in the
scenario where a very small dataset $\mathcal{D}~\sim p$ will be available
for the new task (few-shot learning) or even a single example per class
(one-shot learning). The main insight is that when the tasks are related, as
captured by the distribution $\mathcal{P}$, meta-learning reduces the sample
complexity of learning new tasks drawn by the same distribution
\citep{Baxter2000,maurer2016benefit}. Note that while HPO typically involves
the optimization of a ``black-box'' function, both the meta-training and the
meta-test phases in meta-learning more usually involve the optimization of
functions with a know structure, enabling the application of gradient-based
techniques, as discussed below.

\subsection{Meta-learning of Common Representations}
%
One effective approach to meta-learning consists of using past tasks to
build strong representations, which can be later used for few-shot
learning on the new task. We briefly discuss this approach here, as it is the
most closely related to the HPO problem~\citep{franceschi2018bilevel}.
As in~\citep{Baxter2000}, let us introduce the family of hypothesis
spaces $\mathbb{H}^j=\{\mathcal{H}^j\}$ where each
$\mathcal{H}^j\in\mathbb{H}^j$ is a set of functions
$h^j_{w,\lambda}:\mathcal{X}\mapsto\mathcal{Y}^j$ of the form
$h^j_{w,\lambda}=\gamma_w^j\circ\Phi_\lambda$, i.e., obtained by
composing a shared feature learning function
$\Phi_\lambda:\mathcal{X}\mapsto\mathcal{F}$, parameterized by
$\lambda$ (also called meta-parameter), with a task-specific function
$\gamma_w^j:\mathcal{F}\mapsto\mathcal{Y}_j$, parameterized by $w$.
The set $\mathcal{F}$ is a suitably chosen feature space, for example
$\mathcal{F}=\RSet^d$. This type of function composition is
common in several deep learning architectures where $\Phi$ is in
charge of representation learning while $\gamma$ could be a
classification or regression head. The multitask network introduced
in~\citep{Caruana94} and the feature learning neural network described
in~\citep{Baxter2000} are structured in that way, although all the
weights $\lambda$ and $w$ are optimized jointly, unlike the problem
formulated below.
Note that we have deliberately used the
notation $\lambda$ for the parameters shared across the tasks, to
highlight the fact that they play a role similar to hyperparameters.
In order to construct a ``response function,''
for every source task $\mathcal{T}^j$, $j=1,\dots N$, we split the
given dataset $\mathcal{D}^j$ into a training set
$\mathcal{D}_{\text{tr}}^j$ and a validation set
$\mathcal{D}_{\text{val}}^j$ and define
$$
f(\lambda) = \sum_{j=1}^N \taskloss^j\left(w^j,\lambda,\mathcal{D}^j_{\text{val}}\right).
$$
We are now ready to formulate the following bilevel program
\begin{eqnarray}
  \label{eq:meta_bilevel}
  \min_{\lambda} &~& f(\lambda)\\
  \text{s.t.} &~& w^j \in \text{arg}\min_u \taskloss^j\left(w^j,\lambda,\mathcal{D}^j_{\text{tr}}\right)
\end{eqnarray}
where the lower level (or inner) objectives $\taskloss^j$ are empirical risk
minimization on each training tasks, and the upper (or outer) objective is the validation error on the same tasks. 
If $\Phi$ and $\gamma$ are realized by deep learning modules, the
inner problem would be solved by some iterative gradient-based
algorithm, implying that the solution would only be found
approximately. Moreover, in general, the above problem is not
guaranteed to have a solution. In~\citep{franceschi2018bilevel} it is
shown that a solution is guaranteed to exist if $\gamma$ is linear
(i.e., the classification head consists of a single dense layer) and
an algorithm based on hypergradients (using the method described in~
Section~\ref{sec:gbhpo:iterative}) is presented to solve the
meta-learning problem. A closely related algorithm is proposed
in~\citep{finn_model-agnostic_2017}, although it is not explicitly
formulated as a bilevel program and the emphasis is on fine-tuning to
the target task an architecture pretrained on the sequence of previous
tasks. A solution based on implicit gradients (see
Section~\ref{sec:implicit_hypergradients}) is offered
in~\citep{rajeswaran2019meta}. 

Lastly, while we presented meta-learning as shared feature learning, we note that other tasks' sharing knowledge mechanisms are possible. Notable examples are biased regularization \citep{denevi2018learning,denevi2019learning,balcan2019provable}, in which the meta-parameter $\lambda$ describes a common mean among the tasks, hyper-network approaches, in which $\lambda$ is associated to a function that outputs the parameters of a task-specific model \citep{zhao2020meta,zakerinia2024more}, and conditional meta-learning \citep{denevi2020advantage,denevi2022conditional,rusu2018meta,wang2020structured}, in which $\lambda$ is a conditioning function of side information specific to the task at hand.


\section{Transfer of Hyperparameters Across Model Scale}
\label{sec:hp-transfer-scale}

A foundation model~\citep{brown2020language}, such as a large language model, is a large-scale, pre-trained machine learning model, typically based on deep learning architectures like transformers~\citep{vaswani_attention_2017}, that serves as a general-purpose backbone for various downstream tasks through fine-tuning or prompting.
These models mark a cornerstone in artificial intelligence and have led to several breakthroughs in code generation, natural language processing, and common-sense reasoning.
However, the expensive pre-training phase requires vast amounts of data and tremendous GPU resources.

While this 
also depends on a careful selection of hyperparameters, the hyperparameter optimization methods presented here are not directly applicable.
Even efficient state-of-the-art multi-fidelity approaches (see Chapter~\ref{ch:multi_fidelity}) require at least a few training runs on the full budget, but for contemporary LLMs, this is infeasible.

Inspired by a theoretical perspective from tensor programs, a recent line of work~\citep{yang-arxiv22,everett-icml24a} approaches this problem from a different angle.
While hyperparameters of neural networks, such as the learning rate, change when we increase the width or depth of the network and need to be adjusted accordingly,  the idea here is to change the parameterization of neural networks, such that the optimal hyperparameters stay the same when scaling from smaller models to larger models.
This allows for finding the optimal hyperparameters on a small scale and directly transferring them to the largest scale.

The original parameterization proposed by \citep{yang-arxiv22}, called $\mu$-parameterization, initially provided only a theoretical foundation for scaling model width, however they also provided empirical results confirming its effectiveness for depth scaling.
Recent work by \citep{everett-icml24a} has shown that even hyperparameters such as the learning rate can be transferred across different model widths using the standard parameterization of neural networks by adapting them on a per-layer basis.

While this line of research has shown promising results, it has so far been mostly theoretical and primarily focused on width scaling.
Nevertheless, it has the potential to lead to more practical hyperparameter optimization methods that scale efficiently to large foundation models.

 \chapter{Hyperparameter Optimization Systems}
\label{ch:hposys}

In this chapter, we provide an overview of different open-source libraries and commercial systems that implement methods discussed in the previous chapters. 
HPO software landscape changes frequently as new problems, methods and application domains become relevant:
 the provided list here is by no means exhaustive and only represents a snapshot. 

\section{Open Source Libraries and Frameworks}

The Python package Spearmint
\citep{snoek_practical_2012} is one of the first libraries that provides Bayesian optimization for
hyperparameter optimization.
It implements Bayesian optimization using
Gaussian processes (see Section \ref{sec:bk:hpo:bayesopt}). At the same
time, \citet{bergstra2011algorithms} provided HyperOpt, offering Bayesian
hyperparameter optimization by using the tree Parzen estimator (Section \ref{sec:tpe}) instead of
Gaussian processes (Section \ref{sec:gp}). \citet{Hutter2011:Sequential} introduced SMAC, which
implemented Bayesian optimization strategies using Random Forests.
Initially targeting at more generic algorithm configuration problems,
such as configuring SAT solvers, it also offered functionality for HPO of
machine learning methods.
The current version~\citep{smac3} is written in Python and also provides more
advanced functionalities, such as multi-fidelity optimization (see Chapter
\ref{ch:multi_fidelity}).

Among the first Python libraries that implemented multi-fidelity HPO, such as Fabolas~\citep{klein2017fast} or Hyperband~\citep{li2016hyperband}, was the RoBO library by \citet{klein-bayesopt17} and Dragonfly by \citet{kandasamy-jmlr19}.

While previous libraries, such as Spearmint, allowed to parallelize trials across processes on the same machine, more recent libraries focus more on a large-scale distributed setting to distribute the evaluation of trials across multiple instances or compute nodes.
For example, RayTune~\citep{liaw2018tune}, distributes the evaluation of hyperparameter configurations using the compute framework Ray.
Another example is Orion, which supports both high-performance computing infrastructures and containerized environments.
\citet{salinas-automl22} recently presented Syne Tune, which supports different backends that allow to distribute the HPO process either locally or on the cloud. Furthermore, it also provides a backend to simulate the evaluation of expensive objective functions using surrogate models (see Section~\ref{sec:surrogate}). 
This allows to simulate large-scale asynchronous HPO on a single compute core.

Recent PyTorch implementations of gradient-based hyperparameter optimization algorithms include the libraries HyperTorch~\citep{grazzi2020iteration}, higher~\citep{grefenstette2019generalized} and Betty~\citep{choe2023betty}.
JaxOpt~\citep{blondel2022efficient} is a Jax implementation that focuses on implicit differentiation.

Most HPO libraries or frameworks expect the objective function either as function handle, for example in Python, or as a single training script. The search space is usually defined outside the actual training script and hyperparameters are passed as input arguments. 
Optuna~\citep{akiba-sigkdd19} follows a different approach by a so-called define-by-run API that allows users to dynamically define the search space instead of providing it in advance.

\section{Commercial Systems}

Since hyperparameter optimization plays such an integral role in the machine learning workflow, most machine learning platforms, for example \textit{Weights and Biases}, provide some functionality to select hyperparameters.
There are also several commercial systems specific for automated hyperparameter optimization. For example, Google Vizier~\citep{google_vizier} is an internal service that contributed to several Google products and research efforts. \citet{oss_vizier} also introduces an open-source version of Google Vizier as a stand-alone Python package. 
AWS SageMaker provides a fully automated service called Automated Model Tuner~\citep{perrone2021sagemaker} for gradient-free optimization problems, which allows for large-scale HPO leveraging built-in SageMaker functionalities.
Another example is Sigopt, which provides optimization-as-a-service for hyperparameter optimization utilizing Bayesian optimization techniques.

\section{Benchmarking}\label{sec:surrogate}

A key challenge for an empirical driven research field such as HPO to make progress is the thorough evaluation of new approaches and comparison to baseline methods. However, due to the expensive nature of the underlying optimization problem, it is often prohibitively expensive to obtain statistically reliable results including multiple repetitions, which are necessary to characterize HPO algorithms that often come with intrinsic randomness.

To reduce the computational requirements and enable large-scale experimentation, recent work tries to emulate expensive HPO problems by using cheap-to-evaluate approximations of the underlying objective function. \citet{eggensperger-aaai15} introduced the idea of surrogate benchmarks that replace the original benchmark with the predictions of a machine learning model. This model is first trained in an offline step on data collected from the original benchmark. Unlike the probabilistic model used in Bayesian optimization, where data is generated iteratively, all data in this approach is generated simultaneously, and the model is trained only once. While using a machine learning model to approximate the objective function introduces potentially some bias, single function evaluations are now orders of magnitude cheaper, since they only require a single prediction.

Instead of using a machine learning model to approximate the true objective, \citet{klein-arxiv19a} first discretized the search space of a fully-connected neural network leading to a finite number of hyperparameter configurations and then performed an exhaustive search to collect the performance and runtime of each configuration. This allows to replace the actual training and validation of a hyperparameter configuration by simple table look-ups during the HPO process. To mimic the observation noise that comes with the HPO of neural networks, each HPO configuration was evaluated multiple times. During the HPO process, random entries for each hyperparameter configuration are returned to simulate the observation noise. 
The idea of tabularized benchmarks has since been extended to neural architecture search \citep{ying-icml19, dong-iclr20} (see Section \ref{sec:nas}) leading to several new benchmarks. 
While tabularized benchmarks return for each query truly observed values of the objective function instead of predictions of the model, \citet{pfisterer-automl22} argued that the discretization of the search space eliminates the local structure of the objective function, leading actually to a higher bias than using surrogate models.
The idea of surrogate benchmarks has been extended to define a generative model by \citet{klein-nips19}, which allows sampling an arbitrary set of new benchmarks.

To standardize the interface to different benchmarks, \citet{eggensperger_towards_2013} introduced HPOLib, which consolidates different benchmarks from the literature. In follow-up work \citep{eggensperger-neurips21}, HPOBench containerized these benchmarks to increase reproducibility of benchmarking results.
HPO-B \citep{pineda-neurips21} provides tablularized benchmarks based on OpenML datasets~\citep{vanschoren-sigkdd13a}. In a similar vein, \citet{zimmer-ieee21} introduced LCBench, which provides full learning curves of neural network models trained on OpenML datasets.
\citet{salinas-automl22} also provided a unified interface to tabular and surrogate benchmarks in Syne-Tune, using a fast data format to accelerate the table look-ups.
Another suite for surrogate-based benchmarking is YAHPO \citep{pfisterer-automl22}, consisting of different scenarios, including combined hyperparameter optimization and algorithm selection.

\chapter{Research Directions}
\label{ch:open_questions}

By now, hyperparameter optimization 
may be considered a mature field with well-established algorithmic
solutions and with industry-level software frameworks.
 It is a crucial
ingredient for automated machine learning, an active research area that has even
started a conference in 2022. 
In this monograph, we have reviewed in depth the
major classes of methods (random and quasi-random search, model-based, multi-fidelity, population-based, and
gradient-based) and highlighted connections to closely related areas
such neural architecture search and meta-learning. 

Yet, in all classes of algorithms there remain open issues and
interesting directions for improvement. 
In addition, the development of hybrid algorithms has so far been under-explored, but could potentially bring notable benefits in practical applications. 
We do not discuss these particular aspects here, but rather 
touch upon some general topics which we
feel may become (or rather, are already becoming) 
broad and important research directions in HPO.

\section{Response Functions for Unsupervised Learning}

HPO requires a well-defined \textit{response function} to be minimized (see
Section~\ref{sec:the_HPO_problem}), which in turn requires a well-defined
performance metric. While suitable metrics are abundant for supervised learning,
the same is not true for unsupervised learning. 

A notable example is \textit{anomaly detection}, an area rich of methods,
results, and applications~\citep{pang:2021:deep_learning_anomaly}. The
anomaly or novelty detection problem essentially consists of identifying the
high probability region of the distribution behind the data, so that
points outside this region can be detected as outliers.
For this problem,~\cite{ma:2023:need_unsupervised_outlier} survey a
number of internal (i.e., not using any ground truth labels) approaches for
assessing outlier detection models and conclude that ``none of the existing and
adapted strategies would be practically useful.''

A second important example is generative models. In this case the goal is to
construct a model of the data distribution such that sampling from that
model produces fake data points that are indistinguishable from real data.
Again, the literature is rich in terms of models, ranging from early
generative adversarial
networks~\citep{DBLP:journals/corr/GoodfellowPMXWOCB14} to normalizing
flows~\citep{pmlr-v37-rezende15} and more recently probabilistic diffusion
models~\citep{ho:2020:denoising_diffusion_probabilistic}. In the particular
case of images, popular metrics used to compare different generative models
are the Inception Score~\citep{NIPS2016_8a3363ab} and the Fréchet inception
distance (FID)~\citep{DBLP:conf/nips/HeuselRUNH17}. These metrics only
capture one aspect of the generative process, namely fidelity. However, the
problem of assessing the quality of a model is inherently more complicated:
it is not sufficient to generate faithful points, we also need to cover all
regions where the data distribution is high.

There have been interesting attempts to interpret this perspective in terms
of precision and
recall~\citep{sajjadi:2018:assessing_generative_models,simon:2019:revisiting_precision_recall},
where the two supports of real and fake data are estimated and compared.
Fake points off the support of real distribution are interpreted as false
positives, and real points outside the support of the model's distribution
are interpreted as false negatives. The idea has sparkled a number of
practically computable
metrics~\citep{kynkaanniemi:2019:improved_precision_recall,%
  naeem:2020:reliable_fidelity_diversitya,%
  park:2023:probabilistic_precision_recall} that however have not been shown
to be effective to guide HPO. A unidimensional metric that aims to assess
the similarity between the model and the real data distributions is the
\textit{overlapping area}. Computing the overlaps between these two
distributions requires reliable density estimation, which is however
difficult for high dimensional data. A simple trick to overcome the
associated curse of dimensionality issue is to consider instead the
distribution of intra-set distances (between pairs of real data points) and
inter-set distances (between pairs of fake points). This metric was
introduced by~\citet{yang:2020:evaluation_generative_models} for assessing
the quality of generative models in symbolic music, but distances in that
case are based on domain specific aspects of music such as rhythm or
harmony, and cannot be easily extended to other types of data. A closely
related metric based on maximum mean
discrepancy~\citep{obray:2021:evaluation_metrics_graph}, popular in the
context of graph generative models, has been shown to be complicated due to
its strong sensitivity to hyperparameters in the definition of the metric
itself, leading to a sort of hyper-hyperparameters.

A notable exception in the realm of unsupervised learning is the case of
obtaining disentangled representations, which is the problem of identifying a
small number of explanatory factors behind observed
data~\citep{bengio:2013:representation_learning_reviewa,%
  locatello:2019:challenging_common_assumptions}, such as shape, color and
position for image classification problems. For this particular case, evaluation
metrics have been developed based on informativeness, separability and
inference, and interpretability~\citep{do:2019:theory_evaluation_metrics}. For
this particular problem, a hyperparameter tuning algorithm was developed
in~\citep{duan:2019:unsupervised_model_selection} using ranking scores. Still,
the development of general purpose HPO algorithms for unsupervised learning
remains an open problem.

\section{Learning to Optimize Hyperparameter}

In Section~\ref{sec:meta-learning}, we covered connections between HPO and meta-learning from a problem-setting point of view,
highlighting the connections between meta-parameters of a meta-learning algorithm and hyperparameters of a standard algorithm.
In addition, 
one may also think of meta-learning a hyperparameter optimization algorithm, in that the task distribution is made of HPO problems, each described by a distribution over datasets and response functions of one or several standard learning algorithms.
Samples from such a task distribution may involve various datasets and learning algorithms and traces of evaluations of hyperparameter configurations and their response value (e.g., the validation error).
The goal of the meta-learner in this context would be to learn a general strategy for efficiently searching the hyperparameter space based on
data collected while optimizing hyperparameters across many different problems.

Early attempts use meta-data to warm start model-based methods, or learn loss curves; see \citep[][Chapter 2]{hutter2019automated} for a survey. 
However, more recent works take this perspective one step further, motivated by successes of transformer-based language modeling and learning to optimize, a sub-field of meta-learning \citep{chen2022learning}.
For example, \cite{chen2022towards} proposed using a large language model to directly generate promising hyperparameter configurations based on a natural language description of the dataset and learning algorithm. The language model is fine-tuned on a corpus of hyperparameter optimization traces, allowing it to capture complex relationships between hyperparameters and adapt its search strategy to the problem at hand.

The usage of rich meta-data, past experience, and the capacity to leverage
 similarities across standard learning algorithms may hold the promise of discovering HPO algorithm that are more efficient than classic approaches.
However, significant challenges remain in developing truly general and robust meta-learning for HPO, including handling the diversity of problem types, scalability to high-dimensional spaces, and transferability across vastly different domains and algorithms.

\section{HPO and Foundation Models}

The emergence of foundation models \citep{bommasani2021opportunities} as a new frontier for machine learning applications and research opens up a series of novel scenarios and challenges for hyperparameter optimization. 

Zero-shot \citep{wei2021finetuned} and in-context learning \citep{brown2020language} have become common paradigms for utilizing large language models (LLM), both of which require choosing context (or prompt) templates and in-context examples.
LLMs' performances at downstream tasks may be sensitive to such choices. 
These choices are often carried 
out manually by inspecting few rounds of interactions (prompt engineering).
However, both problems can be formulated as hyperparameter optimization tasks, whereby the hyperparameters may be discrete choices over in-context examples or prompt templates, or even entire sequences over tokens.  
The resulting search space may be large and invariably discrete, and the response functions often involve several potentially expensive and potentially noisy calls to the LLM, which render the resulting HPO problems particularly challenging. 
Initial attempts include AutoPrompt
\citep{shin2020autoprompt}, a gradient-based approach (see Chapter~\ref{ch:gradient-based}) that optimizes for discrete ``trigger'' tokens;
Self-Adaptive In-Context Learning \citep{wu2022self}, which proposes a selection-then-rank framework for selecting instance-dependent examples; TRIPLE \citep{shi2024efficient} which tackles template optimization under a best arm identification angle (algorithms proposed include successive halving; see Section~\ref{sec:sh}); HbBoPs \citep{schneider2024hyperband}, which leverages a structure-aware Gaussian Process surrogate combined with Hyperband to efficiently select prompts in a black-box LLM setting;
Black-box Discrete Prompt Learning \citep{diao2022black} which uses a policy gradient estimator to optimize the prompt templates, among others.
Further research is needed to study more efficient and scalable methods for optimizing prompts and in-context examples for large language models. 
Advances in reinforcement learning and gradient-based approaches with discrete gradient estimators (see Section \ref{sec:gbhpo:discrete}) may prove useful in these directions.

In addition, the emergence of agents and agent networks where various LLMs interact and act by accessing to external tools \citep{schick2024toolformer, yang2024swe} provides another compelling application ground for HPO, in particular considering that many applications leverage closed-source LLMs queried through APIs. 
These agent-based systems introduce new hyperparameters to optimize, such as the selection and configuration of external tools available to the agents, including documentation, output formatting and error reporting, as well as 
the communication protocols and the interactions graph between agents --- see, e.g., GPTSwarm \citep{zhuge2024language}.
Optimizing these hyperparameters presents unique challenges due to the complex interactions between agents, tools, and the underlying LLMs. Traditional HPO methods may need to be adapted or extended to handle the dynamic and potentially non-stationary nature of these systems. Furthermore, the use of closed-source LLMs via APIs introduces additional constraints, such as rate limits and usage costs, which must be considered in the optimization process.

\begin{acknowledgements}
  The work of PF and MP was partially supported by the European Union
  NextGenerationEU and the Italian National Recovery and Resilience Plan
  through the Ministry of University and Research (MUR), under Projects
  CAI4DSA (CUP B13C23005640006) and PE0000013 (CUP J53C22003010006),
  respectively.
\end{acknowledgements}

\backmatter

\printbibliography%

@article{friedler2016possibility,
  title={On the (im)possibility of fairness},
  author={Friedler, S. A. and Scheidegger, C. and Venkatasubramanian, S.},
  journal={arXiv preprint arXiv:1609.07236},
  year={2016}
}

@book{dempe2002foundations,
  title={Foundations of bilevel programming},
  author={Dempe, Stephan},
  year={2002},
  publisher={Springer Science \& Business Media}
}

@article{denevi2018learning,
  title={Learning to learn around a common mean},
  author={Denevi, Giulia and Ciliberto, Carlo and Stamos, Dimitris and Pontil, Massimiliano},
  journal={Advances in neural information processing systems},
  volume={31},
  year={2018}
}

@inproceedings{balcan2019provable,
  title={Provable Guarantees for Gradient-Based Meta-Learning},
  author={Balcan, Maria-Florina and Khodak, Mikhail and Talwalkar, Ameet},
  booktitle={International Conference on Machine Learning},
  pages={424--433},
  year={2019}
}

@article{denevi2020advantage,
  title={The advantage of conditional meta-learning for biased regularization and fine tuning},
  author={Denevi, Giulia and Pontil, Massimiliano and Ciliberto, Carlo},
  journal={Advances in Neural Information Processing Systems},
  volume={33},
  pages={964--974},
  year={2020}
}

@article{denevi2022conditional,
  title={Conditional meta-learning of linear representations},
  author={Denevi, Giulia and Pontil, Massimiliano and Ciliberto, Carlo},
  journal={Advances in Neural Information Processing Systems},
  volume={35},
  pages={253--266},
  year={2022}
}

@article{rusu2018meta,
  title={Meta-learning with latent embedding optimization},
  author={Rusu, Andrei A and Rao, Dushyant and Sygnowski, Jakub and Vinyals, Oriol and Pascanu, Razvan and Osindero, Simon and Hadsell, Raia},
  journal={arXiv preprint arXiv:1807.05960},
  year={2018}
}

@article{wang2020structured,
  title={A structured prediction approach for conditional meta-learning},
  author={Wang, Ruohan and Demiris, Yiannis and Ciliberto, Carlo},
  journal={Advances in Neural Information Processing Systems},
  year={2020}
}

@article{zakerinia2024more,
  title={More Flexible PAC-Bayesian Meta-Learning by Learning Learning Algorithms},
  author={Zakerinia, Hossein and Behjati, Amin and Lampert, Christoph H},
  journal={arXiv preprint arXiv:2402.04054},
  year={2024}
}

@inproceedings{zhao2020meta,
  title={Meta-learning via hypernetworks},
  author={Zhao, Dominic and Kobayashi, Seijin and Sacramento, Jo{\~a}o and von Oswald, Johannes},
  booktitle={4th Workshop on Meta-Learning at NeurIPS 2020 (MetaLearn 2020)},
  year={2020},
  organization={NeurIPS}
}

@incollection{dempe2020bilevel,
  title={Bilevel optimization},
  author={Dempe, Stephan and Zemkoho, Alain},
  booktitle={Springer optimization and its applications},
  volume={161},
  year={2020},
  publisher={Springer}
}

@article{maurer2016benefit,
  title={The benefit of multitask representation learning},
  author={Maurer, Andreas and Pontil, Massimiliano and Romera-Paredes, Bernardino},
  journal={Journal of Machine Learning Research},
  volume={17},
  number={81},
  pages={1--32},
  year={2016}
}

@article{fairness2018verma,
  title={Fairness definitions explained},
  author={Verma, S. and Rubin, J.},
  journal={FairWare},
  year={2018}
}

@article{kleinberg2018inherent,
  title={Inherent trade-offs in algorithmic fairness},
  author={Kleinberg, J.},
  journal={SIGMETRICS},
  year={2018}
}

@article{Mockus1978:Application,
  title={The application of Bayesian methods for seeking the extremum},
  author={Mockus, Jonas and Tiesis, Vytautas and Zilinskas, Antanas},
  journal={Towards global optimization},
  volume={2},
  number={117-129},
  pages={2},
  year={1978}
}

@article{Jones1998:Efficient,
  title={Efficient global optimization of expensive black-box functions},
  author={Jones, Donald R and Schonlau, Matthias and Welch, William J},
  journal={Journal of Global optimization},
  volume={13},
  number={4},
  pages={455--492},
  year={1998},
  publisher={Springer}
}

@article{Kushner1964:New,
    author = {Kushner, H. J.},
    title = "{A New Method of Locating the Maximum Point of an Arbitrary Multipeak Curve in the Presence of Noise}",
    journal = {Journal of Basic Engineering},
    volume = {86},
    number = {1},
    pages = {97-106},
    year = {1964},
    month = {03},
    issn = {0021-9223},
    doi = {10.1115/1.3653121},
    url = {https://doi.org/10.1115/1.3653121},
    eprint = {https://asmedigitalcollection.asme.org/fluidsengineering/article-pdf/86/1/97/5763745/97\_1.pdf},
}

@inproceedings{Emmerich2011:Hypervolume,
  title={Hypervolume-based expected improvement: Monotonicity properties and exact computation},
  author={Emmerich, Michael TM and Deutz, Andr{\'e} H and Klinkenberg, Jan Willem},
  booktitle={2011 IEEE Congress of Evolutionary Computation (CEC)},
  pages={2147--2154},
  year={2011},
  organization={IEEE}
}

@article{Keane2006:Statistical,
  title={Statistical improvement criteria for use in multiobjective design optimization},
  author={Keane, Andy J},
  journal={AIAA journal},
  volume={44},
  number={4},
  pages={879--891},
  year={2006}
}

@article{Svenson2010:Multiobjective,
  title={Multiobjective optimization of expensive black-box functions via expected maximin improvement},
  author={Svenson, Joshua D and Santner, Thomas J},
  journal={The Ohio State University, Columbus, Ohio},
  volume={32},
  year={2010}
}

@article{Picheny2015:Multiobjective,
  title={Multiobjective optimization using Gaussian process emulators via stepwise uncertainty reduction},
  author={Picheny, Victor},
  journal={Statistics and Computing},
  volume={25},
  number={6},
  pages={1265--1280},
  year={2015},
  publisher={Springer}
}

@inproceedings{Wagner2010:Expected,
  title={On expected-improvement criteria for model-based multi-objective optimization},
  author={Wagner, Tobias and Emmerich, Michael and Deutz, Andr{\'e} and Ponweiser, Wolfgang},
  booktitle={International Conference on Parallel Problem Solving from Nature},
  pages={718--727},
  year={2010},
  organization={Springer}
}

@inproceedings{Ponweiser2008:Multiobjective,
  title={Multiobjective optimization on a limited budget of evaluations using model-assisted $\mathcal{S}$-metric selection},
  author={Ponweiser, Wolfgang and Wagner, Tobias and Biermann, Dirk and Vincze, Markus},
  booktitle={International Conference on Parallel Problem Solving from Nature},
  pages={784--794},
  year={2008},
  organization={Springer}
}

@article{Suzuki2019:Multi,
  title={Multi-objective Bayesian Optimization using Pareto-frontier Entropy},
  author={Suzuki, Shinya and Takeno, Shion and Tamura, Tomoyuki and Shitara, Kazuki and Karasuyama, Masayuki},
  journal={arXiv preprint arXiv:1906.00127},
  year={2019}
}

@inproceedings{Hernandez2016:Predictive,
  title={Predictive entropy search for multi-objective Bayesian Optimization},
  author={Hern{\'a}ndez-Lobato, Daniel and Hernandez-Lobato, Jose and Shah, Amar and Adams, Ryan},
  booktitle={International Conference on Machine Learning},
  pages={1492--1501},
  year={2016}
}

@inproceedings{Belakaria2019:Max,
  title={Max-value Entropy Search for Multi-Objective Bayesian Optimization},
  author={Belakaria, Syrine and Deshwal, Aryan and Doppa, Janardhan Rao},
  booktitle={Advances in Neural Information Processing Systems},
  pages={7823--7833},
  year={2019}
}

@inproceedings{Hernandez2014:Predictive,
author = {Henr\'{a}ndez-Lobato, Jos\'{e} Miguel and Hoffman, Matthew W. and Ghahramani, Zoubin},
title = {Predictive Entropy Search for Efficient Global Optimization of Black-Box Functions},
year = {2014},
publisher = {MIT Press},
address = {Cambridge, MA, USA},
booktitle = {Proceedings of the 27th International Conference on Neural Information Processing Systems - Volume 1},
pages = {918–926},
numpages = {9},
location = {Montreal, Canada},
series = {NIPS’14}
}

@inproceedings{Wang2017:Max,
author = {Wang, Zi and Jegelka, Stefanie},
title = {Max-Value Entropy Search for Efficient Bayesian Optimization},
year = {2017},
publisher = {JMLR.org},
booktitle = {Proceedings of the 34th International Conference on Machine Learning - Volume 70},
pages = {3627–3635},
numpages = {9},
location = {Sydney, NSW, Australia},
series = {ICML’17} 
}

@inproceedings{Paria2019:Flexible,
  title={A Flexible Framework for Multi-Objective Bayesian Optimization using Random Scalarizations},
  author={Biswajit Paria and Kirthevasan Kandasamy and Barnab{\'a}s P{\'o}czos},
  booktitle={UAI},
  year={2019}
}

@inproceedings{Belakaria2020:Multi,
  title={Multi-Fidelity Multi-Objective Bayesian Optimization: An Output Space Entropy Search Approach.},
  author={Belakaria, Syrine and Deshwal, Aryan and Doppa, Janardhan Rao},
  booktitle={AAAI},
  pages={10035--10043},
  year={2020}
}

@book{Jin2006:Multi,
  title={Multi-objective Machine Learning},
  author={Jin, Yaochu},
  volume={16},
  year={2006},
  publisher={Springer Science \& Business Media}
}

@inproceedings{Hutter2011:Sequential,
  title={Sequential model-based optimization for general algorithm configuration},
  author={Hutter, Frank and Hoos, Holger H and Leyton-Brown, Kevin},
  booktitle={International conference on learning and intelligent optimization},
  pages={507--523},
  year={2011},
  organization={Springer}
}

@incollection{Igel2005:Evolutionary,
  title={Evolutionary optimization of neural systems: The use of strategy adaptation},
  author={Igel, Christian and Wiegand, Stefan and Friedrichs, Frauke},
  booktitle={Trends and Applications in Constructive Approximation},
  pages={103--123},
  year={2005},
  publisher={Springer}
}

@inproceedings{Nolfi2002:Evolution,
    author = {Stefano Nolfi and Domenico Parisi},
    title = {Evolution of Artificial Neural Networks},
    booktitle = {In Handbook of brain theory and neural networks},
    year = {2002},
    pages = {418--421},
    publisher = {MIT Press}
}

@article{Yao1999:Evolving,
  title={Evolving artificial neural networks},
  author={Yao, Xin},
  journal={Proceedings of the IEEE},
  volume={87},
  number={9},
  pages={1423--1447},
  year={1999},
  publisher={IEEE}
}

@inproceedings{Zuluaga2013:Active,
  title={Active learning for multi-objective optimization},
  author={Zuluaga, Marcela and Sergent, Guillaume and Krause, Andreas and P{\"u}schel, Markus},
  booktitle={International Conference on Machine Learning},
  pages={462--470},
  year={2013}
}

@article{Zhang2009:Expensive,
  title={Expensive multiobjective optimization by MOEA/D with Gaussian process model},
  author={Zhang, Qingfu and Liu, Wudong and Tsang, Edward and Virginas, Botond},
  journal={IEEE Transactions on Evolutionary Computation},
  volume={14},
  number={3},
  pages={456--474},
  year={2009},
  publisher={IEEE}
}

@book{Nakayama2009:Sequential,
  title={Sequential approximate multiobjective optimization using computational intelligence},
  author={Nakayama, Hirotaka and Yun, Yeboon and Yoon, Min},
  year={2009},
  publisher={Springer Science \& Business Media}
}

@article{Golovin2020:Random,
  title={Random Hypervolume Scalarizations for Provable Multi-Objective Black Box Optimization},
  author={Golovin, Daniel and others},
  journal={arXiv preprint arXiv:2006.04655},
  year={2020}
}

@article{Belakaria2020:Information,
  title={Information-Theoretic Multi-Objective Bayesian Optimization with Continuous Approximations},
  author={Belakaria, Syrine and Deshwal, Aryan and Doppa, Janardhan Rao},
  journal={arXiv preprint arXiv:2009.05700},
  year={2020}
}

@string{ CVPR       = {Computer Vision and Pattern Recognition} }

@string{ ICLR       = {Int. Conf. Learning Representations} }

@article{Frazier:08,
  author      = {Frazier, P. and Powell, W. and Dayanik, S.},
  title       = {A Knowledge-Gradient Policy for Sequential Information Collection},
  journal     = {SIAM Journal on Control and Optimization},
  volume      = {47},
  number      = {5},
  pages       = {2410--2439},
  year        = {2008}
}

@InProceedings{Gardner14,
  Title                    = {{Bayesian} optimization with inequality constraints},
  Author                   = {Gardner, Jacob and Kusner, Matt and Xu, Zhixiang and Weinberger, Kilian and Cunningham, John},
  Booktitle                = {Proceedings of the International Conference on Machine Learning (ICML)},
  Year                     = {2014},
  Pages                    = {937--945}
}

@article{Ariafar2019,
  author  = {Setareh Ariafar and Jaume Coll-Font and Dana Brooks and Jennifer Dy},
  title   = {ADMMBO: Bayesian Optimization with Unknown Constraints using ADMM},
  journal = {Journal of Machine Learning Research},
  year    = {2019},
  volume  = {20},
  number  = {123},
  pages   = {1-26},
}

@article{letham2019,
author = "Letham, Benjamin and Karrer, Brian and Ottoni, Guilherme and Bakshy, Eytan",
journal = "Bayesian Analysis",
number = "2",
pages = "495--519",
publisher = "International Society for Bayesian Analysis",
title = "Constrained Bayesian Optimization with Noisy Experiments",
volume = "14",
year = "2019"
}

@article{Picheny2016,
  title                    = {{Bayesian optimization under mixed constraints with a slack-variable augmented Lagrangian}},
  author                   = {Picheny, V. and Gramacy, R. B. and Wild, S. and Le Digabel, S.},
  journal                  = {Advances in Neural Information Processing Systems 29},
  year                     = {2016},
  oages                    = {1435--1443}
}

@inproceedings{Lobato15a,
  author    = {Jos{\'{e}} Miguel Hern{\'{a}}ndez{-}Lobato and
               Michael A. Gelbart and
               Matthew W. Hoffman and
               Ryan P. Adams and
               Zoubin Ghahramani},
  title     = {Predictive Entropy Search for {Bayesian} Optimization with Unknown Constraints},
  booktitle = {Proceedings of the International Conference on Machine Learning (ICML)},
  pages     = {1699--1707},
  year      = {2015}
}

@inproceedings{Gelbart14,
  author    = {Michael A. Gelbart and
               Jasper Snoek and
               Ryan P. Adams},
  title     = {{Bayesian} Optimization with Unknown Constraints},
  booktitle = {Proceedings of the Thirtieth Conference on Uncertainty in Artificial
               Intelligence (UAI)},
  pages     = {250--259},
  year      = {2014}
}

@article{perrone_cmes,
Author = {Valerio Perrone and Iaroslav Shcherbatyi and Rodolphe Jenatton and Cedric Archambeau and Matthias Seeger},
Title = {Constrained Bayesian Optimization with Max-Value Entropy Search},
Year = {2019},
Eprint = {NeurIPS Workshop on Meta-Learning},
}

@inproceedings{obray:2021:evaluation_metrics_graph,
  title = {Evaluation Metrics for Graph Generative Models: Problems, Pitfalls, and Practical Solutions},
  booktitle = {International Conference on Learning Representations},
  author = {O'Bray, Leslie and Horn, Max and Rieck, Bastian and Borgwardt, Karsten},
  year = {2021},
  month = oct,
  url = {https://openreview.net/forum?id=tBtoZYKd9n}
}

@article{nguyen:2009:surrogate_loss_functions,
  title = {On Surrogate Loss Functions and F-Divergences},
  author = {Nguyen, XuanLong and Wainwright, Martin J. and Jordan, Michael I.},
  year = {2009},
  month = apr,
  journal = {The Annals of Statistics},
  volume = {37},
  number = {2},
  pages = {876--904},
  issn = {0090-5364, 2168-8966},
  doi = {10.1214/08-AOS595},
  url = {https://projecteuclid.org/journals/annals-of-statistics/volume-37/issue-2/On-surrogate-loss-functions-and-f-divergences/10.1214/08-AOS595.full}
}

@inproceedings{vezhnevets2017feudal,
  title={Feudal networks for hierarchical reinforcement learning},
  author={Vezhnevets, Alexander Sasha and Osindero, Simon and Schaul, Tom and Heess, Nicolas and Jaderberg, Max and Silver, David and Kavukcuoglu, Koray},
  booktitle={International conference on machine learning},
  pages={3540--3549},
  year={2017},
  organization={PMLR}
}

@article{bouneffouf2020hyper,
  title={Hyper-parameter tuning for the contextual bandit},
  author={Bouneffouf, Djallel and Claeys, Emmanuelle},
  journal={arXiv preprint arXiv:2005.02209},
  year={2020}
}

@book{Birattari2009TuningM,
  title={Tuning Metaheuristics - A Machine Learning Perspective},
  author={Birattari, Mauro},
  publisher={Springer-Verlag},
  series={Studies in Computational Intelligence},
  year={2009},
}

@article{schede:2022:survey_methods_automated,
  title = {A Survey of Methods for Automated Algorithm Configuration},
  author = {Schede, Elias and Brandt, Jasmin and Tornede, Alexander and Wever, Marcel and Bengs, Viktor and H{\"u}llermeier, Eyke and Tierney, Kevin},
  year = {2022},
  month = oct,
  journal = {Journal of Artificial Intelligence Research},
  volume = {75},
  pages = {425--487},
  issn = {1076-9757},
  doi = {10.1613/jair.1.13676},
  url = {https://jair.org/index.php/jair/article/view/13676},
  copyright = {Copyright (c) 2022 Journal of Artificial Intelligence Research}
}

@article{zhuge2024language,
  title={Language agents as optimizable graphs},
  author={Zhuge, Mingchen and Wang, Wenyi and Kirsch, Louis and Faccio, Francesco and Khizbullin, Dmitrii and Schmidhuber, J{\"u}rgen},
  journal={arXiv preprint arXiv:2402.16823},
  year={2024}
}

@article{niculae2023discrete,
  title={Discrete latent structure in neural networks},
  author={Niculae, Vlad and Corro, Caio F and Nangia, Nikita and Mihaylova, Tsvetomila and Martins, Andr{\'e} FT},
  journal={arXiv preprint arXiv:2301.07473},
  year={2023}
}

@article{zantedeschi2023dag,
  title={DAG Learning on the Permutahedron},
  author={Zantedeschi, Valentina and Franceschi, Luca and Kaddour, Jean and Kusner, Matt J and Niculae, Vlad},
  journal={arXiv preprint arXiv:2301.11898},
  year={2023}
}

@article{liu2018darts,
  title={Darts: Differentiable architecture search},
  author={Liu, Hanxiao and Simonyan, Karen and Yang, Yiming},
  journal={arXiv preprint arXiv:1806.09055},
  year={2018}
}

@article{gal2017concrete,
  title={Concrete dropout},
  author={Gal, Yarin and Hron, Jiri and Kendall, Alex},
  journal={Advances in neural information processing systems},
  volume={30},
  year={2017}
}

@article{hutter:2009:paramils_automatic_algorithm,
  title = {ParamILS: An Automatic Algorithm Configuration Framework},
  author = {Hutter, F. and Hoos, H. H. and {Leyton-Brown}, K. and Stuetzle, T.},
  year = {2009},
  month = oct,
  journal = {Journal of Artificial Intelligence Research},
  volume = {36},
  pages = {267--306},
  issn = {1076-9757},
  doi = {10.1613/jair.2861},
  url = {https://jair.org/index.php/jair/article/view/10628},
  copyright = {Copyright (c)}
}

@article{li:2023:survey_evolutionary_deep,
  title = {Survey on Evolutionary Deep Learning: Principles, Algorithms, Applications, and Open Issues},
  author = {Li, Nan and Ma, Lianbo and Yu, Guo and Xue, Bing and Zhang, Mengjie and Jin, Yaochu},
  year = {2023},
  month = sep,
  journal = {ACM Comput. Surv.},
  volume = {56},
  number = {2},
  pages = {41:1--41:34},
  issn = {0360-0300},
  doi = {10.1145/3603704},
  url = {https://doi.org/10.1145/3603704}
}

@article{arlot:2010:survey_crossvalidation_procedures,
  title = {A Survey of Cross-Validation Procedures for Model Selection},
  author = {Arlot, Sylvain and Celisse, Alain},
  year = {2010},
  month = jan,
  journal = {Statistics Surveys},
  volume = {4},
  number = {none},
  pages = {40--79},
  issn = {1935-7516},
  doi = {10.1214/09-SS054},
  url = {https://projecteuclid.org/journals/statistics-surveys/volume-4/issue-none/A-survey-of-cross-validation-procedures-for-model-selection/10.1214/09-SS054.full}
}

@article{heuillet:2024:efficient_automation_neural,
  title = {Efficient Automation of Neural Network Design: A Survey on Differentiable Neural Architecture Search},
  author = {Heuillet, Alexandre and Nasser, Ahmad and Arioui, Hichem and Tabia, Hedi},
  year = {2024},
  month = jun,
  journal = {ACM Comput. Surv.},
  volume = {56},
  number = {11},
  pages = {270:1--270:36},
  issn = {0360-0300},
  doi = {10.1145/3665138},
  url = {https://doi.org/10.1145/3665138}
}

@article{wang:2023:recent_advances_bayesian,
  title = {Recent Advances in Bayesian Optimization},
  author = {Wang, Xilu and Jin, Yaochu and Schmitt, Sebastian and Olhofer, Markus},
  year = {2023},
  month = jul,
  journal = {ACM Comput. Surv.},
  volume = {55},
  number = {13s},
  pages = {287:1--287:36},
  issn = {0360-0300},
  doi = {10.1145/3582078},
  url = {https://doi.org/10.1145/3582078}
}

@article{ren:2021:comprehensive_survey_neurala,
  title = {A Comprehensive Survey of Neural Architecture Search: Challenges and Solutions},
  author = {Ren, Pengzhen and Xiao, Yun and Chang, Xiaojun and Huang, Po-yao and Li, Zhihui and Chen, Xiaojiang and Wang, Xin},
  year = {2021},
  month = may,
  journal = {ACM Comput. Surv.},
  volume = {54},
  number = {4},
  pages = {76:1--76:34},
  issn = {0360-0300},
  doi = {10.1145/3447582},
  url = {https://doi.org/10.1145/3447582}
}

@article{bischl:2023:hyperparameter_optimization_foundations,
  title = {Hyperparameter Optimization: Foundations, Algorithms, Best Practices, and Open Challenges},
  author = {Bischl, Bernd and Binder, Martin and Lang, Michel and Pielok, Tobias and Richter, Jakob and Coors, Stefan and Thomas, Janek and Ullmann, Theresa and Becker, Marc and Boulesteix, Anne-Laure and Deng, Difan and Lindauer, Marius},
  year = {2023},
  journal = {WIREs Data Mining and Knowledge Discovery},
  volume = {13},
  number = {2},
  pages = {e1484},
  issn = {1942-4795},
  doi = {10.1002/widm.1484},
  url = {https://onlinelibrary.wiley.com/doi/abs/10.1002/widm.1484},
}

@article{yang:2020:hyperparameter_optimization_machine,
  title = {On Hyperparameter Optimization of Machine Learning Algorithms: Theory and Practice},
  author = {Yang, Li and Shami, Abdallah},
  year = {2020},
  month = nov,
  journal = {Neurocomputing},
  volume = {415},
  pages = {295--316},
  issn = {09252312},
  doi = {10.1016/j.neucom.2020.07.061},
  url = {https://linkinghub.elsevier.com/retrieve/pii/S0925231220311693},
}

@misc{sharifnassab:2024:metaoptimize_framework_optimizing,
  title = {MetaOptimize: A Framework for Optimizing Step Sizes and Other Meta-Parameters},
  author = {Sharifnassab, Arsalan and Salehkaleybar, Saber and Sutton, Richard},
  year = {2024},
  month = oct,
  number = {arXiv:2402.02342},
  eprint = {2402.02342},
  publisher = {arXiv},
  doi = {10.48550/arXiv.2402.02342},
  url = {http://arxiv.org/abs/2402.02342},
}

@inproceedings{lin:2019:online_hyperparameter_learning,
  title = {Online Hyper-Parameter Learning for Auto-Augmentation Strategy},
  booktitle = {Proceedings of the IEEE/CVF International Conference on Computer Vision},
  author = {Lin, Chen and Guo, Minghao and Li, Chuming and Yuan, Xin and Wu, Wei and Yan, Junjie and Lin, Dahua and Ouyang, Wanli},
  year = {2019},
  pages = {6579--6588},
  url = {https://openaccess.thecvf.com/content_ICCV_2019/html/Lin_Online_Hyper-Parameter_Learning_for_Auto-Augmentation_Strategy_ICCV_2019_paper.html},
}

@inproceedings{hataya:2020:faster_autoaugment_learning,
  title = {Faster AutoAugment: Learning Augmentation Strategies Using Backpropagation},
  booktitle = {Computer Vision -- ECCV 2020},
  author = {Hataya, Ryuichiro and Zdenek, Jan and Yoshizoe, Kazuki and Nakayama, Hideki},
  editor = {Vedaldi, Andrea and Bischof, Horst and Brox, Thomas and Frahm, Jan-Michael},
  year = {2020},
  pages = {1--16},
  publisher = {Springer International Publishing},
  address = {Cham},
  doi = {10.1007/978-3-030-58595-2_1},
  isbn = {978-3-030-58595-2},
}

@inproceedings{mei:2015:using_machine_teaching,
  title = {Using Machine Teaching to Identify Optimal Training-Set Attacks on Machine Learners},
  booktitle = {AAAI},
  author = {Mei, Shike and Zhu, Xiaojin},
  year = {2015},
  pages = {2871--2877},
}

@inproceedings{Zitzler1998:Multiobjective,
  title={Multiobjective optimization using evolutionary algorithms—a comparative case study},
  author={Zitzler, Eckart and Thiele, Lothar},
  booktitle={International conference on parallel problem solving from nature},
  pages={292--301},
  year={1998},
  organization={Springer}
}

@inproceedings{owen1995randomly,
  title={Randomly permuted (t, m, s)-nets and (t, s)-sequences},
  author={Owen, Art B},
  booktitle={Monte Carlo and Quasi-Monte Carlo Methods in Scientific Computing: Proceedings of a conference at the University of Nevada, Las Vegas, Nevada, USA, June 23--25, 1994},
  pages={299--317},
  year={1995},
  organization={Springer}
}

@inproceedings{vicol2021unbiased,
  title={Unbiased gradient estimation in unrolled computation graphs with persistent evolution strategies},
  author={Vicol, Paul and Metz, Luke and Sohl-Dickstein, Jascha},
  booktitle={International Conference on Machine Learning},
  pages={10553--10563},
  year={2021},
  organization={PMLR}
}

@article{micaelli2021gradient,
  title={Gradient-based hyperparameter optimization over long horizons},
  author={Micaelli, Paul and Storkey, Amos J},
  journal={Advances in Neural Information Processing Systems},
  volume={34},
  pages={10798--10809},
  year={2021}
}

@article{wei2021finetuned,
  title={Finetuned language models are zero-shot learners},
  author={Wei, Jason and Bosma, Maarten and Zhao, Vincent Y and Guu, Kelvin and Yu, Adams Wei and Lester, Brian and Du, Nan and Dai, Andrew M and Le, Quoc V},
  journal={arXiv preprint arXiv:2109.01652},
  year={2021}
}

@article{schick2024toolformer,
  title={Toolformer: Language models can teach themselves to use tools},
  author={Schick, Timo and Dwivedi-Yu, Jane and Dess\`i, Roberto and Raileanu, Roberta and Lomeli, Maria and Hambro, Eric and Zettlemoyer, Luke and Cancedda, Nicola and Scialom, Thomas},
  journal={Advances in Neural Information Processing Systems},
  volume={37},
  year={2023}
}

@article{yang2024swe,
  title={Swe-agent: Agent-computer interfaces enable automated software engineering},
  author={Yang, John and Jimenez, Carlos E and Wettig, Alexander and Lieret, Kilian and Yao, Shunyu and Narasimhan, Karthik and Press, Ofir},
  journal={arXiv preprint arXiv:2405.15793},
  year={2024}
}

@article{shin2020autoprompt,
  title={Autoprompt: Eliciting knowledge from language models with automatically generated prompts},
  author={Shin, Taylor and Razeghi, Yasaman and Logan IV, Robert L and Wallace, Eric and Singh, Sameer},
  journal={arXiv preprint arXiv:2010.15980},
  year={2020}
}

@article{diao2022black,
  title={Black-box prompt learning for pre-trained language models},
  author={Diao, Shizhe and Huang, Zhichao and Xu, Ruijia and Li, Xuechun and Lin, Yong and Zhou, Xiao and Zhang, Tong},
  journal={arXiv preprint arXiv:2201.08531},
  year={2022}
}

@article{shi2024efficient,
  title={Efficient prompt optimization through the lens of best arm identification},
  author={Shi, Chengshuai and Yang, Kun and Chen, Zihan and Li, Jundong and Yang, Jing and Shen, Cong},
  journal={arXiv preprint arXiv:2402.09723},
  year={2024}
}

@article{wu2022self,
  title={Self-adaptive in-context learning: An information compression perspective for in-context example selection and ordering},
  author={Wu, Zhiyong and Wang, Yaoxiang and Ye, Jiacheng and Kong, Lingpeng},
  journal={arXiv preprint arXiv:2212.10375},
  year={2022}
}

@book{hutter2019automated,
  title={Automated machine learning: methods, systems, challenges},
  author={Hutter, Frank and Kotthoff, Lars and Vanschoren, Joaquin},
  year={2019},
  publisher={Springer Nature}
}

@article{brown2020language,
  title={Language models are few-shot learners},
  author={Brown, Tom B},
  journal={arXiv preprint arXiv:2005.14165},
  year={2020}
}

@article{bommasani2021opportunities,
  title={On the opportunities and risks of foundation models},
  author={Bommasani, Rishi and Hudson, Drew A and Adeli, Ehsan and Altman, Russ and Arora, Simran and von Arx, Sydney and Bernstein, Michael S and Bohg, Jeannette and Bosselut, Antoine and Brunskill, Emma and others},
  journal={arXiv preprint arXiv:2108.07258},
  year={2021}
}

@article{chandra2022gradient,
  title={Gradient descent: The ultimate optimizer},
  author={Chandra, Kartik and Xie, Audrey and Ragan-Kelley, Jonathan and Meijer, Erik},
  journal={Advances in Neural Information Processing Systems},
  volume={35},
  pages={8214--8225},
  year={2022}
}

@article{shalev2012online,
  title={Online learning and online convex optimization},
  author={Shalev-Shwartz, Shai and others},
  journal={Foundations and Trends{\textregistered} in Machine Learning},
  volume={4},
  number={2},
  pages={107--194},
  year={2012},
  publisher={Now Publishers, Inc.}
}

@book{Lemieux:1338331,
      author        = "Lemieux, Christiane",
      title         = "{Monte Carlo and Quasi-Monte Carlo Sampling}",
      publisher     = "Springer",
      address       = "Dordrecht",
      series        = "Springer series in statistics",
      year          = "2009",
      url           = "https://cds.cern.ch/record/1338331",
      doi           = "10.1007/978-0-387-78165-5",
}

@book{dick2010digital,
  title={Digital nets and sequences: discrepancy theory and quasi--Monte Carlo integration},
  author={Dick, Josef and Pillichshammer, Friedrich},
  year={2010},
  publisher={Cambridge University Press}
}

@article{mckay1979comparison,
  title={A comparison of three methods for selecting values of input variables in the analysis of output from a computer code},
  author={McKay, Michael D and Beckman, Richard J and Conover, William J},
  journal={Technometrics},
  volume={42},
  number={1},
  pages={55--61},
  year={1979},
  publisher={Taylor \& Francis}
}

@article{devroye2006nonuniform,
  title={Nonuniform random variate generation},
  author={Devroye, Luc},
  journal={Handbooks in operations research and management science},
  volume={13},
  pages={83--121},
  year={2006},
  publisher={Elsevier}
}

@article{sobol1979systematic,
  title={On the systematic search in a hypercube},
  author={Sobol’, IM},
  journal={SIAM Journal on Numerical Analysis},
  volume={16},
  number={5},
  pages={790--793},
  year={1979},
  publisher={SIAM}
}

@article{niederreiter1988low,
  title={Low-discrepancy and low-dispersion sequences},
  author={Niederreiter, Harald},
  journal={Journal of number theory},
  volume={30},
  number={1},
  pages={51--70},
  year={1988},
  publisher={Elsevier}
}

@article{Knowles2006:Parego,
  title={ParEGO: a hybrid algorithm with on-line landscape approximation for expensive multiobjective optimization problems},
  author={Knowles, Joshua},
  journal={IEEE Transactions on Evolutionary Computation},
  volume={10},
  number={1},
  pages={50--66},
  year={2006},
  publisher={IEEE}
}

@article{Golovin2020,
  title={Random hypervolume scalarizations for provable multi-objective black box optimization},
  author={Golovin, Daniel and Zhang, Qiuyi},
  journal={arXiv preprint arXiv:2006.04655},
  year={2020}
}

@incollection{mitchell:1983:learning_experimentation_acquiringa,
  title = {Learning by {{Experimentation}}: {{Acquiring}} and {{Refining Problem-Solving Heuristics}}},
  shorttitle = {Learning by {{Experimentation}}},
  booktitle = {Machine {{Learning}}: {{An Artificial Intelligence Approach}}},
  author = {Mitchell, Tom M. and Utgoff, Paul E. and Banerji, Ranan},
  editor = {Michalski, Ryszard S. and Carbonell, Jaime G. and Mitchell, Tom M.},
  year = {1983},
  series = {Symbolic {{Computation}}},
  pages = {163--190},
  publisher = {{Springer}},
  address = {{Berlin, Heidelberg}},
  doi = {10.1007/978-3-662-12405-5_6},
  url = {https://doi.org/10.1007/978-3-662-12405-5_6},
  isbn = {978-3-662-12405-5},
  langid = {english}
}

@inproceedings{maron:1994:hoeffding_races_accelerating,
  title = {Hoeffding {{Races}}: {{Accelerating Model Selection Search}} for {{Classification}} and {{Function Approximation}}},
  shorttitle = {Hoeffding {{Races}}},
  booktitle = {Advances in {{Neural Information Processing Systems}}},
  author = {Maron, Oded and Moore, Andrew},
  year = {1994},
  volume = {6},
  publisher = {{Morgan-Kaufmann}},
  url = {https://proceedings.neurips.cc/paper/1993/hash/02a32ad2669e6fe298e607fe7cc0e1a0-Abstract.html},
}

@Book{holland75,
  author =    {Holland, John H.},
  title =        {Adaptation in Natural and Artificial Systems},
  publisher =    {The MIT Press},
  year =         1975}

@article{rechenberg65,
  title = {Cybernetic Solution Path of an Experimental Problem},
  author = {Rechenberg, I.},
  year = {1965},
  journal = { Royal Aircraft Establishment Library Translation},
  number = {1122}
}

@article{fogel:1965:intelligent_decisionmaking_simulation,
  title = {Intelligent Decision-Making through a Simulation of Evolution},
  author = {Fogel, L.J. and Owens, A.J. and Walsh, M.J.},
  year = {1965},
  month = oct,
  journal = {Simulation},
  volume = {5},
  number = {4},
  pages = {267--279},
  doi = {10.1177/003754976500500413}
}

@inproceedings{birattari:2002:racing_algorithm_configuring,
  title = {A Racing Algorithm for Configuring Metaheuristics},
  booktitle = {Proceedings of the 4th {{Annual Conference}} on {{Genetic}} and {{Evolutionary Computation}}},
  author = {Birattari, Mauro and St{\"u}tzle, Thomas and Paquete, Luis and Varrentrapp, Klaus},
  year = {2002},
  month = jul,
  series = {{{GECCO}}'02},
  pages = {11--18},
  publisher = {{Morgan Kaufmann Publishers Inc.}},
  address = {{San Francisco, CA, USA}},
  isbn = {978-1-55860-878-8},
}

@article{shahriari:2016:taking_human_out,
  title = {Taking the {{Human Out}} of the {{Loop}}: {{A Review}} of {{Bayesian Optimization}}},
  shorttitle = {Taking the {{Human Out}} of the {{Loop}}},
  author = {Shahriari, Bobak and Swersky, Kevin and Wang, Ziyu and Adams, Ryan P. and {de Freitas}, Nando},
  year = {2016},
  month = jan,
  journal = {Proceedings of the IEEE},
  volume = {104},
  number = {1},
  pages = {148--175},
  issn = {1558-2256},
  doi = {10.1109/JPROC.2015.2494218},
}

@article{hestenes1952methods,
  title={Methods of conjugate gradients for solving},
  author={Hestenes, Magnus R and Stiefel, Eduard},
  journal={Journal of research of the National Bureau of Standards},
  volume={49},
  number={6},
  pages={409},
  year={1952}
}

@article{bengio:2013:representation_learning_reviewa,
  title = {Representation Learning: A Review and New Perspectives},
  author = {Bengio, Yoshua and Courville, Aaron and Vincent, Pascal},
  year = {2013},
  journal = {IEEE Transactions on Pattern Analysis and Machine Intelligence},
  volume = {35},
  number = {8},
  pages = {1798--1828},
  url = {https://ieeexplore.ieee.org/document/6472238},
}

@inproceedings{duan:2019:unsupervised_model_selection,
  title = {Unsupervised Model Selection for Variational Disentangled Representation Learning},
  booktitle = {International Conference on Learning Representations},
  author = {Duan, Sunny and Matthey, Loic and Saraiva, Andre and Watters, Nick and Burgess, Chris and Lerchner, Alexander and Higgins, Irina},
  year = {2019},
  month = sep,
  url = {https://openreview.net/forum?id=SyxL2TNtvr}
}

@inproceedings{do:2019:theory_evaluation_metrics,
  title = {Theory and Evaluation Metrics for Learning Disentangled Representations},
  booktitle = {International Conference on Learning Representations},
  author = {Do, Kien and Tran, Truyen},
  year = {2019},
  month = sep,
  url = {https://openreview.net/forum?id=HJgK0h4Ywr},
}

@inproceedings{locatello:2019:challenging_common_assumptions,
  title = {Challenging Common Assumptions in the Unsupervised Learning of Disentangled Representations},
  booktitle = {Proceedings of the 36th International Conference on Machine Learning},
  author = {Locatello, Francesco and Bauer, Stefan and Lucic, Mario and Raetsch, Gunnar and Gelly, Sylvain and Sch{\"o}lkopf, Bernhard and Bachem, Olivier},
  year = {2019},
  month = may,
  pages = {4114--4124},
  publisher = {PMLR},
  url = {https://proceedings.mlr.press/v97/locatello19a.html},
}

@inproceedings{NIPS2016_8a3363ab,
 author = {Salimans, Tim and Goodfellow, Ian and Zaremba, Wojciech and Cheung, Vicki and Radford, Alec and Chen, Xi and Chen, Xi},
 booktitle = {Advances in Neural Information Processing Systems},
 editor = {D. Lee and M. Sugiyama and U. Luxburg and I. Guyon and R. Garnett},
 pages = {},
 publisher = {Curran Associates, Inc.},
 title = {Improved Techniques for Training GANs},
 url = {https://proceedings.neurips.cc/paper_files/paper/2016/file/8a3363abe792db2d8761d6403605aeb7-Paper.pdf},
 volume = {29},
 year = {2016}
}

@inproceedings{DBLP:conf/nips/HeuselRUNH17,
  author       = {Martin Heusel and
                  Hubert Ramsauer and
                  Thomas Unterthiner and
                  Bernhard Nessler and
                  Sepp Hochreiter},
  editor       = {Isabelle Guyon and
                  Ulrike von Luxburg and
                  Samy Bengio and
                  Hanna M. Wallach and
                  Rob Fergus and
                  S. V. N. Vishwanathan and
                  Roman Garnett},
  title        = {GANs Trained by a Two Time-Scale Update Rule Converge to a Local Nash
                  Equilibrium},
  booktitle    = {Advances in Neural Information Processing Systems 30: Annual Conference
                  on Neural Information Processing Systems 2017, December 4-9, 2017,
                  Long Beach, CA, {USA}},
  pages        = {6626--6637},
  year         = {2017},
  url          = {https://proceedings.neurips.cc/paper/2017/hash/8a1d694707eb0fefe65871369074926d-Abstract.html},
}

@article{pang:2021:deep_learning_anomaly,
  title = {Deep Learning for Anomaly Detection: A Review},
  author = {Pang, Guansong and Shen, Chunhua and Cao, Longbing and Hengel, Anton Van Den},
  year = {2021},
  month = mar,
  journal = {ACM Computing Surveys},
  volume = {54},
  number = {2},
  pages = {38:1--38:38},
  issn = {0360-0300},
  doi = {10.1145/3439950},
  url = {https://doi.org/10.1145/3439950},
}

@article{ma:2023:need_unsupervised_outlier,
  title = {The Need for Unsupervised Outlier Model Selection: A Review and Evaluation of Internal Evaluation Strategies},
  author = {Ma, Martin Q. and Zhao, Yue and Zhang, Xiaorong and Akoglu, Leman},
  year = {2023},
  month = jun,
  journal = {ACM SIGKDD Explorations Newsletter},
  volume = {25},
  number = {1},
  pages = {19--35},
  issn = {1931-0145, 1931-0153},
  doi = {10.1145/3606274.3606277},
  url = {https://dl.acm.org/doi/10.1145/3606274.3606277},
}

@inproceedings{kynkaanniemi:2019:improved_precision_recall,
  title = {Improved Precision and Recall Metric for Assessing Generative Models},
  booktitle = {Advances in Neural Information Processing Systems},
  author = {Kynk{\"a}{\"a}nniemi, Tuomas and Karras, Tero and Laine, Samuli and
            Lehtinen, Jaakko and Aila, Timo},
  year = {2019},
  volume = {32},
  url = {
         https://proceedings.neurips.cc/paper/2019/hash/0234c510bc6d908b28c70ff313743079-Abstract.html
         },
}

@inproceedings{naeem:2020:reliable_fidelity_diversitya,
  title = {Reliable Fidelity and Diversity Metrics for Generative Models},
  booktitle = {Proceedings of the 37th International Conference on Machine Learning},
  author = {Naeem, Muhammad Ferjad and Oh, Seong Joon and Uh, Youngjung and Choi, Yunjey and Yoo, Jaejun},
  year = {2020},
  month = nov,
  pages = {7176--7185},
  publisher = {PMLR},
  issn = {2640-3498},
  url = {https://proceedings.mlr.press/v119/naeem20a.html},
}

@inproceedings{park:2023:probabilistic_precision_recall,
  title = {Probabilistic Precision and Recall Towards Reliable Evaluation of Generative Models},
  booktitle = {Proceedings of the IEEE/CVF International Conference on Computer Vision},
  author = {Park, Dogyun and Kim, Suhyun},
  year = {2023},
  pages = {20099--20109},
  url = {https://openaccess.thecvf.com/content/ICCV2023/html/Park_Probabilistic_Precision_and_Recall_Towards_Reliable_Evaluation_of_Generative_Models_ICCV_2023_paper.html},
}

@article{yang:2020:evaluation_generative_models,
  title = {On the Evaluation of Generative Models in Music},
  author = {Yang, Li-Chia and Lerch, Alexander},
  year = {2020},
  month = may,
  journal = {Neural Computing and Applications},
  volume = {32},
  number = {9},
  pages = {4773--4784},
  issn = {0941-0643, 1433-3058},
  doi = {10.1007/s00521-018-3849-7},
  url = {http://link.springer.com/10.1007/s00521-018-3849-7},
}

@inproceedings{sajjadi:2018:assessing_generative_models,
  title = {Assessing Generative Models via Precision and Recall},
  booktitle = {Advances in Neural Information Processing Systems},
  author = {Sajjadi, Mehdi S. M. and Bachem, Olivier and Lucic, Mario and
            Bousquet, Olivier and Gelly, Sylvain},
  year = {2018},
  volume = {31},
  url = {
         https://proceedings.neurips.cc/paper/2018/hash/f7696a9b362ac5a51c3dc8f098b73923-Abstract.html
         },
}

@inproceedings{simon:2019:revisiting_precision_recall,
  title = {Revisiting Precision Recall Definition for Generative Modeling},
  booktitle = {Proceedings of the 36th International Conference on Machine Learning},
  author = {Simon, Loic and Webster, Ryan and Rabin, Julien},
  year = {2019},
  month = may,
  pages = {5799--5808},
  publisher = {PMLR},
  issn = {2640-3498},
  url = {https://proceedings.mlr.press/v97/simon19a.html},
}

@article{DBLP:journals/corr/GoodfellowPMXWOCB14,
  author       = {Ian J. Goodfellow and
                  Jean Pouget{-}Abadie and
                  Mehdi Mirza and
                  Bing Xu and
                  David Warde{-}Farley and
                  Sherjil Ozair and
                  Aaron C. Courville and
                  Yoshua Bengio},
  title        = {Generative Adversarial Networks},
  journal      = {CoRR},
  volume       = {abs/1406.2661},
  year         = {2014},
  url          = {http://arxiv.org/abs/1406.2661},
  eprinttype    = {arXiv},
  eprint       = {1406.2661},
  bibsource    = {dblp computer science bibliography, https://dblp.org}
}

@InProceedings{pmlr-v37-rezende15,
  title = 	 {Variational Inference with Normalizing Flows},
  author = 	 {Rezende, Danilo and Mohamed, Shakir},
  booktitle = 	 {Proceedings of the 32nd International Conference on Machine Learning},
  pages = 	 {1530--1538},
  year = 	 {2015},
  editor = 	 {Bach, Francis and Blei, David},
  volume = 	 {37},
  series = 	 {Proceedings of Machine Learning Research},
  address = 	 {Lille, France},
  month = 	 {Jul},
  publisher =    {PMLR},
  url = 	 {https://proceedings.mlr.press/v37/rezende15.html},
}

@inproceedings{ho:2020:denoising_diffusion_probabilistic,
  author = {Jonathan Ho and Ajay Jain and Pieter Abbeel},
  editor = {Hugo Larochelle and Marc'Aurelio Ranzato and Raia Hadsell and Maria{
            -}Florina Balcan and Hsuan{-}Tien Lin},
  title = {Denoising Diffusion Probabilistic Models},
  booktitle = {Advances in Neural Information Processing Systems 33},
  year = {2020},
}

@inproceedings{mockus:1975:bayesian_methods_seeking,
  title = {On Bayesian Methods for Seeking the Extremum},
  booktitle = {Optimization {{Techniques IFIP Technical Conference Novosibirsk}}, {{July}} 1\textendash 7, 1974},
  author = {Mo{\v c}kus, J.},
  editor = {Marchuk, G. I.},
  year = {1975},
  series = {Lecture {{Notes}} in {{Computer Science}}},
  pages = {400--404},
  publisher = {{Springer}},
  address = {{Berlin, Heidelberg}},
  doi = {10.1007/3-540-07165-2_55},
  isbn = {978-3-540-37497-8},
  langid = {english},
}

@inproceedings{hutter:2009:experimental_investigation_modelbased,
  title = {An {{Experimental Investigation}} of {{Model-based Parameter Optimisation}}: {{SPO}} and {{Beyond}}},
  shorttitle = {An {{Experimental Investigation}} of {{Model-based Parameter Optimisation}}},
  booktitle = {Proceedings of the 11th {{Annual Conference}} on {{Genetic}} and {{Evolutionary Computation}}},
  author = {Hutter, Frank and Hoos, Holger H. and {Leyton-Brown}, Kevin and Murphy, Kevin P.},
  year = {2009},
  series = {{{GECCO}} '09},
  pages = {271--278},
  publisher = {{ACM}},
  address = {{New York, NY, USA}},
  doi = {10.1145/1569901.1569940},
  url = {http://doi.acm.org/10.1145/1569901.1569940},
  urldate = {2018-05-09},
  isbn = {978-1-60558-325-9},
}

@article{box:1951:experimental_attainment_optimum,
  title = {On the {{Experimental Attainment}} of {{Optimum Conditions}}},
  author = {Box, G. E. P. and Wilson, K. B.},
  year = {1951},
  journal = {Journal of the Royal Statistical Society: Series B (Methodological)},
  volume = {13},
  number = {1},
  pages = {1--38},
  issn = {2517-6161},
  doi = {10.1111/j.2517-6161.1951.tb00067.x},
  url = {https://onlinelibrary.wiley.com/doi/abs/10.1111/j.2517-6161.1951.tb00067.x},
  urldate = {2021-12-20},
  langid = {english},
}

@Book{myers16:_respon,
  author =    {Myers, R.H. and Montgomery, D. C., and Anderson Cook, C.M},
  title =        {Response surface methodology: process and product optimization using design experiments},
  publisher =    {John wiley \& sons},
  year =         2016}

@article{sacks:1989:design_analysis_computer,
  title = {Design and {{Analysis}} of {{Computer Experiments}}},
  author = {Sacks, Jerome and Welch, William J. and Mitchell, Toby J. and Wynn, Henry P.},
  year = {1989},
  month = nov,
  journal = {Statistical Science},
  volume = {4},
  number = {4},
  pages = {409--423},
  publisher = {{Institute of Mathematical Statistics}},
  issn = {0883-4237, 2168-8745},
  doi = {10.1214/ss/1177012413},
  url = {https://projecteuclid.org/journals/statistical-science/volume-4/issue-4/Design-and-Analysis-of-Computer-Experiments/10.1214/ss/1177012413.full},
}

@inproceedings{yogotama2015-emnlp15,
  title={Bayesian Optimization of Text Representations},
  author={Yogatama, Dani and Kong, Lingpeng and Smith, Noah A.},
  booktitle={In Proceedings of the 2015 Conference on Empirical Methods in Natural Language Processing (EMNLP)},
  pages={2100--2105},
  year={2015},
  organization={Association for Computational Linguistics}
}

@inproceedings{bornschein-icml20,
   author    = {J. Bornschein and F. Visin and S. Osindero},
   title     = {Small Data, Big Decisions: Model Selection in the Small-Data Regime},
   booktitle = {In Proceedings of the 37th International Conference on Machine Learning (ICML'20)},
   year      = {2020},
}

@inproceedings{ariafar-aistats21,
   author    = {S. Ariafar and Z. Mariet and D. Brooks and J. Dy and J. Snoek},
   title     = {Faster and More Reliable Tuning of Neural Networks: Bayesian Optimization with Importance Sampling},
   booktitle = {International Conference on Artificial Intelligence and Statistics (AISTATS'21)},
   year      = {2021},
}

@inproceedings{karnin-icml13,
   author    = {Karnin, Z. and Koren, T. and Somekh, O.},
   title     = {Almost optimal exploration in multi-armed bandits},
   booktitle = {International Conference on Machine Learning 30},
   editor    = {Globerson, A. and Sha, F.},
   publisher = {Omni Press},
   year      = {2013},
}

@article{AMT2020,
Author = {Valerio Perrone and Huibin Shen and Aida Zolic and Iaroslav Shcherbatyi and Amr Ahmed and Tanya Bansal and Michele Donini and Fela Winkelmolen and Rodolphe Jenatton and Jean Baptiste Faddoul and Barbara Pogorzelska and Miroslav Miladinovic and Krishnaram Kenthapadi and Matthias Seeger and Cédric Archambeau},
Title = {Amazon SageMaker Automatic Model Tuning: Scalable Black-box Optimization},
journal= {arXiv:2012.08489},
Year = {2020}
}

@Article{Thompson1933,
  Title                    = {On the likelihood that one unknown probability exceeds another in view of the evidence of two samples},
  Author                   = {Thompson, William R},
  Journal                  = {Biometrika},
  Year                     = {1933},
  Pages                    = {285--294},
  Publisher                = {JSTOR},
}

@InProceedings{Hoffman2014,
  Title                    = {On correlation and budget constraints in model-based bandit optimization with application to automatic machine learning},
  Author                   = {Hoffman, Matthew and Shahriari, Bobak and de Freitas, Nando},
  Booktitle                = {Proceedings of the International Conference on Artificial Intelligence and Statistics (AISTATS)},
  Year                     = {2014},
  Pages                    = {365--374},
}

@Article{Srinivas2009,
author = {Srinivas, Niranjan and Krause, Andreas and Kakade, Sham and Seeger, Matthias},
title = {Gaussian Process Optimization in the Bandit Setting: No Regret and Experimental Design},
journal = {ICML},
Year = {2010},
pages = {1015–1022},
}

@Book{Griewank08:_evaluat_deriv,
  author =    {Griewank, Andreas and Walther, Andrea},
  title =        {Evaluating Derivatives: Principles and Techniques of Algorithmic Differentiation},
  publisher =    {SIAM},
  year =         2008,
  edition =   {Second Edition}}

@article{DBLP:journals/neco/WilliamsP90,
  author    = {Ronald J. Williams and
               Jing Peng},
  title     = {An Efficient Gradient-Based Algorithm for On-Line Training of Recurrent
               Network Trajectories},
  journal   = {Neural Comput.},
  volume    = {2},
  number    = {4},
  pages     = {490--501},
  year      = {1990},
  url       = {https://doi.org/10.1162/neco.1990.2.4.490},
  doi       = {10.1162/neco.1990.2.4.490}
}

@Article{Villemonteix2009,
  Title                    = {An informational approach to the global optimization of expensive-to-evaluate functions},
  Author                   = {Villemonteix, Julien and Vazquez, Emmanuel and Walter, Eric},
  Journal                  = {Journal of Global Optimization},
  Year                     = {2009},
  Number                   = {4},
  Pages                    = {509--534},
  Volume                   = {44},
  Publisher                = {Springer},
}

@Article{Hennig2012,
  Title                    = {Entropy search for information-efficient global optimization},
  Author                   = {Hennig, Philipp and Schuler, Christian J},
  Journal                  = {Journal of Machine Learning Research},
  Year                     = {2012},
  Number                   = {1},
  Pages                    = {1809--1837},
  Volume                   = {98888},
}

@TechReport{Hernandez-Lobato2014,
  Title                    = {Predictive Entropy Search for Efficient Global Optimization of Black-box Functions},
  Author                   = {Hern{\'a}ndez-Lobato, Jos{\'e} Miguel and Hoffman, Matthew W and Ghahramani, Zoubin},
  Institution              = {preprint arXiv:1406.2541},
  Year                     = {2014},
}

@InProceedings{Wilson2018,
  author    = {Wilson, James and Hutter, Frank and Deisenroth, Marc},
  title     = {Maximizing acquisition functions for Bayesian optimization},
  booktitle = {NeurIPS},
  year      = {2018},
  pages     = {9906--9917},
}

@article{hoffman2011portfolio,
  title={Portfolio Allocation for Bayesian Optimization.},
  author={Hoffman, Matthew D and Brochu, Eric and de Freitas, Nando},
  journal={UAI},
  year={2011},
}

@inproceedings{wu2018learning,
  title={Learning to teach with dynamic loss functions},
  author={Wu, Lijun and Tian, Fei and Xia, Yingce and Fan, Yang and Qin, Tao and Jian-Huang, Lai and Liu, Tie-Yan},
  booktitle={Advances in Neural Information Processing Systems},
  pages={6466--6477},
  year={2018}
}

@article{chen2022learning,
  title={Learning to optimize: A primer and a benchmark},
  author={Chen, Tianlong and Chen, Xiaohan and Chen, Wuyang and Heaton, Howard and Liu, Jialin and Wang, Zhangyang and Yin, Wotao},
  journal={Journal of Machine Learning Research},
  volume={23},
  number={189},
  pages={1--59},
  year={2022}
}

@inproceedings{thornton2013auto,
  title={Auto-WEKA: Combined selection and hyperparameter optimization of classification algorithms},
  author={Thornton, Chris and Hutter, Frank and Hoos, Holger H and Leyton-Brown, Kevin},
  booktitle={Proceedings of the 19th ACM SIGKDD international conference on Knowledge discovery and data mining},
  pages={847--855},
  year={2013}
}

@inproceedings{fusi2018probabilistic,
  title={Probabilistic matrix factorization for automated machine learning},
  author={Fusi, Nicolo and Sheth, Rishit and Elibol, Melih},
  booktitle={Advances in Neural Information Processing Systems},
  pages={3348--3357},
  year={2018}
}

@article{williams_learning_1989,
	title = {A learning algorithm for continually running fully recurrent neural networks},
	volume = {1},
	number = {2},
	journal = {Neural computation},
	author = {Williams, Ronald J. and Zipser, David},
	year = {1989},
	pages = {270--280},
}

@article{stone1974cross,
  title={Cross-validatory choice and assessment of statistical predictions},
  author={Stone, Mervyn},
  journal={Journal of the Royal Statistical Society: Series B (Methodological)},
  volume={36},
  number={2},
  pages={111--133},
  year={1974},
  publisher={Wiley Online Library}
}

@inproceedings{hutter_sequential_2011,
	title = {Sequential model-based optimization for general algorithm configuration},
	booktitle = {Int. {Conf.} on {Learning} and {Intelligent} {Optimization}},
	publisher = {Springer},
	author = {Hutter, Frank and Hoos, Holger H. and Leyton-Brown, Kevin},
	year = {2011},
	pages = {507--523},
}

@article{hutter_beyond_2015,
  title = {Beyond {Manual} {Tuning} of {Hyperparameters}},
  volume = {29},
  issn = {0933-1875, 1610-1987},
  url = {http://link.springer.com/10.1007/s13218-015-0381-0},
  doi = {10.1007/s13218-015-0381-0},
  language = {en},
  number = {4},
  journal = {KI - K{\"u}nstliche Intelligenz},
  author = {Hutter, Frank and Lücke, Jörg and Schmidt-Thieme, Lars},
  month = nov,
  year = {2015},
  pages = {329--337},
}

@inproceedings{swersky_multi-task_2013,
  title = {Multi-task {Bayesian} optimization},
  booktitle = {Advances in Neural Information Processing Systems 26},
  author = {Swersky, Kevin and Snoek, Jasper and Adams, Ryan P.},
  year = {2013},
  pages = {2004--2012},
}

@inproceedings{snoek_practical_2012,
  title = {Practical bayesian optimization of machine learning algorithms},
  booktitle = {Advances in neural information processing systems},
  author = {Snoek, Jasper and Larochelle, Hugo and Adams, Ryan P.},
  year = {2012},
  pages = {2951--2959},
}

@article{chen2022towards,
  title={Towards learning universal hyperparameter optimizers with transformers},
  author={Chen, Yutian and Song, Xingyou and Lee, Chansoo and Wang, Zi and Zhang, Richard and Dohan, David and Kawakami, Kazuya and Kochanski, Greg and Doucet, Arnaud and Ranzato, Marc'aurelio and others},
  journal={Advances in Neural Information Processing Systems},
  volume={35},
  pages={32053--32068},
  year={2022}
}

@inproceedings{eggensperger_towards_2013,
	title = {Towards an empirical foundation for assessing bayesian optimization of hyperparameters},
	booktitle = {{NIPS} workshop on {Bayesian} {Optimization} in {Theory} and {Practice}},
	author = {Eggensperger, Katharina and Feurer, Matthias and Hutter, Frank and Bergstra, James and Snoek, Jasper and Hoos, Holger and Leyton-Brown, Kevin},
	year = {2013},
	pages = {1--5},
}

@inproceedings{snoek_scalable_2015,
	title = {Scalable {Bayesian} {Optimization} {Using} {Deep} {Neural} {Networks}.},
	booktitle = {{International Conference on Machine Learning}},
	author = {Snoek, Jasper and Rippel, Oren and Swersky, Kevin and Kiros, Ryan and Satish, Nadathur and Sundaram, Narayanan and Patwary, Md Mostofa Ali and Prabhat, Mr and Adams, Ryan P.},
	year = {2015},
	pages = {2171--2180},
}

@article{bergstra_making_2013,
	title = {Making a {Science} of {Model} {Search}: {Hyperparameter} {Optimization} in {Hundreds} of {Dimensions} for {Vision} {Architectures}.},
	volume = {28},
	shorttitle = {Making a {Science} of {Model} {Search}},
	journal = {International Conference on Machine Learning},
	author = {Bergstra, James and Yamins, Daniel and Cox, David D.},
	year = {2013},
	pages = {115--123},
}

@article{bergstra_random_2012,
  title = {Random search for hyper-parameter optimization},
  volume = {13},
  url = {http://www.jmlr.org/papers/v13/bergstra12a.html},
  number = {Feb},
  journal = {Journal of Machine Learning Research},
  author = {Bergstra, James and Bengio, Yoshua},
  year = {2012},
  pages = {281--305},
}

@article{bengio_gradient-based_2000,
	title = {Gradient-based optimization of hyperparameters},
	volume = {12},
    number = {8},
	journal = {Neural computation},
	author = {Bengio, Yoshua},
	year = {2000},
	pages = {1889--1900},
}

@article{baydin2017automatic,
  author    = {Atilim Gunes Baydin and
               Barak A. Pearlmutter and
               Alexey Andreyevich Radul and
               Jeffrey Mark Siskind},
  title     = {Automatic differentiation in machine learning: a survey},
  journal   = {Journal of Machine Learning Research},
  volume    = {18},
  pages     = {1--43},
  year      = {2017},
  url       = {http://jmlr.org/papers/v18/papers/v18/17-468.html}
}

@inproceedings{franceschi2017forward,
  title={Forward and reverse gradient-based hyperparameter optimization},
  author={Franceschi, Luca and Donini, Michele and Frasconi, Paolo and Pontil, Massimiliano},
  booktitle={Proceedings of the 34th International Conference on Machine Learning-Volume 70},
  pages={1165--1173},
  year={2017},
}

@inproceedings{donini2019opt,
  title={MARTHE: Scheduling the Learning Rate Via Online Hypergradients},
  author={Donini, Michele and Franceschi, Luca and Majumder, Orchid and Pontil, Massimiliano and Frasconi, Paolo},
  booktitle={IJCAI},
  year={2020}
}

@inproceedings{larsen_design_1996,
	title = {Design and regularization of neural networks: the optimal use of a validation set},
	shorttitle = {Design and regularization of neural networks},
	booktitle = {Neural {Networks} for {Signal} {Processing} [1996] {VI}. {Proceedings} of the 1996 {IEEE} {Signal} {Processing} {Society} {Workshop}},
	publisher = {IEEE},
	author = {Larsen, Jan and Hansen, Lars Kai and Svarer, Claus and Ohlsson, M.},
	year = {1996},
	pages = {62--71}
}

@inproceedings{franceschi2019learning,
  title={Learning Discrete Structures for Graph Neural Networks},
  author={Franceschi, Luca and Niepert, Mathias and Pontil, Massimiliano and He, Xiao},
  booktitle={International Conference on Machine Learning},
  pages={1972--1982},
  year={2019}
}

@incollection{bengio2012practical,
  title={Practical recommendations for gradient-based training of deep architectures},
  author={Bengio, Yoshua},
  booktitle={Neural networks: Tricks of the trade},
  pages={437--478},
  year={2012},
  publisher={Springer}
}

@article{zoph2016neural,
  title={Neural architecture search with reinforcement learning},
  author={Zoph, Barret and Le, Quoc V},
  journal={International Conference on Learning Representations},
  year={2017}
}

@inproceedings{golovin2017google,
  title={{Google Vizier}: A service for black-box optimization},
  author={Golovin, Daniel and Solnik, Benjamin and Moitra, Subhodeep and Kochanski, Greg and Karro, John and Sculley, D},
  booktitle={Proceedings of the 23rd ACM SIGKDD international conference on knowledge discovery and data mining},
  pages={1487--1495},
  year={2017}
}

@article{li2016hyperband,
  title={Hyperband: A novel bandit-based approach to hyperparameter optimization},
  author={Li, Lisha and Jamieson, Kevin and DeSalvo, Giulia and Rostamizadeh, Afshin and Talwalkar, Ameet},
  journal={arXiv preprint arXiv:1603.06560},
  year={2016}
}

@article{loshchilov2016cma,
  title={CMA-ES for hyperparameter optimization of deep neural networks},
  author={Loshchilov, Ilya and Hutter, Frank},
  journal={arXiv preprint arXiv:1604.07269},
  year={2016}
}

@article{ravi_optimization_2016,
	title = {Optimization as a model for few-shot learning},
	urldate = {2017-08-29},
	author = {Ravi, Sachin and Larochelle, Hugo},
    	journal = {International Conference on Learning Representations},
	year = {2017}
}

@inproceedings{chen_learning_2017,
	title = {Learning to {Learn} without {Gradient} {Descent} by {Gradient} {Descent}},
	url = {http://proceedings.mlr.press/v70/chen17e.html},
	urldate = {2017-08-29},
	booktitle = {International {Conference} on {Machine} {Learning}},
	author = {Chen, Yutian and Hoffman, Matthew W. and Colmenarejo, Sergio Gómez and Denil, Misha and Lillicrap, Timothy P. and Botvinick, Matt and Freitas, Nando},
	year = {2017},
	pages = {748--756}
}

@inproceedings{finn_model-agnostic_2017,
  author    = {Chelsea Finn and
               Pieter Abbeel and
               Sergey Levine},
  title     = {Model-Agnostic Meta-Learning for Fast Adaptation of Deep Networks},
  booktitle = {Proceedings of the 34th International Conference on Machine Learning},
  pages     = {1126--1135},
  year      = {2017},
}

@inproceedings{Caruana94,
  author       = {Rich Caruana},
  editor       = {Gerald Tesauro and
                  David S. Touretzky and
                  Todd K. Leen},
  title        = {Learning Many Related Tasks at the Same Time with Backpropagation},
  booktitle    = {Advances in Neural Information Processing Systems 7, {[NIPS} Conference,
                  Denver, Colorado, USA, 1994]},
  pages        = {657--664},
  publisher    = {{MIT} Press},
  year         = {1994},
  url          = {http://papers.nips.cc/paper/959-learning-many-related-tasks-at-the-same-time-with-backpropagation},
}

@article{kaplan-arxiv19,
  title = {Scaling Laws for Neural Language Models},
  author = {J. Kaplan and S. McCandlish and T. Henighan and T. B. Brown and B. Chess and R. Child and S. Gray and A. Radford and J. Wu and D. Amodei},
journal = {arXiv:2001.08361 [cs.LG]},
  year={2020}
}

@article{hoffmann-arxiv22,
  title = {Training Compute-Optimal Large Language Models},
  author = {J. Hoffmann and S. Borgeaud and A. Mensch and E. Buchatskaya and T. Cai and E. Rutherford and D. de Las Casas and L. A. Hendricks and J. Welbl and A. Clark and T. Hennigan and E. Noland and K. Millican and G. van den Driessche and B. Damoc and A. Guy and S. Osindero and K. Simonyan and E. Elsen and J. W. Rae and O. Vinyals and L. Sifre},
journal = {arXiv:2203.15556 [cs.CL]},
  year={2022}
}

@inproceedings{vaswani_attention_2017,
  author    = {Ashish Vaswani and
               Noam Shazeer and
               Niki Parmar and
               Jakob Uszkoreit and
               Llion Jones and
               Aidan N. Gomez and
               Lukasz Kaiser and
               Illia Polosukhin},
  title     = {Attention is All you Need},
  booktitle = {Advances in Neural Information Processing Systems},
  pages     = {6000--6010},
  year      = {2017}
}

@article{auer2002finite,
  title={Finite-time analysis of the multiarmed bandit problem},
  author={Auer, Peter and Cesa-Bianchi, Nicolo and Fischer, Paul},
  journal={Machine learning},
  volume={47},
  pages={235--256},
  year={2002},
  publisher={Springer}
}

@inproceedings{li2010contextual,
  title={A contextual-bandit approach to personalized news article recommendation},
  author={Li, Lihong and Chu, Wei and Langford, John and Schapire, Robert E},
  booktitle={Proceedings of the 19th international conference on World wide web},
  pages={661--670},
  year={2010}
}

@article{ding2022syndicated,
  title={Syndicated bandits: A framework for auto tuning hyper-parameters in contextual bandit algorithms},
  author={Ding, Qin and Kang, Yue and Liu, Yi-Wei and Lee, Thomas Chun Man and Hsieh, Cho-Jui and Sharpnack, James},
  journal={Advances in Neural Information Processing Systems},
  volume={35},
  pages={1170--1181},
  year={2022}
}

@article{kang2024online,
  title={Online continuous hyperparameter optimization for generalized linear contextual bandits},
  author={Kang, Yue and Hsieh, Cho-Jui and Lee, Thomas},
  journal={TMLR},
  year={2024}
}

@article{schneider2024hyperband,
  title={Hyperband-based Bayesian Optimization for Black-box Prompt Selection},
  author={Schneider, Lennart and Wistuba, Martin and Klein, Aaron and Golebiowski, Jacek and Zappella, Giovanni and Merra, Felice Antonio},
  journal={arXiv preprint arXiv:2412.07820},
  year={2024}
}

@inproceedings{agrawal2013thompson,
  title={Thompson sampling for contextual bandits with linear payoffs},
  author={Agrawal, Shipra and Goyal, Navin},
  booktitle={International conference on machine learning},
  pages={127--135},
  year={2013},
  organization={PMLR}
}

@article{colson2007overview,
  title={An overview of bilevel optimization},
  author={Colson, Beno{\^\i}t and Marcotte, Patrice and Savard, Gilles},
  journal={Annals of operations research},
  volume={153},
  number={1},
  pages={235--256},
  year={2007},
  publisher={Springer}
}

@article{benmeziane2021comprehensive,
  title={A Comprehensive Survey on Hardware-Aware Neural Architecture Search},
  author={Benmeziane, Hadjer and Maghraoui, Kaoutar El and Ouarnoughi, Hamza and Niar, Smail and Wistuba, Martin and Wang, Naigang},
  journal={arXiv preprint arXiv:2101.09336},
  year={2021}
}

@inproceedings{wu2017bayesian,
  title={Bayesian optimization with gradients},
  author={Wu, Jian and Poloczek, Matthias and Wilson, Andrew G and Frazier, Peter},
  booktitle={Advances in Neural Information Processing Systems},
  pages={5267--5278},
  year={2017}
}

@book{krantz2012implicit,
  title={The implicit function theorem: history, theory, and applications},
  author={Krantz, Steven G and Parks, Harold R},
  year={2012},
  publisher={Springer Science \& Business Media}
}

@Article{Baxter2000,
  Title                    = {A model of inductive bias learning},
  Author                   = {Baxter, Jonathan},
  Journal                  = {Journal of Artificial Intelligence Research},
  Year                     = {2000},
  Pages                    = {149--198},
  Volume                   = {12},
  }

@article{kouw2018introduction,
  title={An introduction to domain adaptation and transfer learning},
  author={Kouw, Wouter M and Loog, Marco},
  journal={arXiv preprint arXiv:1812.11806},
  year={2018}
}

@article{pan2009survey,
  title={A survey on transfer learning},
  author={Pan, Sinno Jialin and Yang, Qiang},
  journal={IEEE Transactions on knowledge and data engineering},
  volume={22},
  number={10},
  pages={1345--1359},
  year={2009},
  publisher={IEEE}
}

@inproceedings{liao2018reviving,
  title={Reviving and Improving Recurrent Back-Propagation},
  author={Liao, Renjie and Xiong, Yuwen and Fetaya, Ethan and Zhang, Lisa and Yoon, KiJung and Pitkow, Xaq and Urtasun, Raquel and Zemel, Richard},
  booktitle={International Conference on Machine Learning},
  pages={3088--3097},
  year={2018}
}

@Article{he:2021:automl,
  author       = {Xin He and Kaiyong Zhao and Xiaowen Chu},
  title        = {AutoML: {A} survey of the state-of-the-art},
  year         = 2021,
  volume       = 212,
  pages        = 106622,
  doi          = {10.1016/j.knosys.2020.106622},
  url          = {https://doi.org/10.1016/j.knosys.2020.106622},
  journal      = {Knowl. Based Syst.},
  bibsource    = {dblp computer science bibliography, https://dblp.org}
}

@inproceedings{franceschi2018bilevel,
  title={Bilevel Programming for Hyperparameter Optimization and Meta-Learning},
  author={Franceschi, Luca and Frasconi, Paolo and Salzo, Saverio and Grazzi, Riccardo and Pontil, Massimiliano},
  booktitle={International Conference on Machine Learning},
  pages={1563--1572},
  year={2018}
}

@article{brooks1958discussion,
  title={A discussion of random methods for seeking maxima},
  author={Brooks, Samuel H},
  journal={Operations research},
  volume={6},
  number={2},
  pages={244--251},
  year={1958},
  publisher={INFORMS}
}

@article{bousquet2017critical,
  title={Critical hyper-parameters: No random, no cry},
  author={Bousquet, Olivier and Gelly, Sylvain and Kurach, Karol and Teytaud, Olivier and Vincent, Damien},
  journal={arXiv preprint arXiv:1706.03200},
  year={2017}
}

@inproceedings{li2019random,
  title={Random search and reproducibility for neural architecture search},
  author={Li, Liam and Talwalkar, Ameet},
  booktitle={Uncertainty in Artificial Intelligence},
  pages={367--377},
  year={2020},
}

@inproceedings{grazzi2020iteration,
  title={On the iteration complexity of hypergradient computation},
  author={Grazzi, Riccardo and Franceschi, Luca and Pontil, Massimiliano and Salzo, Saverio},
  booktitle={International Conference on Machine Learning},
  pages={3748--3758},
  year={2020},
  organization={PMLR}
}

@misc{tuningplaybookgithub,
  author = {Varun Godbole and George E. Dahl and Justin Gilmer and Christopher J. Shallue and Zachary Nado},
  title = {Deep Learning Tuning Playbook},
  url = {http://github.com/google-research/tuning_playbook},
  year = {2023},
  note = {Version 1.0}
}

@inproceedings{maclaurin2015gradient,
  title={Gradient-based hyperparameter optimization through reversible learning},
  author={Maclaurin, Dougal and Duvenaud, David and Adams, Ryan},
  booktitle={International Conference on Machine Learning},
  pages={2113--2122},
  year={2015}
}

@inproceedings{
choe2023betty,
title={Betty: An Automatic Differentiation Library for Multilevel Optimization},
author={Sang Keun Choe and Willie Neiswanger and Pengtao Xie and Eric Xing},
booktitle={The Eleventh International Conference on Learning Representations },
year={2023},
url={https://openreview.net/forum?id=LV_MeMS38Q9}
}

@article{grefenstette2019generalized,
  title={Generalized Inner Loop Meta-Learning},
  author={Grefenstette, Edward and Amos, Brandon and Yarats, Denis and Htut, Phu Mon and Molchanov, Artem and Meier, Franziska and Kiela, Douwe and Cho, Kyunghyun and Chintala, Soumith},
  journal={arXiv preprint arXiv:1910.01727},
  year={2019}
}

@article{blondel2022efficient,
  title={Efficient and modular implicit differentiation},
  author={Blondel, Mathieu and Berthet, Quentin and Cuturi, Marco and Frostig, Roy and Hoyer, Stephan and Llinares-L{\'o}pez, Felipe and Pedregosa, Fabian and Vert, Jean-Philippe},
  journal={Advances in neural information processing systems},
  volume={35},
  pages={5230--5242},
  year={2022}
}

@article{back1997handbook,
  title={Handbook of evolutionary computation},
  author={B{\"a}ck, Thomas and Fogel, David B and Michalewicz, Zbigniew},
  journal={Release},
  volume={97},
  number={1},
  pages={B1},
  year={1997}
}

@book{simon2013evolutionary,
  title={Evolutionary optimization algorithms},
  author={Simon, Dan},
  year={2013},
  publisher={John Wiley \& Sons}
}

@article{hansen2016cma,
  title={The CMA evolution strategy: A tutorial},
  author={Hansen, Nikolaus},
  journal={arXiv preprint arXiv:1604.00772},
  year={2016}
}

@inproceedings{kunanusont2017ntuple,
  title={The N-Tuple Bandit Evolutionary Algorithm for Automatic Game Improvement},
  author={Kunanusont, Kamolwan and Gaina, Raluca D. and Liu, Jialin and Perez-Liebana, Diego and Lucas, Simon M.},
  booktitle={2017 IEEE Congress on Evolutionary Computation (CEC)},
  note={\url{https://arxiv.org/pdf/1705.01080.pdf}},
  year={2017}
}

@article{hansen2003reducing,
  title={Reducing the time complexity of the derandomized evolution strategy with covariance matrix adaptation (CMA-ES)},
  author={Hansen, Nikolaus and M{\"u}ller, Sibylle D and Koumoutsakos, Petros},
  journal={Evolutionary computation},
  volume={11},
  number={1},
  pages={1--18},
  year={2003},
  publisher={MIT Press}
}

@article{tao2020learning,
  title={Learning to mutate with hypergradient guided population},
  author={Tao, Zhiqiang and Li, Yaliang and Ding, Bolin and Zhang, Ce and Zhou, Jingren and Fu, Yun},
  journal={Advances in Neural Information Processing Systems},
  volume={33},
  pages={17641--17651},
  year={2020}
}

@misc{hansen2019pycma,
  author       = {Nikolaus Hansen and Youhei Akimoto and Petr Baudis},
  title        = {{CMA-ES/pycma} on {G}ithub},
  howpublished = {Zenodo, DOI:10.5281/zenodo.2559634},
  month        = feb,
  year         = 2019,
  doi          = {10.5281/zenodo.2559634},
  url          = {https://doi.org/10.5281/zenodo.2559634},
}

@incollection{miikkulainen2024evolving,
  title={Evolving deep neural networks},
  author={Miikkulainen, Risto and Liang, Jason and Meyerson, Elliot and Rawal, Aditya and Fink, Dan and Francon, Olivier and Raju, Bala and Shahrzad, Hormoz and Navruzyan, Arshak and Duffy, Nigel and others},
  booktitle={Artificial intelligence in the age of neural networks and brain computing},
  pages={269--287},
  year={2024},
  publisher={Elsevier}
}

@article{werbos1990backpropagation,
  title={Backpropagation Through Time: What it Does and How To Do It},
  author={Werbos, Paul J},
  journal={Proceedings of the IEEE},
  volume={78},
  number={10},
  pages={1550--1560},
  year={1990},
  publisher={IEEE}
}

@article{scarselli2009graph,
	title={The graph neural network model},
	author={Scarselli, Franco and Gori, Marco and Tsoi, Ah Chung and Hagenbuchner, Markus and Monfardini, Gabriele},
	journal={IEEE Transactions on Neural Networks},
	volume={20},
	number={1},
	pages={61--80},
	year={2009},
	publisher={IEEE}
}

@inproceedings{pedregosa_hyperparameter_2016,
  author    = {Fabian Pedregosa},
  title     = {Hyperparameter optimization with approximate gradient},
  booktitle = {International Conference on Machine Learning},
  pages     = {737--746},
  year      = {2016}
}

@article{loshchilov2017sgdr,
  title={Sgdr: Stochastic gradient descent with warm restarts},
  author={Loshchilov, Ilya and Hutter, Frank},
  journal={International Conference on Learning Representations},
  year={2017}
}

@article{lorraine2018stochastic,
  title={Stochastic hyperparameter optimization through hypernetworks},
  author={Lorraine, Jonathan and Duvenaud, David},
  journal={arXiv preprint arXiv:1802.09419},
  year={2018}
}

@article{mackay2019self,
  title={Self-Tuning Networks: Bilevel Optimization of Hyperparameters using Structured Best-Response Functions},
  author={MacKay, Matthew and Vicol, Paul and Lorraine, Jon and Duvenaud, David and Grosse, Roger},
  journal={International Conference on Learning Representations},
  year={2019}
}

@inproceedings{orabona2016coin,
  title={Coin betting and parameter-free online learning},
  author={Orabona, Francesco and P{\'a}l, D{\'a}vid},
  booktitle={Advances in Neural Information Processing Systems},
  pages={577--585},
  year={2016}
}

@article{hansen2016using,
  title={Using deep Q-learning to control optimization hyperparameters},
  author={Hansen, Samantha},
  journal={arXiv preprint arXiv:1602.04062},
  year={2016}
}

@article{rumelhart1986learning,
  title={Learning representations by back-propagating errors},
  author={Rumelhart, David E and Hinton, Geoffrey E and Williams, Ronald J},
  journal={nature},
  volume={323},
  number={6088},
  pages={533--536},
  year={1986},
  publisher={Nature Publishing Group}
}

@inproceedings{baydin_2017_online,
  author    = {Atilim Gunes Baydin and
               Robert Cornish and
               David Martinez-Rubio and
               Mark Schmidt and
               Frank Wood},
  title     = {Online Learning Rate Adaptation with Hypergradient Descent},
  booktitle = {6th International Conference on Learning Representations},
  year      = {2018},
  url       = {https://openreview.net/forum?id=BkrsAzWAb}
}

@inproceedings{bergstra2011algorithms,
  title={Algorithms for hyper-parameter optimization},
  author={Bergstra, James S and Bardenet, R{\'e}mi and Bengio, Yoshua and K{\'e}gl, Bal{\'a}zs},
  booktitle={Advances in Neural Information Processing Systems},
  pages={2546--2554},
  year={2011}
}

@inproceedings{seeger2007cross,
  title={Cross-validation optimization for large scale hierarchical classification kernel methods},
  author={Seeger, Matthias},
  booktitle={Advances in neural information processing systems},
  pages={1233--1240},
  year={2007}
}

@article{chapelle2002choosing,
  title={Choosing multiple parameters for support vector machines},
  author={Chapelle, Olivier and Vapnik, Vladimir and Bousquet, Olivier and Mukherjee, Sayan},
  journal={Machine learning},
  volume={46},
  number={1-3},
  pages={131--159},
  year={2002},
  publisher={Springer}
}

@inproceedings{krizhevsky-nips12,
   title     = {{ImageNet} Classification with Deep Convolutional Neural Networks},
   author    = {A. Krizhevsky and I. Sutskever and G. Hinton},
   booktitle = {Proceedings of the 25th International Conference on Advances in Neural Information Processing Systems (NIPS'12)},
   year      = {2012},
}

@inproceedings{ilievski2017efficient,
  title={Efficient hyperparameter optimization for deep learning algorithms using deterministic RBF surrogates},
  author={Ilievski, Ilija and Akhtar, Taimoor and Feng, Jiashi and Shoemaker, Christine Annette},
  booktitle={Thirty-First AAAI Conference on Artificial Intelligence},
  year={2017}
}

@inproceedings{springenberg2016bayesian,
  title={Bayesian optimization with robust Bayesian neural networks},
  author={Springenberg, Jost Tobias and Klein, Aaron and Falkner, Stefan and Hutter, Frank},
  booktitle={Advances in neural information processing systems},
  pages={4134--4142},
  year={2016}
}

@inproceedings{hansen1996adapting,
  title={Adapting arbitrary normal mutation distributions in evolution strategies: The covariance matrix adaptation},
  author={Hansen, Nikolaus and Ostermeier, Andreas},
  booktitle={Proceedings of IEEE international conference on evolutionary computation},
  pages={312--317},
  year={1996},
  organization={IEEE}
}

@article{hospedales2020meta,
  title={Meta-learning in neural networks: A survey},
  author={Hospedales, Timothy and Antoniou, Antreas and Micaelli, Paul and Storkey, Amos},
  journal={arXiv preprint arXiv:2004.05439},
  year={2020}
}

@inproceedings{perrone2018scalable,
  title={Scalable hyperparameter transfer learning},
  author={Perrone, Valerio and Jenatton, Rodolphe and Seeger, Matthias W and Archambeau, C{\'e}dric},
  booktitle={Advances in Neural Information Processing Systems},
  pages={6845--6855},
  year={2018}
}

@inproceedings{real2019regularized,
  title={Regularized evolution for image classifier architecture search},
  author={Real, Esteban and Aggarwal, Alok and Huang, Yanping and Le, Quoc V},
  booktitle={Proceedings of the aaai conference on artificial intelligence},
  volume={33},
  pages={4780--4789},
  year={2019}
}

@article{akaike:1974:new_look_statistical,
  title = {A New Look at the Statistical Model Identification},
  author = {Akaike, H.},
  year = {1974},
  month = dec,
  journal = {IEEE Transactions on Automatic Control},
  volume = {19},
  number = {6},
  pages = {716--723},
  doi = {10.1109/TAC.1974.1100705},
}

@article{schwarz1978estimating,
  title={Estimating the dimension of a model},
  author={Schwarz, Gideon and others},
  journal={The annals of statistics},
  volume={6},
  number={2},
  pages={461--464},
  year={1978},
  publisher={Institute of Mathematical Statistics}
}

@article{friedrichs2005evolutionary,
  title={Evolutionary tuning of multiple SVM parameters},
  author={Friedrichs, Frauke and Igel, Christian},
  journal={Neurocomputing},
  volume={64},
  pages={107--117},
  year={2005},
  publisher={Elsevier}
}

@inproceedings{jamieson2016non,
  title={Non-stochastic best arm identification and hyperparameter optimization},
  author={Jamieson, Kevin and Talwalkar, Ameet},
  booktitle={Artificial Intelligence and Statistics},
  pages={240--248},
  year={2016}
}

@article{jaderberg2017population,
  title={Population based training of neural networks},
  author={Jaderberg, Max and Dalibard, Valentin and Osindero, Simon and Czarnecki, Wojciech M and Donahue, Jeff and Razavi, Ali and Vinyals, Oriol and Green, Tim and Dunning, Iain and Simonyan, Karen and others},
  journal={arXiv preprint arXiv:1711.09846},
  year={2017}
}

@article{li2017hyperband,
  title={Hyperband: A novel bandit-based approach to hyperparameter optimization},
  author={Li, Lisha and Jamieson, Kevin and DeSalvo, Giulia and Rostamizadeh, Afshin and Talwalkar, Ameet},
  journal={The Journal of Machine Learning Research},
  volume={18},
  number={1},
  pages={6765--6816},
  year={2017},
  publisher={JMLR. org}
}

@article{real2020automl,
  title={Automl-zero: Evolving machine learning algorithms from scratch},
  author={Real, Esteban and Liang, Chen and So, David R and Le, Quoc V},
  journal={arXiv preprint arXiv:2003.03384},
  year={2020}
}

@inproceedings{lorenzo2017particle,
  title={Particle swarm optimization for hyper-parameter selection in deep neural networks},
  author={Lorenzo, Pablo Ribalta and Nalepa, Jakub and Kawulok, Michal and Ramos, Luciano Sanchez and Pastor, Jos{\'e} Ranilla},
  booktitle={Proceedings of the genetic and evolutionary computation conference},
  pages={481--488},
  year={2017}
}

@inproceedings{klein2017fast,
  title={Fast bayesian optimization of machine learning hyperparameters on large datasets},
  author={Klein, Aaron and Falkner, Stefan and Bartels, Simon and Hennig, Philipp and Hutter, Frank},
  booktitle={Artificial Intelligence and Statistics},
  pages={528--536},
  year={2017}
}

@article{wang2017max,
  title={Max-value entropy search for efficient Bayesian optimization},
  author={Wang, Zi and Jegelka, Stefanie},
  journal={International Conference on Machine Learning},
  year={2017}
}

@book{murphy2012machine,
  title={Machine learning: a probabilistic perspective},
  author={Murphy, Kevin P},
  year={2012},
  publisher={MIT press}
}

@book{williams2006gaussian,
  title={Gaussian processes for machine learning},
  author={Williams, Christopher KI and Rasmussen, Carl Edward},
  volume={2},
  year={2006},
  publisher={MIT press Cambridge, MA}
}

@inproceedings{luketina2016scalable,
  title={Scalable gradient-based tuning of continuous regularization hyperparameters},
  author={Luketina, Jelena and Berglund, Mathias and Greff, Klaus and Raiko, Tapani},
  booktitle={International Conference on Machine Learning},
  pages={2952--2960},
  year={2016}
}

@article{mockus1978application,
  title={The application of Bayesian methods for seeking the extremum},
  author={Mo{\v{c}}kus, Jonas and Tiesis, Vytautas and Zilinskas, Antanas},
  journal={Towards global optimization},
  volume={2},
  number={117-129},
  pages={2},
  year={1978}
}

@Misc{sigopt-web-page,
  author = {Scott Clark and Patrick Hayes},
  title =  {{SigOpt} {W}eb page},
  url =  {https://sigopt.com},
  howpublished = {\url{https://sigopt.com}},
  year = {2019}
}

@article{falkner2018bohb,
  title={BOHB: Robust and efficient hyperparameter optimization at scale},
  author={Falkner, Stefan and Klein, Aaron and Hutter, Frank},
  journal={International Conference on Machine Learning},
  year={2018}
}

@inproceedings{feurer2015initializing,
  title={Initializing bayesian hyperparameter optimization via meta-learning},
  author={Feurer, Matthias and Springenberg, Jost Tobias and Hutter, Frank},
  booktitle={Twenty-Ninth AAAI Conference on Artificial Intelligence},
  year={2015}
}

@article{hazan2017hyperparameter,
  title={Hyperparameter optimization: A spectral approach},
  author={Hazan, Elad and Klivans, Adam and Yuan, Yang},
  journal={International Conference on Learning Representations},
  year={2018}
}

@article{swersky2014freeze,
  title={Freeze-thaw Bayesian optimization},
  author={Swersky, Kevin and Snoek, Jasper and Adams, Ryan Prescott},
  journal={arXiv preprint arXiv:1406.3896},
  year={2014}
}

@article{jones1998efficient,
  title={Efficient global optimization of expensive black-box functions},
  author={Jones, Donald R and Schonlau, Matthias and Welch, William J},
  journal={Journal of Global optimization},
  volume={13},
  number={4},
  pages={455--492},
  year={1998},
  publisher={Springer}
}

@article{miller2019stable,
  title={Stable recurrent models},
  author={Miller, John and Hardt, Moritz},
  journal={International Conference on Learning Representations},
  year={2019}
}

@inproceedings{rajeswaran2019meta,
  title={Meta-learning with implicit gradients},
  author={Rajeswaran, Aravind and Finn, Chelsea and Kakade, Sham M and Levine, Sergey},
  booktitle={Advances in Neural Information Processing Systems},
  pages={113--124},
  year={2019}
}

@incollection{larsen1998adaptive,
  title={Adaptive Regularization in Neural Network Modeling},
  author={Larsen, Jan and Svarer, Claus and Andersen, Lars Nonboe and Hansen, Lars Kai},
  booktitle={Neural Networks: Tricks of the Trade},
  pages={113--132},
  year={1998},
  publisher={Springer}
}

@inproceedings{shaban2019truncated,
  title={Truncated Back-propagation for Bilevel Optimization},
  author={Shaban, Amirreza and Cheng, Ching-An and Hatch, Nathan and Boots, Byron},
  booktitle={The 22nd International Conference on Artificial Intelligence and Statistics},
  pages={1723--1732},
  year={2019}
}

@inproceedings{almeida1987learning,
  title={A learning rule for asynchronous perceptrons with feedback in a combinatorial environment.},
  author={Almeida, Luis B},
  booktitle={Proceedings, 1st First International Conference on Neural Networks},
  volume={2},
  pages={609--618},
  year={1987},
}

@inproceedings{bai2019deep,
  title={Deep equilibrium models},
  author={Bai, Shaojie and Kolter, J Zico and Koltun, Vladlen},
  booktitle={Advances in Neural Information Processing Systems},
  pages={688--699},
  year={2019}
}

@article{elsken2019neural,
  title={Neural Architecture Search: A Survey.},
  author={Elsken, Thomas and Metzen, Jan Hendrik and Hutter, Frank},
  journal={Journal of Machine Learning Research},
  volume={20},
  number={55},
  pages={1--21},
  year={2019}
}

@article{goyal2017accurate,
  title={Accurate, large minibatch sgd: Training imagenet in 1 hour},
  author={Goyal, Priya and Doll{\'a}r, Piotr and Girshick, Ross and Noordhuis, Pieter and Wesolowski, Lukasz and Kyrola, Aapo and Tulloch, Andrew and Jia, Yangqing and He, Kaiming},
  journal={arXiv preprint arXiv:1706.02677},
  year={2017}
}

@article{harrison1978hedonic,
  title={Hedonic housing prices and the demand for clean air},
  author={Harrison Jr, David and Rubinfeld, Daniel L},
  journal={Journal of environmental economics and management},
  volume={5},
  number={1},
  pages={81--102},
  year={1978},
  publisher={Elsevier}
}

@article{neumann1966theory,
  title={Theory of self-reproducing automata},
  author={Neumann, J von},
  journal={Edited by Arthur W. Burks},
  year={1966}
}

@inproceedings{kennedy1995particle,
  title={Particle swarm optimization},
  author={Kennedy, James and Eberhart, Russell},
  booktitle={Proceedings of ICNN'95-international conference on neural networks},
  volume={4},
  pages={1942--1948},
  year={1995},
  organization={ieee}
}

@InProceedings{denevi2019learning,
  title = 	 {Learning-to-learn stochastic gradient descent with biased regularization},
  author = 	 {Denevi, Giulia and Ciliberto, Carlo and Grazzi, Riccardo and Pontil, Massimiliano},
  booktitle = 	 {Proc. 36th International Conference on Machine Learning},
  pages = 	 {1566--1575},
  year = 	 {2019},
  volume = 	 {97},
  series = 	 {PMLR},
}

@Article{haibe-kains:2020:transparency,
  author       = {Haibe-Kains, Benjamin and Adam, George Alexandru and
                  Hosny, Ahmed and Khodakarami, Farnoosh and Waldron,
                  Levi and Wang, Bo and McIntosh, Chris and
                  Goldenberg, Anna and Kundaje, Anshul and et al.},
  title        = {Transparency and reproducibility in artificial
                  intelligence},
  year         = 2020,
  volume       = 586,
  number       = 7829,
  month        = {10},
  pages        = {E14-–E16},
  issn         = {1476-4687},
  doi          = {10.1038/s41586-020-2766-y},
  url          = {http://dx.doi.org/10.1038/s41586-020-2766-y},
  journal      = {Nature},
  publisher    = {Springer Science and Business Media LLC}
}

@InProceedings{kohavi:1995:study_cross,
  author       = {Ron Kohavi},
  title        = {A Study of Cross-Validation and Bootstrap for
                  Accuracy Estimation and Model Selection},
  year         = 1995,
  booktitle    = {Proceedings of the Fourteenth International Joint
                  Conference on Artificial Intelligence},
  pages        = {1137-1145},
  url          = {http://ijcai.org/Proceedings/95-2/Papers/016.pdf},
  bibsource    = {dblp computer science bibliography, https://dblp.org}
}

@techreport{Li:19,
  author      = {Li, L. and Jamieson, K. and Rostamizadeh, A. and Gonina, E. and Hardt, M. and Recht, B. and Talwalkar, A},
  title       = {Massively Parallel Hyperparameter Tuning},
  number      = {1810.05934v4 [cs.LG]},
  institution = {ArXiv},
  year        = {2019}
}

@InProceedings{zoph:2018:learning_transferable,
  author       = {Barret Zoph and Vijay Vasudevan and Jonathon Shlens and Quoc V. Le},
  title        = {Learning Transferable Architectures for Scalable Image Recognition},
  year         = 2018,
  booktitle    = {{IEEE} Conference on Computer Vision and Pattern Recognition},
  pages        = {8697--8710},
  doi          = {10.1109/CVPR.2018.00907},
  url          = {http://openaccess.thecvf.com/content\_cvpr\_2018/html/Zoph\_Learning\_Transferable\_Architectures\_CVPR\_2018\_paper.html},
}

@Article{silver:2017:mastering_go,
  author       = {Silver, David and Schrittwieser, Julian and
                  Simonyan, Karen and Antonoglou, Ioannis and Huang,
                  Aja and Guez, Arthur and Hubert, Thomas and Baker,
                  Lucas and Lai, Matthew and Bolton, Adrian and et
                  al.},
  title        = {Mastering the game of Go without human knowledge},
  year         = 2017,
  volume       = 550,
  number       = 7676,
  month        = 10,
  pages        = {354–359},
  issn         = {1476-4687},
  doi          = {10.1038/nature24270},
  url          = {http://dx.doi.org/10.1038/nature24270},
  journal      = {Nature},
  publisher    = {Springer Science and Business Media LLC}
}

@Article{chen:2018:bayesian_optimization_alphago,
  author       = {Yutian Chen and Aja Huang and Ziyu Wang and Ioannis
                  Antonoglou and Julian Schrittwieser and David Silver
                  and Nando de Freitas},
  title        = {Bayesian Optimization in AlphaGo},
  year         = 2018,
  volume       = {abs/1812.06855},
  eprint       = {1812.06855},
  url          = {http://arxiv.org/abs/1812.06855},
  journal      = {CoRR},
  archiveprefix= {arXiv},
  bibsource    = {dblp computer science bibliography,
                  https://dblp.org}
}

@InProceedings{perrone2021sagemaker,
  author       = {Perrone, V. and Shen, H. and Zolic, A. and Shcherbatyi, I. and Ahmed, A. and Bansal, T. and Donini, M. and Winkelmolen, F. and Jenatton, R. Faddoul, J. and Pogorzelska, B. and Miladinovic, M. and Kenthapadi, K. and Seeger, M. and Archambeau, C.},
  title        = {{Amazon SageMaker} Automatic Model Tuning: Scalable Gradient-Free Optimization},
  year         = 2021,
  booktitle    = {ACM SIGKDD Conference on Knowledge Discovery and Data Mining}
}

@inproceedings{Alvi:19,
  author      = {Alvi, A. and Ru, B and Calliess, J. and Roberts, J. and Osborne, M.},
  title       = {Asynchronous Batch {Bayesian} Optimisation with Improved Local Penalisation},
  booktitle   = {International Conference on Machine Learning 36},
  year        = {2019},
  editor      = {Chaudhuri, K. and Salakhutdinov, R.},
  publisher   = {Proceedings of Machine Learning Research},
  volume      = {97},
  pages       = {253--262}
}

@article{Desautels:14,
  author      = {Desautels, T. and Krause, A. and Burdick, J.},
  title       = {Parallelizing Exploration-Exploitation Tradeoffs in {Gaussian} Process Bandit Optimization},
  journal     = {Journal of Machine Learning Research},
  volume      = {15},
  number      = {1},
  pages       = {3873--3923},
  year        = {2014}
}

@inproceedings{Gonzales:16,
  author      = {Gonz\'{a}lez, J. and Dai, Z. and Hennig, P. and Lawrence, N.},
  title       = {Batch {Bayesian} Optimization via Local Penalization},
  booktitle   = {Workshop on Artificial Intelligence and Statistics 19},
  editor      = {Gretton, A. and Robert, C.},
  year        = {2016},
  pages       = {648--657}
}

@incollection{Ginsbourger:10,
  author      = {Ginsbourger, D. and Le Riche, R. and Carraro, L.},
  title       = {Kriging is Well-Suited to Parallelize Optimization},
  booktitle   = {Computational Intelligence in Expensive Optimization Problems},
  publisher   = {Springer},
  pages       = {131--162},
  year        = {2010}
}

@techreport{Klein:20,
  author      = {Klein, A. and Tiao, L. and Lienart, T. and Archambeau, C. and Seeger, M.},
  title       = {Model-based Asynchronous Hyperparameter and Neural Architecture Search},
  number      = {2003.10865 [cs.LG]},
  institution = {ArXiv},
  year        = {2020}
}

@book{neal1996bayesian,
  author      = {Neal, R.~M.},
  title       = {Bayesian Learning for Neural Networks},
  number      = {118},
  publisher   = {Springer},
  series      = {Lecture Notes in Statistics},
  year        = {1996}
}

@inproceedings{kandasamy-neurips18,
 title     = {Neural Architecture Search with {B}ayesian Optimisation and Optimal Transport},
 author    = {K. Kandasamy and W. Neiswanger and J. Schneider and B. Poczos and E. P. Xing},
 booktitle = {Proceedings of the 31th International Conference on Advances in Neural Information Processing Systems (NeurIPS'18)},
 year      = {2018},
}

@inproceedings{Kandasamy:18,
  author      = {Kandasamy, K. and Krishnamurthy, A. and Schneider, J. and Poczos, B.},
  title       = {Parallelised {Bayesian} Optimisation via {Thompson} Sampling},
  booktitle   = {21st International Conference on Artificial Intelligence and Statistics},
   year       = {2018},
  pages       = {133-142}
}

@inproceedings{Tiao:21,
  author      = {Louis C Tiao and Aaron Klein and Matthias W Seeger and Edwin V. Bonilla and Cedric Archambeau and Fabio Ramos},
  title       = {{BORE}: {Bayesian} Optimization by Density-Ratio Estimation},
  booktitle   = {Proceedings of the 38th International Conference on Machine Learning},
   year       = {2021},
  pages       = {10289-10300}
}

@inproceedings{klein-bayesopt17,
 author    = {A. Klein and S. Falkner and N. Mansur and F. Hutter},
 title     = {RoBO: A Flexible and Robust Bayesian Optimization Framework in Python},
 booktitle = {NeurIPS Workshop on {B}ayesian Optimization (BayesOpt'17)},
 year      = {2017},
}

@incollection{torrey2010transfer,
  title={Transfer learning},
  author={Torrey, Lisa and Shavlik, Jude},
  booktitle={Handbook of research on machine learning applications and trends: algorithms, methods, and techniques},
  pages={242--264},
  year={2010},
  publisher={IGI global}
}

@misc{ruder:2017:overview_multitask_learning,
  title = {An Overview of Multi-Task Learning in Deep Neural Networks},
  author = {Ruder, Sebastian},
  year = {2017},
  month = jun,
  number = {arXiv:1706.05098},
  eprint = {1706.05098},
  publisher = {arXiv},
  doi = {10.48550/arXiv.1706.05098},
}

@article{hospedales:2021:metalearning_neural_networks,
  title = {Meta-Learning in Neural Networks: A Survey},
  author = {Hospedales, Timothy M and Antoniou, Antreas and Micaelli, Paul and Storkey, Amos J.},
  year = {2021},
  journal = {IEEE Transactions on Pattern Analysis and Machine Intelligence},
  pages = {1--1},
  issn = {1939-3539},
  doi = {10.1109/TPAMI.2021.3079209},
}

@article{lake:2019:omniglot_challenge_3year,
  title = {The Omniglot Challenge: A 3-Year Progress Report},
  author = {Lake, Brenden M and Salakhutdinov, Ruslan and Tenenbaum, Joshua B},
  year = {2019},
  month = oct,
  journal = {Current Opinion in Behavioral Sciences},
  series = {Artificial Intelligence},
  volume = {29},
  pages = {97--104},
  issn = {2352-1546},
  doi = {10.1016/j.cobeha.2019.04.007},
  url = {https://www.sciencedirect.com/science/article/pii/S2352154619300051},
}

@inproceedings{eggensperger-aaai15,
  author    =  {K. Eggensperger and F. Hutter and H.H. Hoos and K. Leyton-Brown},
  title     = {Efficient Benchmarking of Hyperparameter Optimizers via Surrogates},
  booktitle = {Proceedings of the 29th National Conference on Artificial Intelligence (AAAI'15)},
  year      = {2015},
}

@inproceedings{klein-nips19,
 title     = {Meta-Surrogate Benchmarking for Hyperparameter Optimization},
 author    = {A. Klein and Z. Dai and F. Hutter and N. Lawrence and J. Gonzalez},
 booktitle = {Proceedings of the 32th International Conference on Advances in Neural Information Processing Systems (NeurIPS'19)},
 year      = {2019},
}

@InProceedings{salinas-automl22,
  title     = {Syne Tune: A Library for Large Scale Hyperparameter Tuning and Reproducible Research},
  author    = {D. Salinas and M. Seeger and A. Klein and V. Perrone and M. Wistuba and C. Archambeau},
  booktitle = {First Conference on Automated Machine Learning (Main Track)},
  year      = {2022},
}

@article{klein-arxiv19a,
 author  = {A. Klein and F. Hutter},
 year    = {2019},
 title   = {Tabular Benchmarks for Joint Architecture and Hyperparameter Optimization},
 journal = {arXiv:1905.04970 [cs.LG]}
}

@inproceedings{eggensperger-neurips21,
title={{HPOB}ench: A Collection of Reproducible Multi-Fidelity Benchmark Problems for {HPO}},
author={K. Eggensperger and P. M{\"u}ller and N. Mallik and M. Feurer and R. Sass and A. Klein and N. Awad and M. Lindauer and F. Hutter},
booktitle={Thirty-fifth Conference on Neural Information Processing Systems Datasets and Benchmarks Track (Round 2)},
year={2021},
}

@article{smac3,
  author  = {Marius Lindauer and Katharina Eggensperger and Matthias Feurer and André Biedenkapp and Difan Deng and Carolin Benjamins and Tim Ruhkopf and René Sass and Frank Hutter},
  title   = {SMAC3: A Versatile Bayesian Optimization Package for Hyperparameter Optimization},
  journal = {Journal of Machine Learning Research},
  year    = {2022},
  volume  = {23},
  number  = {54},
  pages   = {1--9},
  url     = {http://jmlr.org/papers/v23/21-0888.html}
}

@inproceedings{kandasamy-jmlr19,
 author    = {K. Kandasamy and K. R. Vysyaraju and W. Neiswanger and B. Paria and C. R. Collins and J. Schneider and B. Poczos and E. P. Xing},
 title     = {Tuning Hyperparameters without Grad Students: Scalable and Robust Bayesian Optimisation with Dragonfly},
 booktitle = {Journal of Machine Learning Research 2020, Special Issue on Bayesian Optimization},
 year      = {2019}
}

@article{dong-iclr20,
 title     = {NAS-Bench-201: Extending the Scope of Reproducible Neural Architecture Search},
 author    = {X. Dong and Y. Yang},
 booktitle = {International Conference on Learning Representations (ICLR'20)},
 year      = {2020}
}

@inproceedings{oss_vizier,
  author    = {Xingyou Song and
               Sagi Perel and
               Chansoo Lee and
               Greg Kochanski and
               Daniel Golovin},
  title     = {Open Source Vizier: Distributed Infrastructure and API for Reliable and Flexible Black-box Optimization},
  booktitle = {Automated Machine Learning Conference, Systems Track (AutoML-Conf Systems)},
  year      = {2022},
}

@inproceedings{akiba-sigkdd19,
    title={Optuna: A Next-generation Hyperparameter Optimization Framework},
    author={T. Akiba and S. Sano and T. Yanase and T. Ohta and M. Koyama},
    booktitle={Proceedings of the 25th {ACM} {SIGKDD} International Conference on Knowledge Discovery and Data Mining},
    year={2019}
}

@article{liaw2018tune,
    title={Tune: A Research Platform for Distributed Model Selection and Training},
    author={Liaw, Richard and Liang, Eric and Nishihara, Robert
            and Moritz, Philipp and Gonzalez, Joseph E and Stoica, Ion},
    journal={arXiv preprint arXiv:1807.05118},
    year={2018}
}

@inproceedings{pfisterer-automl22,
  title     = {YAHPO Gym - An Efficient Multi-Objective Multi-Fidelity Benchmark for Hyperparameter Optimization. In International Conference on Automated Machine Learning},
  author    = {
F. Pfisterer and L. Schneider and J. Moosbauer and M. Binder and B. Bischl},
  booktitle = {First Conference on Automated Machine Learning (Main Track)},
  year      = {2022},
}

@article{zimmer-ieee21,
  author = {L. Zimmer and M. Lindauer and F. Hutter},
  title = {Auto-PyTorch Tabular: Multi-Fidelity MetaLearning for Efficient and Robust AutoDL},
  journal = {IEEE Transactions on Pattern Analysis and Machine Intelligence},
  year = {2021},
}

@inproceedings{ying-icml19,
  author    = {C. Ying and A. Klein and E. Real and E. Christiansen and K. Murphy and F. Hutter},
  title     = {{NAS-Bench-101}: Towards Reproducible Neural Architecture Search},
  booktitle = {Proceedings of the 36th International Conference on Machine Learning (ICML'19)},
  year      = {2019},
}

@article{vanschoren-sigkdd13a,
 author  = {J. Vanschoren and J. van Rijn and B. Bischl and L. Torgo},
 title   = {{OpenML}: Networked Science in Machine Learning},
 journal = {SIGKDD Explorations},
 year    = {2014}
}

@inproceedings{pineda-neurips21,
 author = {S. Pineda Arango and H. Jomaa and M. Wistuba and J. Grabocka},
 booktitle = {Proceedings of the Neural Information Processing Systems Track on Datasets and Benchmarks},
 title = {HPO-B: A Large-Scale Reproducible Benchmark for Black-Box HPO based on OpenML },
 year = {2021}
}

@inproceedings{google_vizier,
  author    = {Daniel Golovin and
               Benjamin Solnik and
               Subhodeep Moitra and
               Greg Kochanski and
               John Karro and
               D. Sculley},
  title     = {Google Vizier: {A} Service for Black-Box Optimization},
  booktitle = {Proceedings of the 23rd {ACM} {SIGKDD} International Conference on
               Knowledge Discovery and Data Mining, Halifax, NS, Canada, August 13
               - 17, 2017},
  pages     = {1487--1495},
  publisher = {{ACM}},
  year      = {2017},
  url       = {https://doi.org/10.1145/3097983.3098043},
  doi       = {10.1145/3097983.3098043},
}

@InProceedings{tan-cvpr19,
author = {M. Tan and B. Chen and R. Pang and V. Vasudevan and M. Sandler and A. Howard and Q. V. Le},
title = {MnasNet: Platform-Aware Neural Architecture Search for Mobile},
booktitle = {Proceedings of the IEEE/CVF Conference on Computer Vision and Pattern Recognition (CVPR'19)},
year = {2019}
}

@article{white-arxiv23,
  author = {C. White and M. Safari and R. Sukthanker and B. Ru and T. Elsken and A. Zela and D. Dey and F. Hutter},
  title = {Neural Architecture Search: Insights from 1000 Papers},
  journal = {arXiv:2301.08727 [cs.LG]},
  year = {2023}
}

@article{griffiths:2021,
  title = {Achieving Robustness to Aleatoric Uncertainty with Heteroscedastic Bayesian Optimisation},
  author = {Griffiths, Ryan-Rhys and Aldrick, Alexander A. and {Garcia-Ortegon}, Miguel and Lalchand, Vidhi and Lee, Alpha A.},
  year = {2021},
  month = nov,
  journal = {Machine Learning: Science and Technology},
  volume = {3},
  number = {1},
  pages = {015004},
  publisher = {IOP Publishing},
  issn = {2632-2153},
  doi = {10.1088/2632-2153/ac298c},
  url = {https://dx.doi.org/10.1088/2632-2153/ac298c},
}

@inproceedings{pham-icml18,
 title     = {Efficient Neural Architecture Search via Parameters Sharing},
 author    = {H. Pham and M. Guan and B. Zoph and Q. Le and J. Dean},
 booktitle = {Proceedings of the 35th International Conference on Machine Learning (ICML'18)},
 year      = {2018},
}

\end{document}